\documentclass{article}

\usepackage[preprint]{neurips_2026}

\usepackage[utf8]{inputenc} 
\usepackage[T1]{fontenc}    
\usepackage{hyperref}       
\usepackage{url}            
\usepackage{booktabs}       
\usepackage{amsfonts}       
\usepackage{nicefrac}       
\usepackage{microtype}      
\usepackage{xcolor}         

\usepackage{amsmath}
\usepackage{amssymb}
\usepackage{mathtools}
\usepackage{amsthm} 
\usepackage{cases}
\usepackage{bm}
\usepackage{bbm}
\usepackage{multirow}
\usepackage{colortbl}
\usepackage{listings}
\usepackage{makecell}
\usepackage{enumitem}
\usepackage{xurl}
\usepackage{graphicx}
\usepackage{subcaption}
\usepackage{caption}         
\usepackage[dvipsnames]{xcolor}            
\usepackage{wrapfig}
\usepackage{fancyvrb}
\usepackage{tcolorbox}  
\tcbuselibrary{breakable}
\usepackage{titletoc}
\usepackage{marvosym}

\title{WebGen-R1: Incentivizing Large Language Models to Generate Functional and Aesthetic Websites with Reinforcement Learning}

\author{
Juyong Jiang$^1$\,$^2$\thanks{Work done during a research internship at Alibaba.}\quad 
Chenglin Cai$^3$\quad
Chansung Park$^4$\quad
Jiasi Shen$^2$\quad
Sunghun Kim$^1$\\
\textbf{Jianguo Li}$^5$\quad
\textbf{Yue Wang}$^3$\thanks{Corresponding authors.}\\
$^1$The Hong Kong University of Science and Technology (Guangzhou) \\
$^2$The Hong Kong University of Science and Technology\\
$^3$Tongyi Lab, Alibaba Group\\
$^4$Electronics and Telecommunications Research Institute\,
$^5$Ant Group \\
\Letter~\texttt{yue.w@alibaba-inc.com}
}

\begin{document}

\maketitle

\begin{abstract}
While Large Language Models (LLMs) excel at function-level code generation, project-level tasks such as generating functional and visually aesthetic multi-page websites remain highly challenging.  
Existing works are often limited to single-page static websites, while agentic frameworks typically rely on multi-turn execution with proprietary models, leading to substantial token costs, high latency, and brittle integration.  
Training a small LLM end-to-end with reinforcement learning (RL) is a promising alternative, yet it faces a critical bottleneck in designing reliable and computationally feasible rewards for website generation. 
Unlike single-file coding tasks that can be verified by unit tests, website generation requires evaluating inherently subjective aesthetics, cross-page interactions, and functional correctness. 
To this end, we propose WebGen-R1, an end-to-end RL framework tailored for project-level website generation. 
We first introduce a scaffold-driven structured generation paradigm that constrains the large open-ended action space and preserves architectural integrity. 
We then design a novel cascaded multimodal reward that seamlessly couples structural guarantees with execution-grounded functional feedback and vision-based aesthetic supervision. 
Extensive experiments demonstrate that our WebGen-R1 substantially transforms a 7B base model from generating nearly nonfunctional websites into producing deployable, aesthetically aligned multi-page websites. 
Remarkably, our WebGen-R1 not only consistently outperforms heavily scaled open-source models (up to 72B), but also rivals the state-of-the-art DeepSeek-R1 (671B) in functional success, while substantially exceeding it in valid rendering and aesthetic alignment. 
These results position WebGen-R1 as a viable path for scaling small open models from function-level code generation to project-level web application generation. 
To facilitate future research, we release our code and data at \url{https://github.com/juyongjiang/WebGen-R1}. 
\end{abstract}

\section{Introduction}

Recent advances in Large Language Models (LLMs) have significantly expanded their capabilities in automated code generation \citep{jaech2024openai,hui2024qwen2,guo2025deepseek,team2025kimi,yang2025qwen3,park2026tarot,jiang2026reflexicoder}, achieving human-competitive performance on function-level benchmarks \citep{chen2021evaluating,li2022competition}. However, moving from generating isolated functions to architecting project-level software remains a frontier challenge \citep{jimenez2023swe,bi2024iterative,zhuo2025bigcodearena,zan2025multi,li2026osvbench,jiang2026survey}.

Among these challenges, end-to-end website generation, encompassing multi-page routing, dynamic functionality, modern user interface (UI) design, and responsive layouts, is particularly demanding \citep{wan2024mrweb,xiao2024interaction2code,lu2025webgen,zhang2025artifactsbench}. Unlike single-file coding tasks, real-world websites require consistent architectural patterns, multi-file codebases with intricate dependencies, long-range contextual coherence, and adherence to design principles that balance functionality with visual appeal. This requires reasoning not only over software engineering constraints but also over aesthetic and user-experience (UX) considerations, which have traditionally been difficult to formalize and evaluate in automated code generation.

Despite promising early progress, current approaches to LLM-driven website generation face significant scalability and quality limitations. One paradigm simplifies the task to single-page static sites \citep{webdev_arena}, abstracting away the intricacies of modern web applications, such as dynamic routing, state management, authentication flows, and cross-page navigation in modern web applications. Conversely, multi-agent orchestration frameworks attempt to decompose the task by assigning different specialized sub-agents to implement discrete subtasks, such as UI layout, backend logic, and testing, and then integrating their outputs \citep{hong2023metagpt,he2024webvoyager,he2024openwebvoyager,lu2025webgen}. However, such modularity introduces brittle inter-agent dependency chains, where small inconsistencies in contracts, file names, or interface definitions can cascade into non-functional builds. Although this issue can be mitigated through multi-turn execution with refined feedback, doing so results in substantial token costs and high latency. Furthermore, neither paradigm effectively optimizes for the intersection of functionality and aesthetics, often yielding websites that are technically executable but visually incoherent or functionally sterile. 

To address these limitations, we propose WebGen-R1, a reinforcement learning (RL) framework designed to incentivize the end-to-end generation of functional and aesthetic multi-page websites. However, integrating RL into such open-ended generative tasks presents a critical bottleneck in \textit{designing a reward signal that is both reliable and computationally feasible} \citep{guo2025deepseek,zeng2025simplerl,wen2025reinforcement,yang2025depth,mroueh2025reinforcement}. Unlike single-file coding tasks that can be verified by unit tests, verifying a website's functionality requires interacting with it, clicking buttons, navigating routes, and checking state changes. While deploying GUI agents for automated exploration seems promising, it is prohibitively expensive for RL training. The vast exploration space of web interactions, combined with the substantial token costs and high latency of multi-turn execution, introduces unacceptable cost and training delay. Moreover, the reliability of such agents is largely limited by the capabilities of their underlying models, often leading to noisy reward signals. Relying solely on visual inspection via Vision-Language Models (VLMs) also leaves an important gap, since a visually perfect interface may still conceal broken event handlers or invalid logic chains, rendering the page non-functional despite its appearance.

To resolve this dilemma, we first propose a scaffold-driven structured generation paradigm that constrains the large open-ended action space and preserves architectural integrity. Specifically, instead of generating the whole project from scratch, the LLM operates within a standardized, pre-validated \texttt{React} template. It generates and modifies components within this robust skeleton, ensuring that the fundamental build pipeline, routing architecture, and server configurations are inherently correct by design. Subsequently, we implement a rigorous filtering stage, namely a hierarchical verification and rendering pipeline, before costly reward modeling. This stage includes static compliance verification, which checks the required artifact/action structure, the presence of project files and commands, and selected content-level rules, followed by automated build and rendering. This step eliminates low-level execution errors and provides immediate, dense penalty signals for structural hallucinations. 
Built on this template-driven, statically verified foundation that strongly guarantees functionality and executability, we further propose a novel cascaded multimodal reward that leverages VLMs for semantic visual alignment.
The VLM assesses the rendered website for aesthetic harmony, such as UI layout, typographic consistency, visual cues of functionality, and stylistic alignment with the user's intent. 

We empirically demonstrate that our approach is both efficient and effective. By constraining the action space to valid architectural patterns and rewarding structural compliance, WebGen-R1 significantly reduces the generation of non-renderable code. Extensive experiments on real-world benchmarks reveal that WebGen-R1 substantially transforms a 7B base model from generating nearly nonfunctional websites into producing deployable, aesthetically aligned multi-page websites. We observe a dramatic increase in the valid render ratio from 30.56\% to 95.89\%, alongside a 44.32\% improvement in aesthetic scoring and a leap in functional quality metrics from 1.59\% to 29.21\%. Remarkably, WebGen-R1 not only consistently outperforms heavily scaled open-source models up to 72B, but also rivals the state-of-the-art DeepSeek-R1 (671B) in functional success, while substantially exceeding it in valid rendering and aesthetic alignment. These results position WebGen-R1 as a viable path for scaling small open models from function-level code generation to project-level web application generation. 
In summary, our contributions are as follows:
\begin{itemize}
\item We introduce WebGen-R1, the first RL framework for optimizing end-to-end multi-page website generation in small open-source LLMs, eliminating the need for complex multi-agent orchestration with proprietary models.
\item We propose a scaffold-driven structured generation paradigm with a hierarchical verification and rendering pipeline, ensuring high functional reliability without the prohibitive cost of GUI-agent exploration.
\item We design a cascaded multimodal reward model that seamlessly couples structural guarantees with execution-grounded functional feedback and vision-based aesthetic supervision, aligning optimization objectives with human standards for both functional correctness and visual design.
\item We establish a multi-dimensional evaluation protocol and demonstrate through extensive experiments on real-world benchmarks that our WebGen-R1 is both efficient and effective, scaling small open models from function-level code generation to project-level web application generation.
\end{itemize}
\section{Related Work}\label{sec:related_work} 

\subsection{LLMs for Project‑Level Code Generation}
Large language models (LLMs) have demonstrated remarkable proficiency in function-level code generation, achieving near-human performance on function-level benchmarks \citep{chen2021evaluating,austin2021program,li2022competition,park2026tarot,jiang2026reflexicoder}. Advances in instruction tuning \citep{ouyang2022training,wang2023self,chung2024scaling,wang2024kasa,park2025llamaduo} and tool-augmented prompting \citep{schick2023toolformer,zhang2023repocoder} have further improved zero-shot and few-shot code generation capabilities. However, these achievements predominantly concern single-file or self-contained scripts, often constrained to producing one function or module per task. Such settings abstract away the complexities of real-world software engineering, where projects span multiple files, require intricate inter-module dependencies, and must satisfy both functional and non-functional requirements. 
Compared to function-level coding, project-level code generation poses qualitatively different challenges. Prior attempts to extend LLMs to multi-file outputs include hierarchical prompting \citep{shrivastava2023repository,zhang2023repocoder}, iterative refinement \citep{chen2023teaching,olausson2023self,shinn2024reflexion}, and agent-based pipelines \citep{li2024codetree,zhang2024codeagent,deepswe2025}. 
In the context of web development, one line of work simplifies the task to single-page static sites \citep{webdev_arena}, abstracting away the intricacies of modern web applications, such as dynamic routing, state management, authentication flows, and cross-page navigation. Multi-agent approaches \citep{hong2023metagpt,he2024webvoyager,he2024openwebvoyager,lu2025webgen} partition functionality across specialized sub-agents, such as front-end generation, API design, and testing. However, integration often suffers from inconsistent shared states and fragile inter-component linking. In addition, multi-turn execution results in substantial token costs and high latency. These shortcomings may result in generated projects that build and run but still fail to align with the holistic end-to-end specifications of real-world websites. 
This work explores employing a small open-source LLM to generate project-level multi-page websites in an end-to-end manner, without relying on task decomposition or agent orchestration with proprietary models.

\subsection{Reinforcement Learning for Code Generation}
Reinforcement learning (RL) has emerged as a crucial technique for aligning LLM behavior with human preferences and task-specific objectives \citep{ouyang2022training,achiam2023gpt,bai2022constitutional,lee2023rlaif}. Yet adapting RL to open-ended code generation introduces unique obstacles, including a vast action space, undefined or ambiguous ground truth, and outputs that cannot be trivially benchmarked against static gold standards. 
Reinforcement Learning with Verifiable Rewards (RLVR) \citep{shao2024deepseekmath,guo2025deepseek,team2025kimi,yu2025dapo} addresses part of this issue through deterministic binary success checks, such as unit test pass rates, which work well for algorithmic correctness but fail to capture subjective quality dimensions such as style, maintainability, or visual experience. Existing RL-based work on coding tasks \citep{le2022coderl,shen2023pangu,shojaee2023execution,dou2024stepcoder,deepcoder2025} therefore typically optimizes purely functional metrics, leaving a large gap for project-level tasks such as website generation, where both functional correctness and aesthetics are first-class objectives. 
While prior works \citep{hong2023metagpt,he2024webvoyager,he2024openwebvoyager,lu2025webgen,webdev_arena} have advanced LLM-driven website generation, none simultaneously optimizes multi-file structural dependencies, execution validity, functional correctness, and visual quality within an integrated RL framework. 
Our work shows that RL can transform a small open-source LLM from generating nearly nonfunctional websites into producing deployable multi-page websites with substantially improved functional correctness and visual quality, suggesting a viable path toward project-level code generation. 

\begin{figure*}[t]
	\centering
	\includegraphics[width=\linewidth]{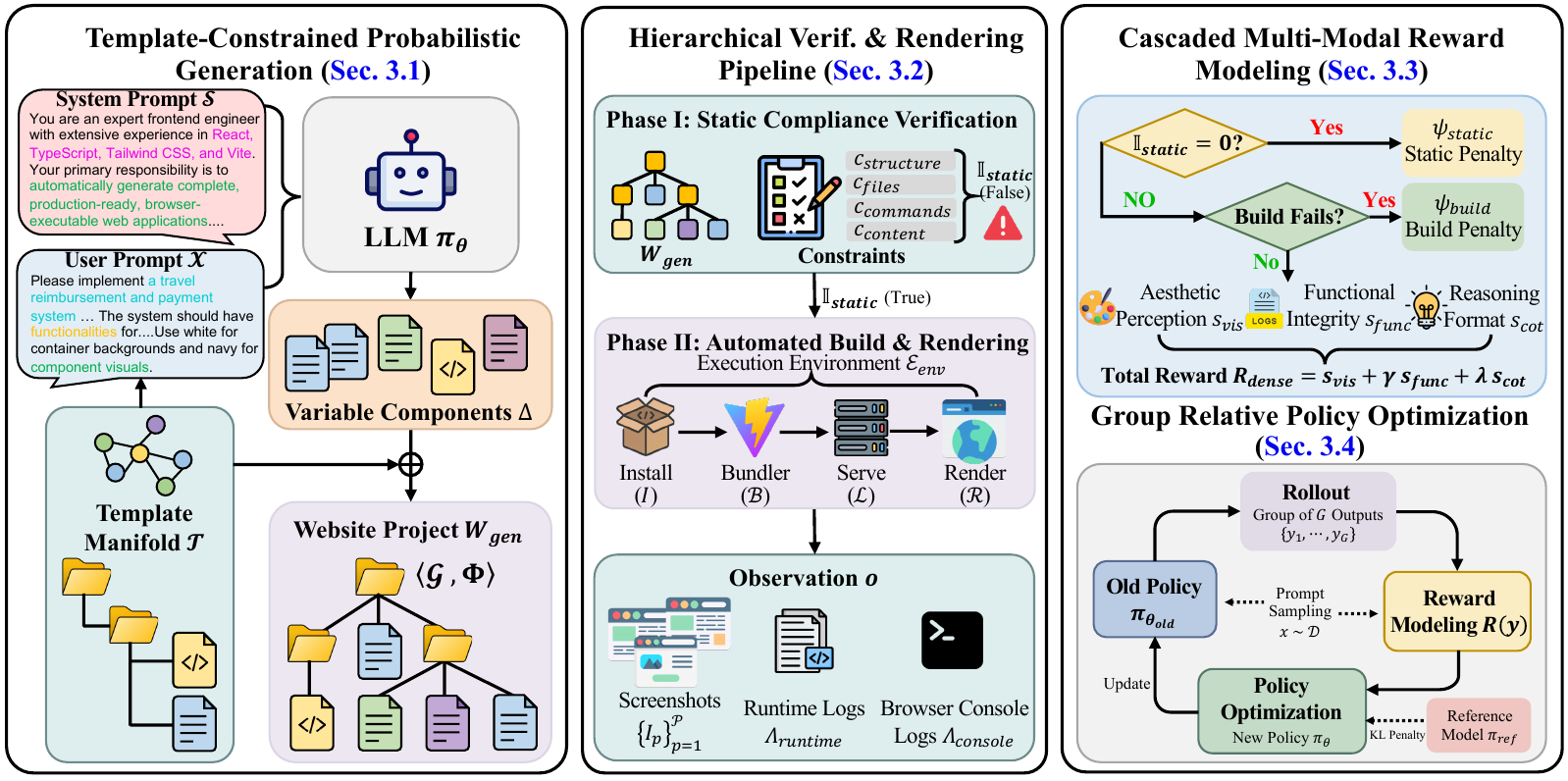} 
    \caption{Overview of WebGen-R1, a reinforcement learning framework for functional and aesthetic website generation. Given a system prompt, a user specification, and a template manifold, the policy $\pi_\theta$ generates only the variable project components, which are then injected into predefined scaffold slots to form a complete website project. The generated project is evaluated by a hierarchical verification and rendering pipeline. In Phase I, static compliance verification checks structural, file-level, command-level, and content-level constraints and terminates invalid samples early. In Phase II, statically valid projects proceed through dependency installation, bundling, local serving, and browser-based rendering, producing screenshots together with runtime and console logs. Based on these observations, WebGen-R1 uses a cascaded multimodal reward that assigns scalar rewards to static or execution failures and otherwise combines a VLM-based aesthetic perception score, a functional integrity score derived from execution logs, and a reasoning format reward. The resulting reward is used to optimize the model with GRPO. 
    This design combines template-level structural constraints, execution-grounded functional reward supervision, and visually informed aesthetic reward supervision, enabling efficient end-to-end RL for high-quality multi-page website generation.}
\label{fig:webgen-r1}
\end{figure*}

\section{Methodology} 

We formulate automated end-to-end website generation as a conditional structured generation problem under template constraints. 
Instead of generating the whole project from scratch, WebGen-R1 operates within a scaffold-driven structured generation paradigm based on a standardized, pre-validated \texttt{React} template. 
In this framework, the scaffold guarantees the structural integrity of the project, while the LLM is responsible for the semantic content and visual realization.

\subsection{Template-Constrained Probabilistic Generation}

Let $\mathcal{D}$ denote the space of natural language specifications and $\mathcal{W}$ denote the space of valid website projects. A valid project $W \in \mathcal{W}$ is defined as a tuple $W = \langle \mathcal{G}, \Phi \rangle$, where $\mathcal{G} = (\mathcal{V}, \mathcal{E})$ represents the directory graph structure, with nodes $\mathcal{V}$ as files or folders and edges $\mathcal{E}$ as import dependencies, and $\Phi = \{f_v\}_{v \in \mathcal{V}}$ denotes the semantic content of each file. 
To reduce the brittle dependency chains common in multi-page website generation, we introduce a fixed template manifold $\mathcal{T}$. The template defines an invariant subgraph $\mathcal{G}_{\mathcal{T}} \subset \mathcal{G}$ that contains the build configuration, routing skeleton, and server-side logic. The policy $\pi_\theta$ is restricted to generating the variable components outside the fixed scaffold, denoted by $\Delta = W \setminus \mathcal{T}$. Formally, given a system prompt $\mathcal{S}$, a user prompt $x \in \mathcal{D}$, and the template context $\mathcal{T}$, website generation is modeled as:
\begin{equation}
W_{\text{gen}} = \mathcal{T} \oplus \pi_{\theta}(\Delta \mid \mathcal{S}, x, \mathcal{T}),
\end{equation}
where $\oplus$ denotes the injection of generated component code into predefined scaffold slots. The policy $\pi_\theta$ is factorized autoregressively over the tokens of the variable components:
\begin{equation}
\pi_{\theta}(\Delta \mid \mathcal{S}, x, \mathcal{T}) = \prod_{k=1}^{K} \prod_{t=1}^{T_k} P(y_{k,t} \mid \mathcal{S}, x, \mathcal{T}, \Delta_{<k}, y_{k,<t}),
\end{equation}
where $K$ is the number of required components, such as pages, commands, and styles, and $y_{k,t}$ is the $t$-th token of the $k$-th component. This formulation ensures that the fundamental structural properties of the website, such as a valid entry point and correct build scripts, are satisfied a priori.

\subsection{Hierarchical Verification and Rendering Pipeline} 
To eliminate low-level execution errors, we introduce a rigorous filtering stage before costly reward modeling. We model the environment transition as a composite function $\mathcal{H}: \mathcal{W} \to \mathcal{O}$ that maps a generated project to an observation state $o$. This process is composed of two distinct phases as follows.

\textbf{Phase I. Static Compliance Verification.}
Before execution, the generated website $W_{\text{gen}}$ undergoes a static validation procedure $\mathbb{I}_{\text{static}}(\cdot)$. We define a set of constraints
$\mathcal{C} = \{c_{\text{structure}}, c_{\text{files}}, c_{\text{commands}}, c_{\text{content}}\}$,
which verify compliance with the required website structure, the presence of project files and commands, and selected content-level rules such as valid ``package.json'' fields and component export patterns. The static validation procedure is formally defined as:
\begin{equation}
\mathbb{I}_{\text{static}}(W_{\text{gen}}) = \prod_{c \in \mathcal{C}} \mathbbm{1}\left[ c(W_{\text{gen}}) \text{ is satisfied} \right],
\end{equation}
where $\mathbbm{1}[\cdot]$ denotes the indicator function. If $\mathbb{I}_{\text{static}} = 0$, the process terminates early and returns rule-based failure signals for the generated website format and project scaffold.

\textbf{Phase II. Automated Build and Rendering.}
If $\mathbb{I}_{\text{static}} = 1$, the project enters the execution environment $\mathcal{E}_{\text{env}}$. In our setup, this stage is implemented as a deterministic pipeline: 
\begin{equation}
o = \mathcal{R}_{\text{render}}\left( \mathcal{L}_{\text{serve}}\left( \mathcal{B}_{\text{bundler}}\left( \mathcal{I}_{\text{install}}(W_{\text{gen}}) \right) \right) \right),
\end{equation}
where $\mathcal{I}$ resolves package dependencies, $\mathcal{B}$ executes the build script such as Vite or Webpack, $\mathcal{L}$ starts the local server, and $\mathcal{R}$ captures the visual state with a headless browser. The resulting observation $o$ is a tuple:
\begin{equation}
o = \left\langle \{I_p\}_{p \in \mathcal{P}}, \Lambda_{\text{runtime}}, \Lambda_{\text{console}} \right\rangle,
\end{equation}
where $\{I_p\}_{p \in \mathcal{P}}$ denotes screenshots over a set of routes $\mathcal{P}$, and $\Lambda_{\text{runtime}}$ and $\Lambda_{\text{console}}$ denote runtime logs and browser console logs, respectively.

\subsection{Cascaded Multi-Modal Reward Modeling} 
Designing a reliable and computationally feasible reward model for website generation is challenging because functional correctness depends on objective rule-based verification, while aesthetic quality and user experience (UX) depend on subjective and continuous perceptual assessment.
To address this challenge, we propose a cascaded multimodal reward $R(W_{\text{gen}}, o, x)$ that strictly penalizes structural failures while incentivizing aesthetic alignment, including UI layout, typographic consistency, visual functional correctness, and stylistic alignment with the user's intent, via Vision-Language Models (VLMs).
Formally, the total reward is computed hierarchically:
\begin{equation}\label{eq:reward}
R(y) = R(W_{\text{gen}}, o, x) =
\begin{cases}
\psi_{\text{static}}(W_{\text{gen}}) & \text{if } \mathbb{I}_{\text{static}}(W_{\text{gen}}) = 0 \\
\psi_{\text{build}}(\Lambda_{\text{runtime}}) & \text{if build fails} \\
R_{\text{dense}}(y, o, x) & \text{otherwise}
\end{cases}
\end{equation}
where $\psi_{\text{static}}$ and $\psi_{\text{build}}$ map static-analysis and build failures to scalar rewards, respectively. For simplicity, in this work we set both mappings to 0.
For successfully rendered projects, the dense reward $R_{\text{dense}}$ is a weighted linear combination of three objectives, where $\gamma$ and $\lambda$ are weighting coefficients:
\begin{equation}
R_{\text{dense}} = s_{\text{vis}} + \gamma \cdot s_{\text{func}} + \lambda \cdot s_{\text{cot}}.
\end{equation}

\begin{itemize}
    \item \textbf{Aesthetic Perception Score ($s_{\text{vis}}$).} We employ a VLM as a proxy for human preference judgment. The VLM processes the rendered screenshots $\{I_p\}_{p \in \mathcal{P}}$ together with the prompt $x$ and outputs a scalar score $s_{\text{vis}} \in [0, 5]$. This score reflects layout harmony, color palettes, typographic hierarchy, visually observable functional quality, and stylistic alignment with the user's intent.
    \item \textbf{Functional Integrity Score ($s_{\text{func}}$).} This reward quantifies the absence of runtime errors. Let $N_{\text{err}}$ denote the number of errors in $\Lambda_{\text{console}}$ and $\Lambda_{\text{runtime}}$. We define $s_{\text{func}} = 1$ if $N_{\text{err}} = 0$, and $s_{\text{func}} = 0$ otherwise, thereby encouraging error-free execution.
    \item \textbf{Reasoning Format Score ($s_{\text{cot}}$).} To encourage the model to plan the component architecture before coding, such as organizing directory hierarchies, configuring frameworks appropriately, and maintaining coherent shared state, we reward outputs that contain structured chain-of-thought (CoT) enclosed in explicit tags. $s_{\text{cot}}$ is a binary format reward that equals 1 when the reasoning is enclosed in the required \texttt{\textless think\textgreater} tags, and 0 otherwise.
\end{itemize}

This cascaded reward design ensures that expensive visual perception reasoning is only triggered for candidates that satisfy structural requirements, which significantly improves the computational efficiency of RL training.

\subsection{Group Relative Policy Optimization} 
We formulate the RL objective as maximizing the expected reward $R(y)$ in Eq. \ref{eq:reward}.
A key challenge in website generation is the high variance of the reward, since even a minor syntax error can lead to rendering failure and reduce the reward from a high value to zero. Such sparse and highly volatile reward signals make standard actor-critic methods such as PPO \citep{schulman2017proximal} difficult to optimize reliably.
Therefore, we optimize the policy $\pi_\theta$ using Group Relative Policy Optimization (GRPO) \citep{shao2024deepseekmath}, which computes group-wise relative advantages by normalizing rewards within a set of sampled outputs for the same prompt and using the group mean as a dynamic baseline. This design reduces variance and improves optimization stability.
Specifically, for each task $x$, we sample a group of $G$ outputs $\{y_1, \dots, y_G\}$ from the old policy $\pi_{\theta_{\text{old}}}$. The objective function is formally defined as:  
\begin{equation}
\begin{aligned}
\mathcal{J}(\theta) =
\mathbb{E}_{x \sim \mathcal{D}, \{y_i\}_{i=1}^G \sim \pi_{\theta_{\mathrm{old}}}(\cdot \mid x)} \Bigg[ \frac{1}{G} \sum_{i=1}^G \frac{1}{|y_i|} \sum_{t=1}^{|y_i|} 
\bigg( & \min \Big[r_{i,t}(\theta) \, \hat{A}_{i,t}, \mathrm{clip}\big(r_{i,t}(\theta), 1 - \varepsilon, 1 + \varepsilon\big) \, \hat{A}_{i,t} \Big] \\
& - \beta \, D_{\mathrm{KL}}\big( \pi_{\theta}(\cdot \mid x) \| \pi_{\mathrm{ref}}(\cdot \mid x) \big) \bigg) \Bigg],
\label{eq:grpo_obj_final}
\end{aligned}
\end{equation}
\begin{equation}
\hat{A}_{i,t} \triangleq \frac{R(y_i) - \mathrm{mean}(\{ R(y_j) \}_{j=1}^G)}{\mathrm{std}(\{ R(y_j) \}_{j=1}^G)}, \quad r_{i,t}(\theta) \triangleq \frac{\pi_{\theta}\big( y_{i,t} \mid x, y_{i,<t} \big)}{\pi_{\theta_{\mathrm{old}}}\big( y_{i,t} \mid x, y_{i,<t} \big)},
\label{eq:grpo_ratio}
\end{equation}
where $\hat{A}_{i,t}$ denotes the group-relative normalized advantage for token position $t$ in response $y_i$, $\varepsilon > 0$ is the clipping parameter, and $\beta$ controls the strength of the KL regularization.
This normalization enables the model to learn relative preferences among candidate outputs for the same prompt, even when absolute reward scales vary substantially across tasks. The KL term $D_{\text{KL}}$ constrains the policy from drifting excessively from the reference model, thereby preserving the syntactic and linguistic quality of the generated code.
\section{Experiments}
\subsection{Experimental Setup} 
\begin{figure*}[t]
    \centering
    \begin{subfigure}[t]{0.4\linewidth}
        \centering
        \includegraphics[width=\linewidth]{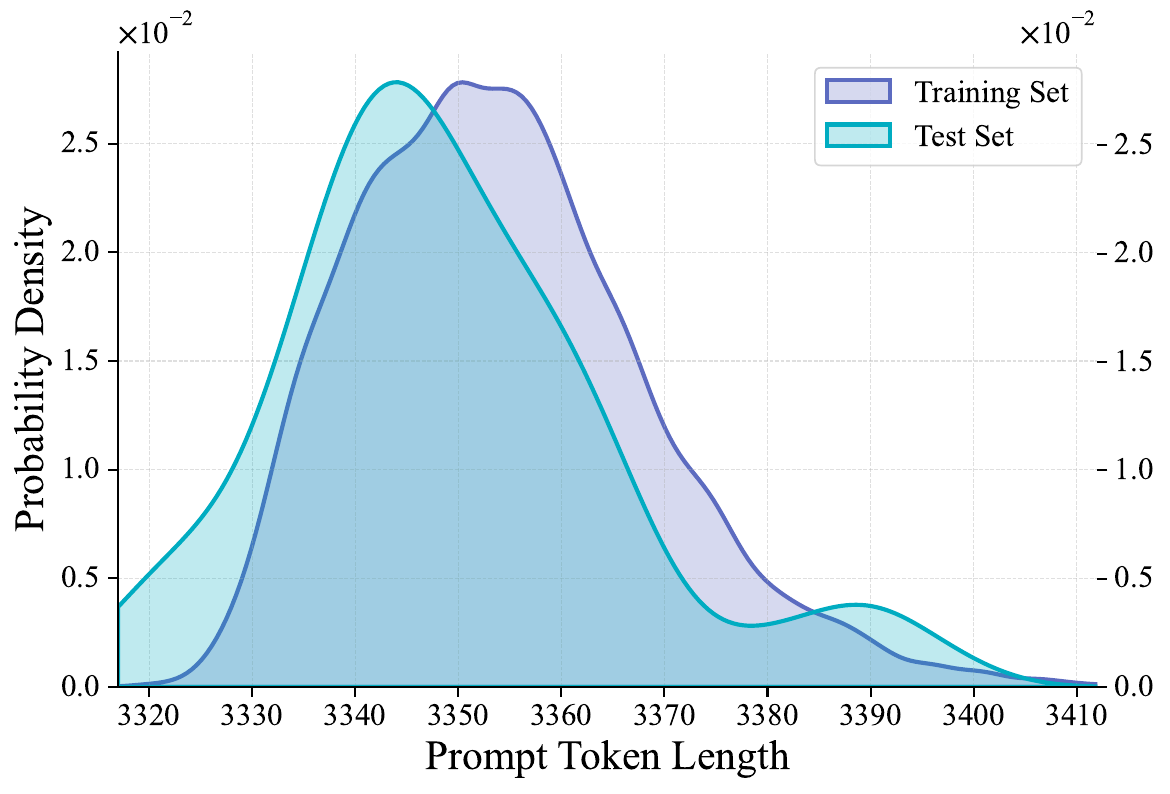}
    \end{subfigure}
    \begin{subfigure}[t]{0.4\linewidth}
        \centering
        \includegraphics[width=\linewidth]{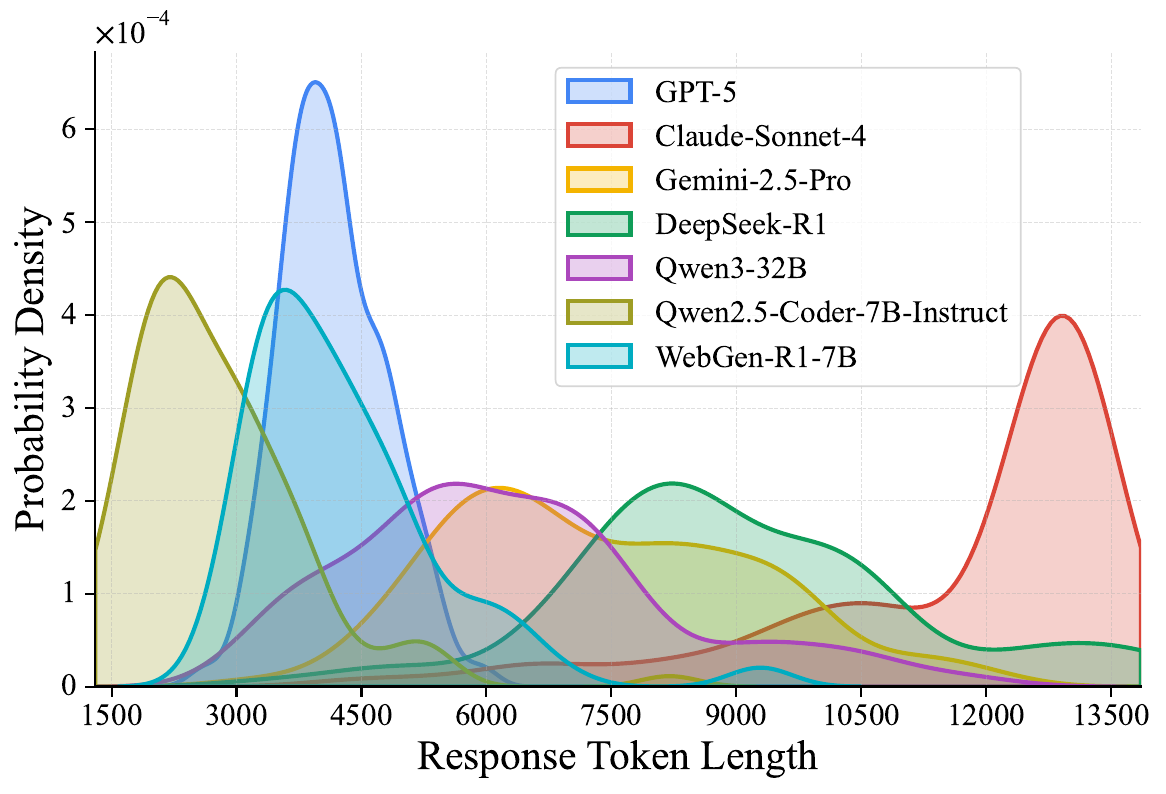}
    \end{subfigure}
    \begin{subfigure}[t]{0.4\linewidth}
        \centering
        \includegraphics[width=\linewidth]{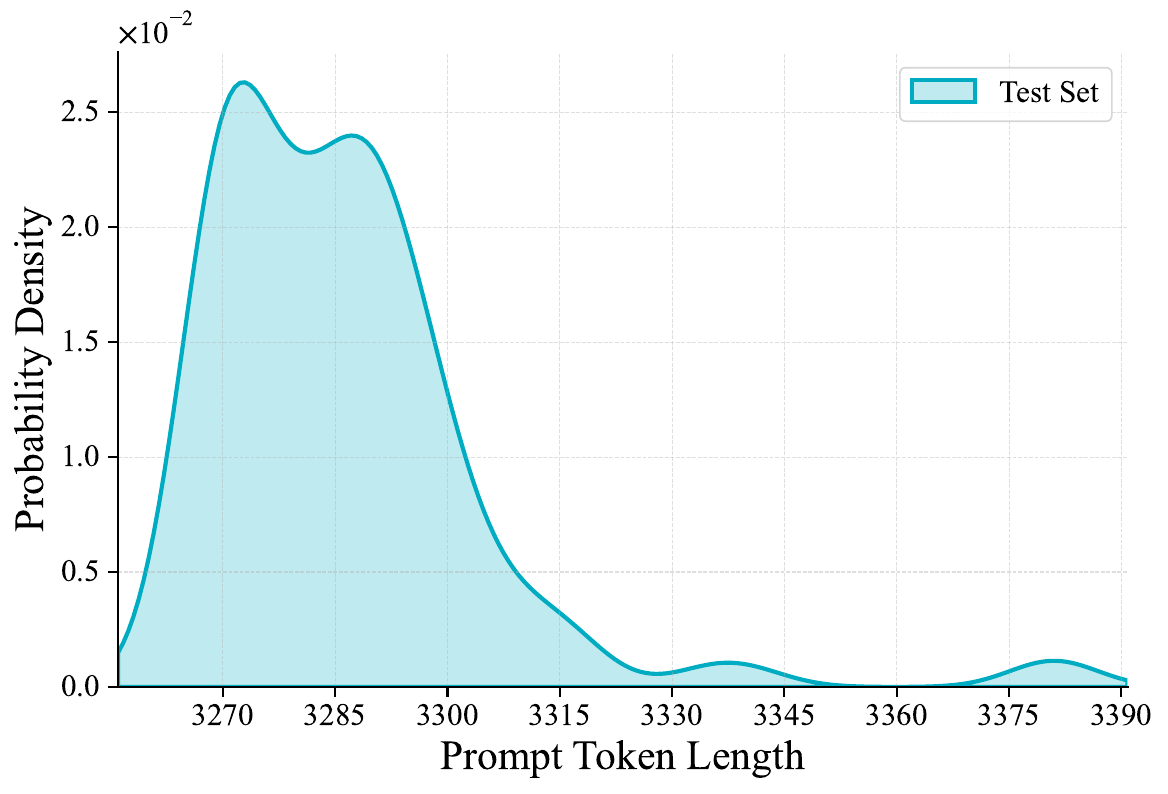}
        \label{fig:sub1}
    \end{subfigure}
    \begin{subfigure}[t]{0.4\linewidth}
        \centering
        \includegraphics[width=\linewidth]{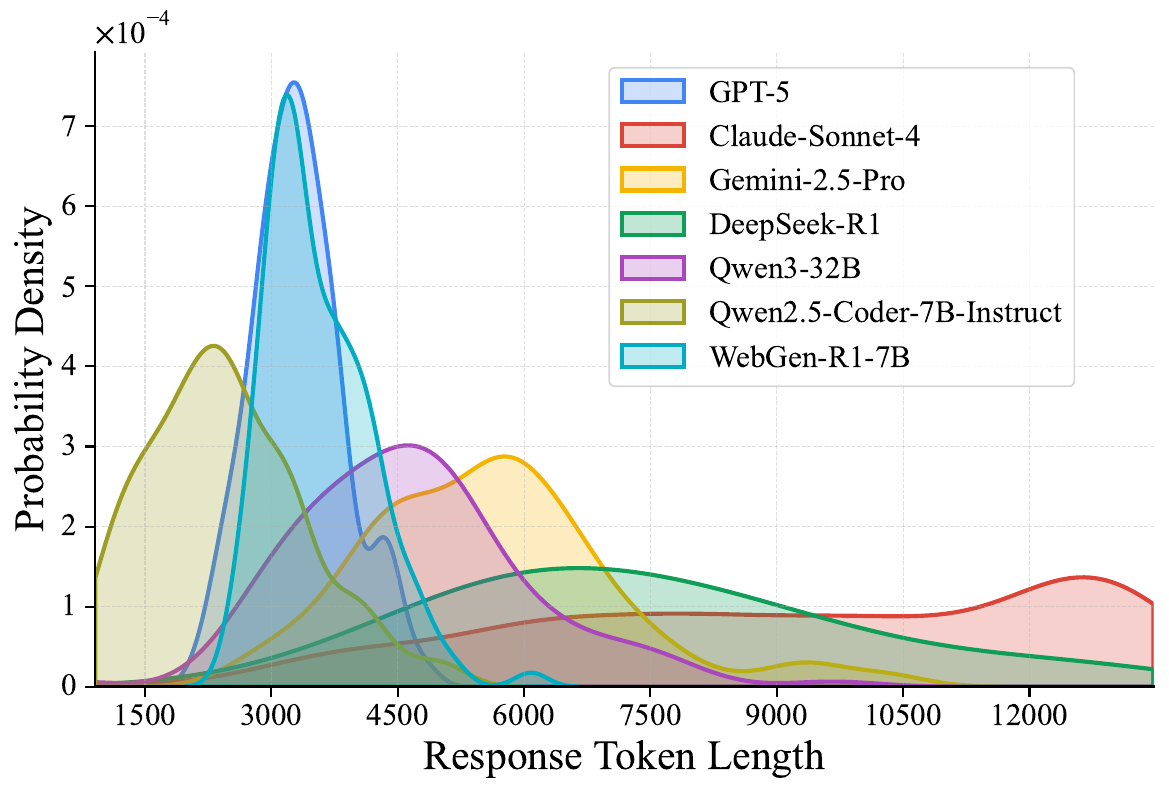}
        \label{fig:sub2}
    \end{subfigure}
    \caption{
    Token length distributions of prompts and generated responses for several state-of-the-art LLMs on end-to-end multi-page website generation tasks across WebGen-Instruct, WebGen-Bench, and WebDev Arena. For each instance, the prompt is formed by concatenating the system prompt with the corresponding natural-language website design instruction. 
    The first two panels show the prompt length distributions on WebGen-Instruct (training set) and WebGen-Bench (test set), together with the response length distributions aggregated over WebGen-Bench. 
    The last two panels show the prompt length distributions and response length distributions aggregated over WebDev Arena (test set). These statistics provide a quantitative characterization of input and output lengths across models and datasets in realistic website generation scenarios.}
    \label{fig:data}
\end{figure*}

\textbf{Datasets and Benchmarks.}  
We use WebGen-Instruct \citep{lu2025webgen} as our training corpus. It comprises 6,667 end-to-end website generation tasks covering a broad range of real-world web application domains.  
For evaluation, we use the WebGen-Bench \citep{lu2025webgen} benchmark, which contains 101 carefully curated website generation tasks ranging from minimalist portfolios to complex data-driven dashboards, including e-commerce frontends and real-time trackers. 
Each benchmark instance is paired with a comprehensive and repeatedly validated test suite, which enables reliable measurement of both functional behavior and stylistic conformance of generated websites. 
The natural-language task descriptions explicitly specify both functional requirements and visual design expectations, enabling precise evaluation.  
To evaluate out-of-distribution generalization, we curate a specialized evaluation set from the WebDev-Arena open-source dataset \footnote{\href{https://huggingface.co/datasets/lmarena-ai/webdev-arena-preference-10k}{https://huggingface.co/datasets/lmarena-ai/webdev-arena-preference-10k}}, which contains 10,000 real-world web development tasks. 
Given the noise in large-scale scraped data, we adopt an LLM-as-a-Judge strategy to filter high-quality and representative front-end development tasks. 
The prompt used for this filtering process is provided in Table~\ref{prompt:judgement}. This procedure yields a benchmark with 119 high-fidelity tasks. 
Figure \ref{fig:data} shows the distributions of prompt and response lengths. Each prompt is formed by concatenating the system prompt with the website task specification, and the responses are collected from several leading LLMs on end-to-end website generation tasks. 
Detailed data statistics and analysis are provided in Appendix \ref{sec:data_analysis}.

\textbf{Baselines.} 
We compare WebGen-R1 with 15 advanced LLMs, including eight proprietary models and seven open-source models. Details are provided in Appendix \ref{sec:details}.    
We exclude WebGen-LM \citep{lu2025webgen} because it is fine-tuned on Bolt.diy \footnote{\href{https://github.com/stackblitz-labs/bolt.diy}{https://github.com/stackblitz-labs/bolt.diy}} website generation agent trajectories collected with DeepSeek-V3 \citep{liu2024deepseek}. 
As a result, the model is closely tied to a specific agent-based framework, which limits its applicability to our end-to-end generation setting. Empirically, we also observe that it produces almost entirely non-functional websites in this setting.

\textbf{Metrics.}
To thoroughly evaluate LLMs for end-to-end website generation, we establish several quantitative metrics.  
(1) Functional Success Rate (FSR) is the percentage of generated websites that pass predefined interactive checks such as button clicks and form submissions.  
(2) Aesthetic Alignment Score (AAS) is the average score assigned by the VLM, which measures how well the rendered webpage aligns with user requirements and human aesthetic preferences. Unlike FSR, which directly measures functional correctness, AAS can reflect certain aspects of functionality through visual cues, such as the presence of forms, dropdown menus, or search inputs. 
(3) Valid Render Ratio (VRR) is the percentage of generated websites that can be rendered successfully without execution errors. 
(4) Lint \& Dependency Pass Rate (LDPR) is the fraction of projects that pass static code analysis, such as ESLint, and automatically resolve dependencies, which indicates readiness for deployment.  
Together, these metrics provide a multi-dimensional evaluation protocol that covers functional correctness, aesthetic quality, execution reliability, and deployability.

\textbf{Implementation Details.} 
We conduct all experiments on a cluster with $8 \times$ NVIDIA H100 GPUs (80\,GB) using the TRL framework~\citep{openr1}.  
The training pipeline begins with a supervised fine-tuning warm-up stage to instill structural priors for complex web development projects. We sample 600 instances from WebGen-Instruct based on the ``\texttt{application\_type}'' field to preserve the original domain distribution and avoid generalization drift. For these tasks, silver-standard responses are distilled from GPT-4.1-04-14 with temperature set to 0.6, nucleus sampling $\mathrm{top}_p$ set to 0.95, and a maximum token limit of 8,192 to ensure complete project generation in a single inference step. We then fine-tune Qwen2.5-Coder-7B-Instruct for 2 epochs with a learning rate of $1.0 \times 10^{-5}$, a batch size of 32, a maximum sequence length of 32k tokens, and a warmup ratio of 0.03.  
We subsequently further fine-tune the model for 400 optimization steps under the RL objective defined in Eq. \ref{eq:grpo_obj_final}. 
The reward weights are set to $\gamma=0.1$ and $\lambda=0.1$ to balance the optimization landscape.  
The training hyperparameters are a global batch size of $256$, group size $G = 16$, clipping parameter $\epsilon = 0.2$, learning rate $\mathrm{lr} = 5 \times 10^{-6}$, and KL-divergence coefficient $\beta = 0.01$. 
The maximum context length is set to $4{,}096$ tokens for prompts and $8{,}192$ tokens for model outputs. 
For inference, we use a decoding temperature of $0.7$ and a nucleus sampling value $\mathrm{top}_p$ of $0.95$.   
Following \citep{lu2025webgen}, we use WebVoyager \citep{he2024webvoyager} as the GUI-agent framework to evaluate the functional correctness of generated websites, since it is specifically designed for web environments. 
For each WebGen-Bench task, WebVoyager performs the required interactive behaviors, such as button clicks, form submissions, and multi-page navigation, while monitoring DOM changes and UI responses to verify whether the benchmark-defined behavior checks are satisfied. 
We compute FSR as the proportion of tasks that satisfy all checks. This automated interaction testing provides a scalable and reproducible evaluation of functional correctness without human annotators. 
To reduce cost and improve efficiency, we standardize all multimodal components, including the GUI-agent engine and VLM, on GPT-4o-11-20 during both training and evaluation.  
Additional implementation details are provided in Appendix \ref{sec:details}.

\begin{table*}[t]
\centering
\small
\setlength{\tabcolsep}{5pt} 
\caption{
Performance comparison of WebGen-R1 and various state-of-the-art LLMs from multiple institutions on the WebGen-Bench benchmark, evaluated by FSR, AAS, and VRR. We also report the score improvement ($\pm$) of our WebGen-R1 relative to the base model, Qwen2.5-Coder-7B-Instruct. Bold values indicate the best results.
}
\begin{tabular}{l|l|ccc}
\toprule
\textbf{Institution} & \textbf{Model} & \textbf{FSR}(\%) & \textbf{AAS} & \textbf{VRR}(\%) \\
\midrule
\multirow{5}{*}{\includegraphics[height=8pt]{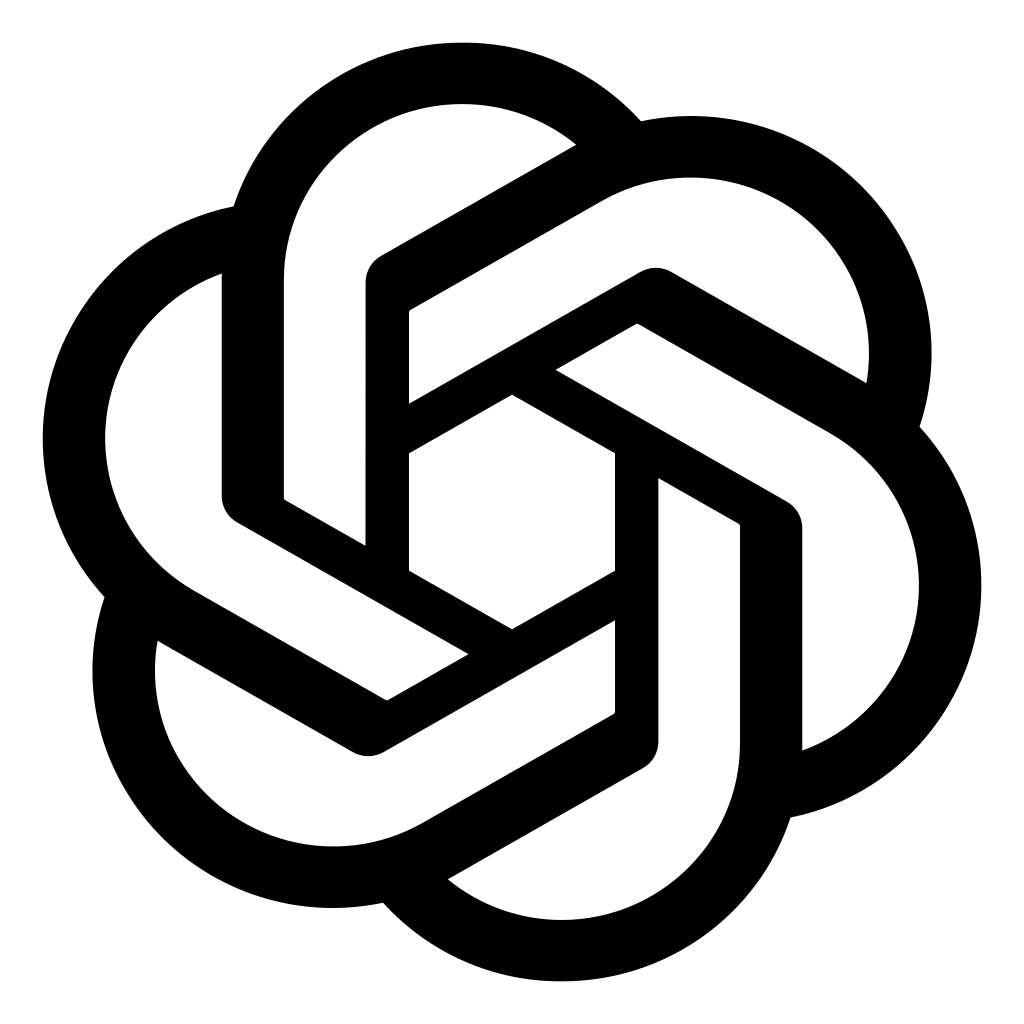} OpenAI}
 & GPT-5       &   46.53     &  3.34    &   90.43    \\
 & GPT-4.1     &   43.91     &   3.78  &   82.09    \\
 & o3          &   42.86    &    3.55   &  81.08      \\
 & o4-mini     &   27.29    &   3.31   &   56.52     \\
 & GPT-4o      &   21.60     &   3.31    &  85.71    \\
\midrule
\multirow{2}{*}{\includegraphics[height=6pt]{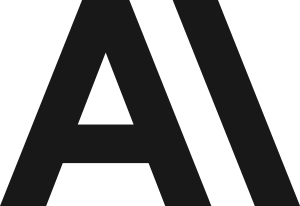} Anthropic}
 & Claude-Sonnet-4   &   46.13     &   3.86   &  86.05       \\
 & Claude-3.7-Sonnet &   \textbf{57.72}     &    3.90   &  84.00    \\
\midrule
\multirow{1}{*}{\includegraphics[height=8pt]{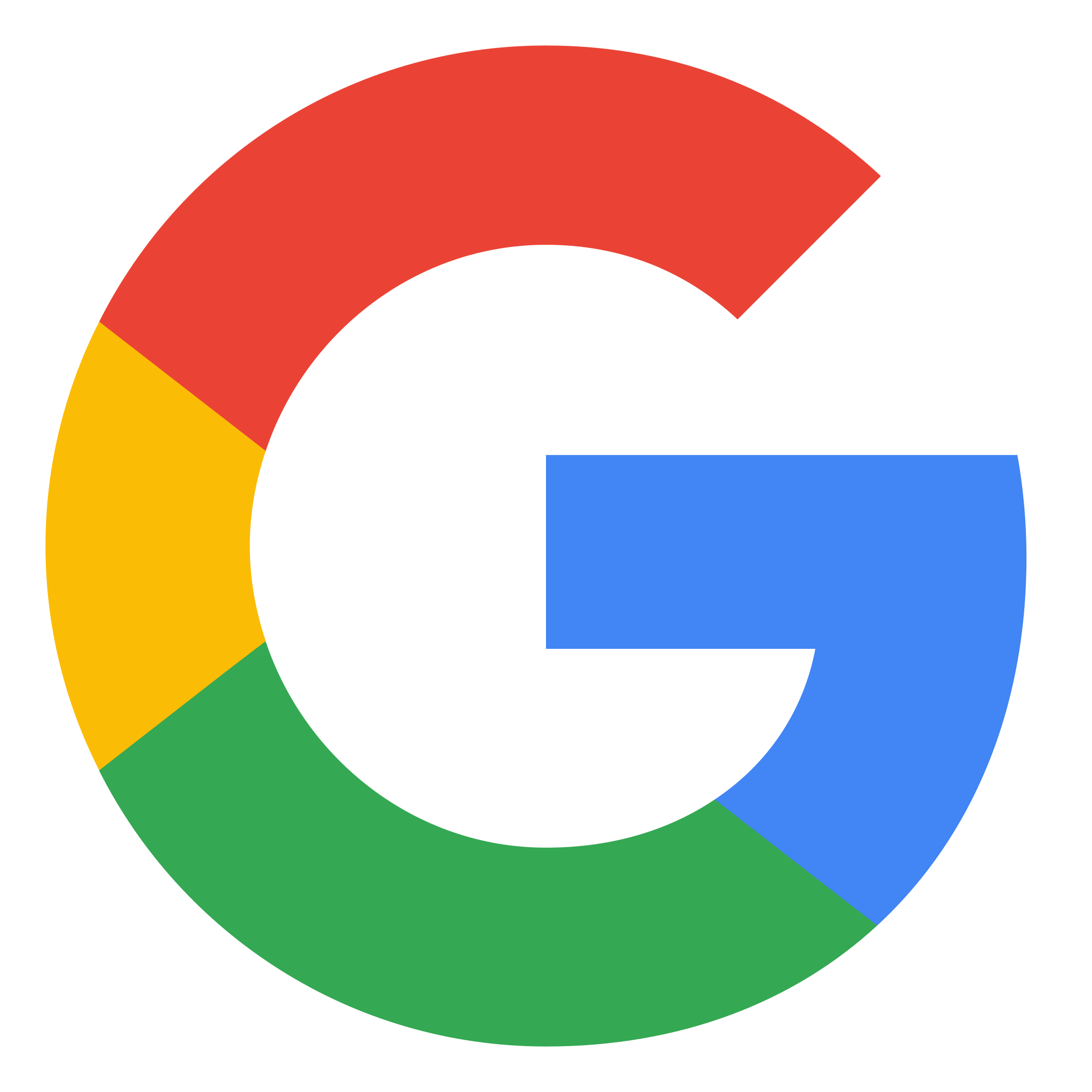} Google}
 & Gemini-2.5-Pro    &   36.31      &   3.89   &   83.33    \\
\midrule
\multirow{1}{*}{\includegraphics[height=10pt]{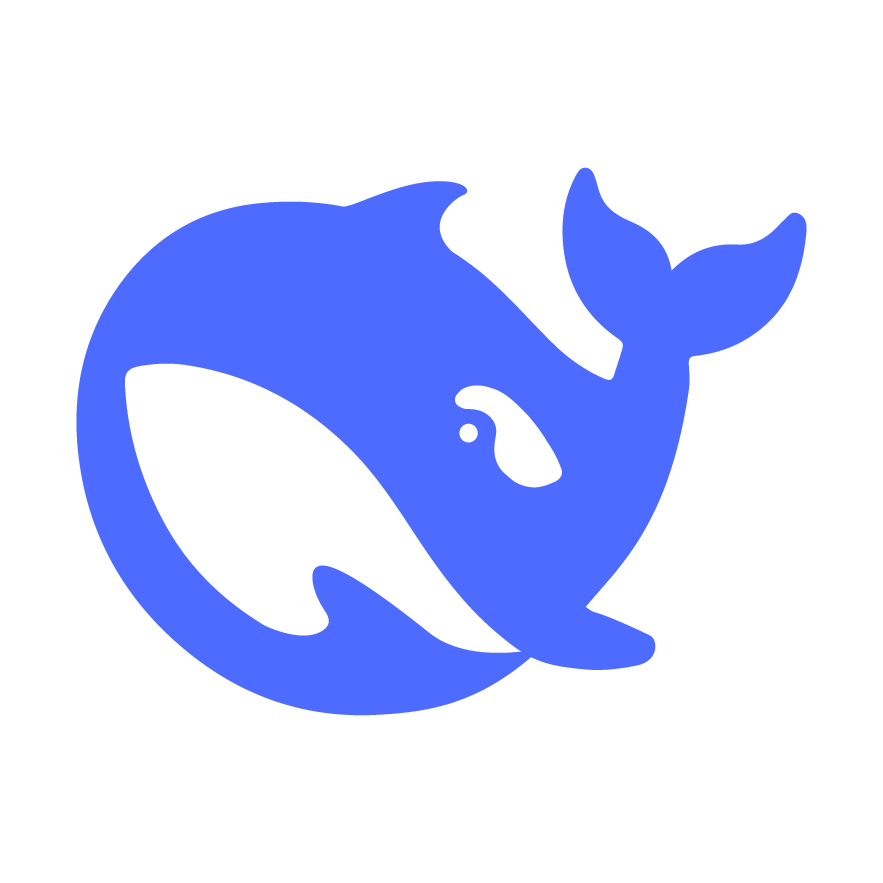} DeepSeek}
 & DeepSeek-R1       &    30.25    &    3.67   &  42.86    \\
\midrule
\multirow{5}{*}{\includegraphics[height=12pt]{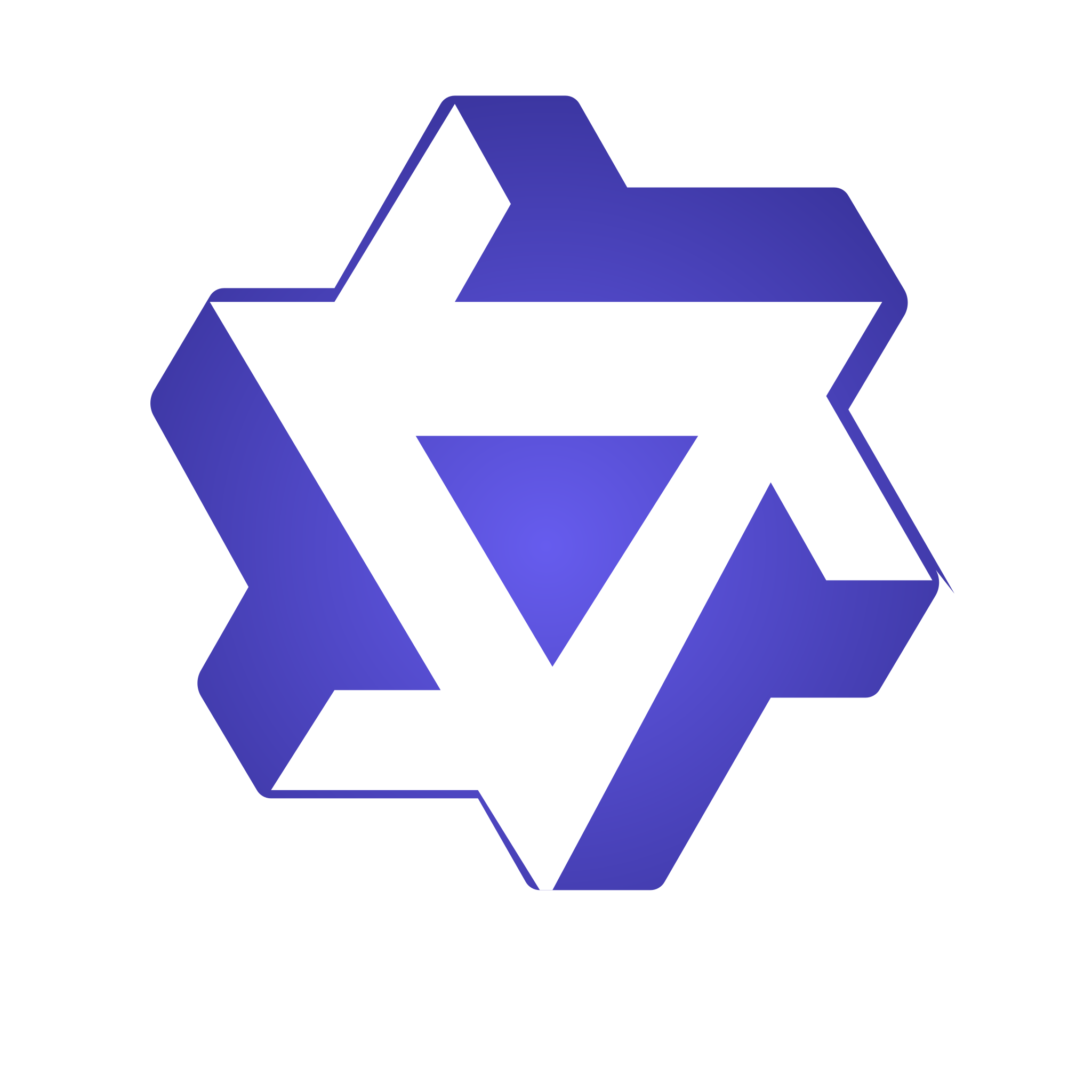} Alibaba}
 & Qwen2.5-Coder-7B-Instruct     & 1.59 & 2.73   &  30.56    \\
 & Qwen2.5-72B-Instruct    &   2.54     &  3.14   &   8.86    \\
 & Qwen3-8B                &   3.72     &   2.57  &   12.50    \\
 & Qwen3-32B               &   18.69     &   3.39  &  59.42   \\
 & Qwen3-30B-A3B-Thinking-2507 &   9.30     &   2.60    &   23.81   \\
 & Qwen3-Coder-30B-A3B-Instruct     &  6.06   &  2.90  &  32.81  \\
\midrule
\multirowcell{1}{\quad Ours} & \textbf{WebGen-R1-7B} & 29.21{\tiny\textcolor{ForestGreen}{+27.62}}  &  \textbf{3.94}{\tiny\textcolor{ForestGreen}{+44.32}} &  \textbf{95.89}{\tiny\textcolor{ForestGreen}{+65.33}} \\
\bottomrule
\end{tabular}
\label{tab:main_results}
\vspace{-2mm}
\end{table*}

\subsection{Main Results} 
We evaluate WebGen-R1 against state-of-the-art LLMs on WebGen-Bench using FSR, AAS, and VRR to examine whether reinforcement learning can improve functionality, design aesthetics, and rendering reliability in generated websites. 
As shown in Table \ref{tab:main_results}, WebGen-R1 achieves the best performance in design aesthetics and rendering reliability, with the highest AAS of 3.94 and the highest VRR of 95.89\%. 
In terms of functionality, WebGen-R1 attains an FSR of 29.21\%, improving by 27.62 percentage points over its base model. It outperforms six of the seven open-source baselines up to 72B and closely matches DeepSeek-R1-671B at 30.25\%. 
These results demonstrate that WebGen-R1 can transform a 7B base model from generating almost non-functional websites into producing deployable and aesthetically aligned multi-page websites.  
Notably, although the highest FSR is achieved by the much larger proprietary model Claude-3.7-Sonnet, this advantage in functionality does not translate into the best design aesthetics or rendering reliability.  
Moreover, we observe that design aesthetics appear easier to scale with model size than functional correctness. For example, large open-source models such as Qwen3-32B with AAS 3.39 and Qwen2.5-72B-Instruct with AAS 3.14 achieve AAS scores comparable to proprietary models, yet still lag far behind in FSR. 
This suggests that generating functionally correct websites remains substantially more challenging than producing visually appealing designs, because front-end functional requirements often involve precise logic and interaction handling beyond surface-level layout generation.  
Overall, our RL optimization tailored to project-level web application generation effectively encourages the LLM to produce functional and aesthetic multi-page websites in an end-to-end manner. 

\begin{table*}[t]
\centering
\setlength{\tabcolsep}{2pt}
\caption{
Category-wise performance comparison of WebGen-R1 and various LLMs from multiple institutions on the WebGen-Bench benchmark, evaluated by FSR. We also report the score improvement ($\pm$) of our WebGen-R1 relative to the base model, Qwen2.5-Coder-7B-Instruct. The first three columns correspond to categories of website-generation instructions, and the last three columns correspond to categories of test cases, following \citep{lu2025webgen}.
}
\resizebox{0.87\textwidth}{!}{
\begin{tabular}{l|l|cccccc}
\toprule
\textbf{Institution} & \textbf{Model} & 
\makecell{\textbf{Content} \\ \textbf{Presenta}-\\\textbf{tion}} &
\makecell{\textbf{User} \\ \textbf{Inter}-\\\textbf{action}} &
\makecell{\textbf{Data} \\ \textbf{Manage}-\\\textbf{ment}} &
\makecell{\textbf{Func}-\\\textbf{tional} \\ \textbf{Testing}} &
\makecell{\textbf{Data} \\ \textbf{Display} \\ \textbf{Testing}} &
\makecell{\textbf{Design} \\ \textbf{Validation} \\ \textbf{Testing}} \\
\midrule
\multirow{2}{*}{\includegraphics[height=8pt]{images/logos/openai.png} OpenAI}
 & GPT-5       & 55.29 & 46.21 & 36.81 & 36.93 & 53.14 & 62.73  \\
 & GPT-4.1     & 46.27 & 31.03 & 61.26 & 38.05 & 44.85 & 57.69    \\
\midrule
\multirow{2}{*}{\includegraphics[height=6pt]{images/logos/anthropic.png} Anthropic}
 & Claude-Sonnet-4   & 52.88 & 39.42 & 53.33 & 41.09 & 47.67 & 55.36  \\
 & Claude-3.7-Sonnet & 66.67 & 43.40 & 64.29 & 46.48 & 72.09 & 62.86 \\
\midrule
\multirow{1}{*}{\includegraphics[height=8pt]{images/logos/google.png} Google}
 & Gemini-2.5-Pro    & 31.54 & 37.34 & 39.71 & 29.60 & 43.05 & 44.90 \\
\midrule
\multirow{1}{*}{\includegraphics[height=10pt]{images/logos/deepseek.png} DeepSeek}
 & DeepSeek-R1       & 31.11 & 20.45 & 58.62 & 24.69 & 37.25 & 33.33  \\
\midrule
\multirow{3}{*}{\includegraphics[height=12pt]{images/logos/qwen.png} Alibaba}
 & Qwen2.5-Coder-7B-Instruct     & 0.00 & 3.17 & 3.12 & 0.00 & 3.57 & 0.00   \\
 & Qwen3-32B               & 24.81 & 17.02 & 14.41 & 11.42 & 22.73 & 32.47  \\
 & Qwen3-Coder-30B-A3B-Instruct     & 16.44 & 1.06 & 8.82 & 2.81 & 13.25 & 5.80 \\
\midrule
\multirowcell{1}{\quad Ours} & \textbf{WebGen-R1-7B} & 
35.29{\tiny\textcolor{ForestGreen}{+35.29}} & 
27.25{\tiny\textcolor{ForestGreen}{+24.08}} & 
26.88{\tiny\textcolor{ForestGreen}{+23.76}} & 
15.90{\tiny\textcolor{ForestGreen}{+15.90}} & 
30.43{\tiny\textcolor{ForestGreen}{+26.86}} & 
54.92{\tiny\textcolor{ForestGreen}{+54.92}} \\
\bottomrule
\end{tabular}}
\label{tab:fsr_results}
\end{table*}

\begin{figure}[t]
	\centering
    \includegraphics[width=0.9\linewidth]{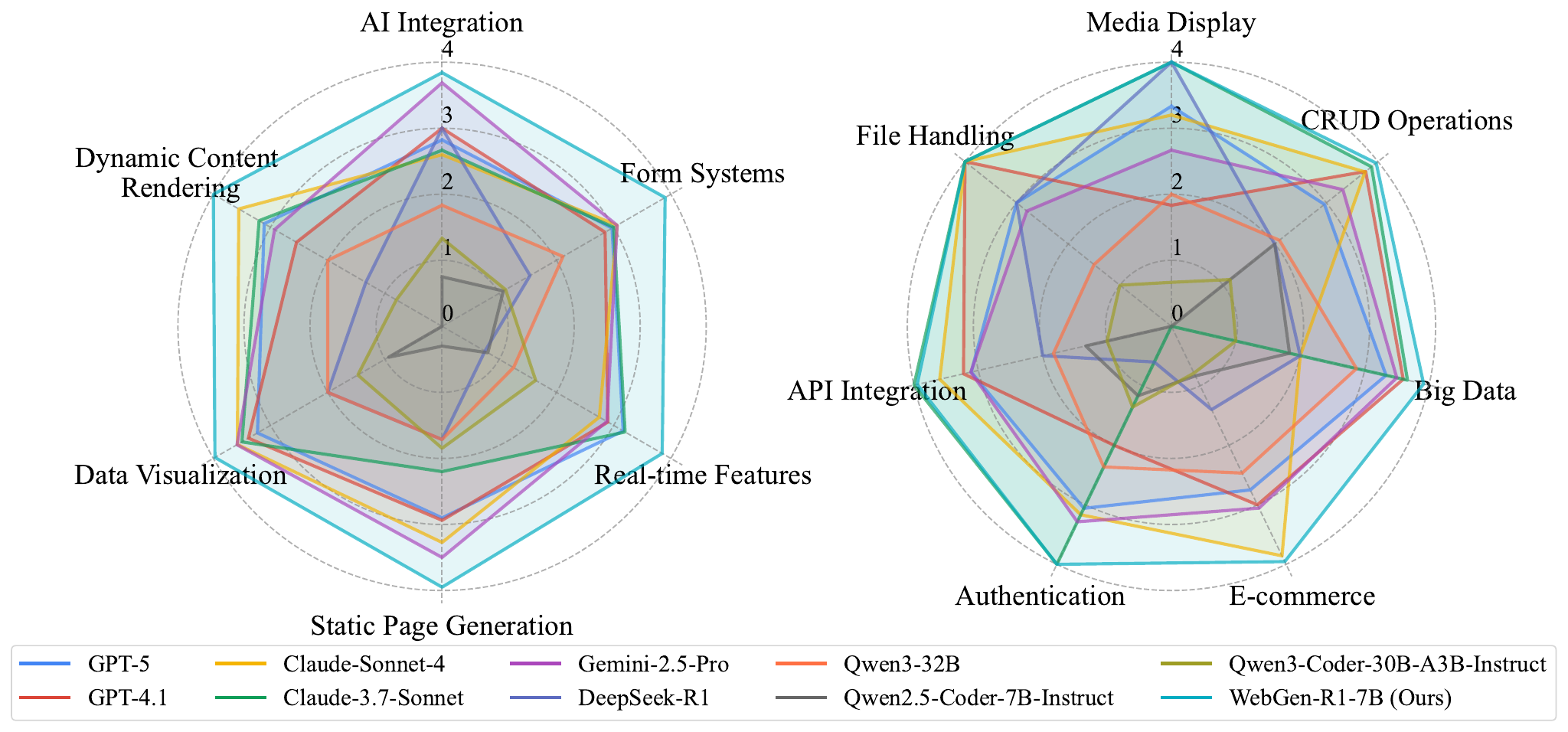}
    \caption{
    Comparison of WebGen-R1 and baseline LLMs across 13 multi-scenario front-end development tasks from WebGen-Bench, evaluating both functional correctness and visual fidelity.
    }
\label{fig:task_category}
\vspace{-5mm}
\end{figure}

\subsection{In-depth Analysis and Insights.}

\textbf{Multi-Scenario Web Environments.} 
We evaluate WebGen-R1 against various advanced LLMs on multi-scenario front-end development tasks from WebGen-Bench \citep{lu2025webgen}, including AI integration, form systems, real-time features, static page generation, data visualization, dynamic content rendering, media display, CRUD operations, big data, e-commerce, authentication, API integration, and file handling.   
As shown in Figure \ref{fig:task_category}, our WebGen-R1 achieves superior AAS across all 13 categories, indicating robust improvements in diverse front-end development scenarios.
Moreover, following \citep{lu2025webgen}, we report FSR for each category of website instructions and test cases in Table \ref{tab:fsr_results}. WebGen-R1 consistently achieves substantial gains over the base model across all categories. Notably, WebGen-R1 delivers competitive performance with much stronger baselines such as DeepSeek-R1 and Gemini-2.5-Pro in the Content Presentation and Design Validation Testing categories. 
We attribute these improvements to the proposed cascaded multimodal reward model, which combines structural guarantees, execution-grounded functional feedback, and vision-based aesthetic supervision, and aligns optimization with human standards for both functional correctness and visual design. 
The consistent gains across diverse front-end scenarios indicate that WebGen-R1 is robust across tasks and domains and can better balance functional correctness with aesthetic alignment, which highlights its practical value for real-world web development.

\begin{figure}[tp]
\centering
\begin{minipage}[b]{0.49\linewidth}
\centering 
\includegraphics[width=\textwidth]{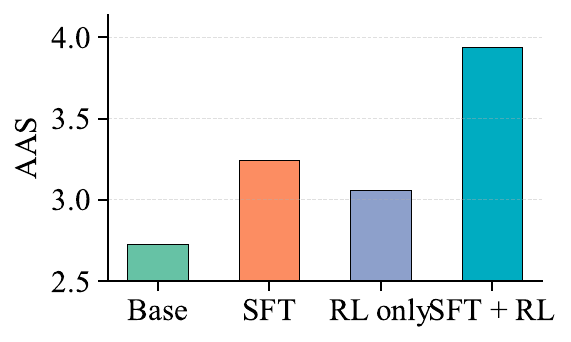}
\caption{Performance comparison of SFT, RL-only, and SFT+RL on the WebGen-Bench benchmark under the AAS metric. 
}
\label{fig:sft_rl} 
\end{minipage}
\hspace{0.2em}
\begin{minipage}[b]{0.49\linewidth} 
\centering 
\includegraphics[width=\textwidth]{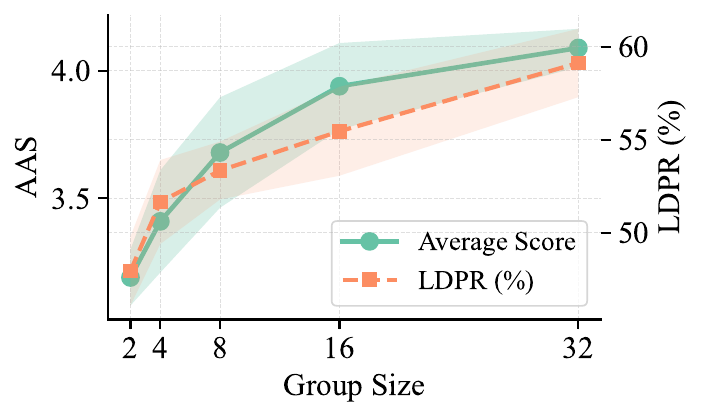}
\caption{Impact of group size $G$ in GRPO on WebGen-Bench performance, measured by AAS and LDPR metrics.
}
\label{fig:group_size} 
\end{minipage}
\end{figure}

\begin{table}[t]
\centering
\begin{minipage}{0.48\textwidth}
\caption{Performance comparison of SFT, RL-only, and SFT+RL on the WebGen-Bench benchmark under the FSR and VRR metrics.}
\label{tab:fsr_vrr_1}
\centering
\resizebox{\textwidth}{!}{
\begin{tabular}{lcccc}
\toprule
\textbf{Metrics} & \textbf{Base} & \textbf{SFT} & \textbf{RL only} & \textbf{SFT + RL} \\
\midrule
FSR & 1.59 & 20.08 & 18.23 & 29.21 \\
VRR & 30.56 & 30.69 & 26.82 &  95.89 \\
\bottomrule
\end{tabular}}
\end{minipage}
\quad
\begin{minipage}{0.48\textwidth}
\caption{Impact of group size $G$ in GRPO on WebGen-Bench performance, measured by FSR and VRR metrics.}
\label{tab:fsr_vrr_2}
\centering
\resizebox{\textwidth}{!}{
\begin{tabular}{lccccc}
\toprule
\textbf{Metrics} & \textbf{2} & \textbf{4} & \textbf{8} & \textbf{16} & \textbf{32} \\
\midrule
FSR & 22.39 & 22.92 & 24.99 & 29.21 & 34.79 \\
VRR & 90.98 & 91.07 & 93.52 & 95.89 & 98.27\\
\bottomrule
\end{tabular}}
\end{minipage}
\end{table}

\textbf{Fine-Tuning Strategy.} 
We compare RL and supervised fine-tuning on the website generation task. As shown in Figure~\ref{fig:sft_rl}, fine-tuning the base model Qwen2.5-Coder-7B-Instruct on 600 GPT-4.1-generated examples yields an 18.68\% improvement in AAS, substantially improving structural correctness and semantic consistency.
The RL-only, also known as R1-Zero \citep{guo2025deepseek}, attains a 12.09\% improvement, which demonstrates the effectiveness of reward signals for optimizing both functionality and visual appearance. 
While SFT alone outperforms RL alone, combining SFT initialization with RL leads to even stronger results and surpasses SFT-only and RL-only by 21.60\% and 28.76\%, respectively.
A similar trend is observed for FSR and VRR, as shown in Table \ref{tab:fsr_vrr_1}.
These results suggest that SFT provides the model with a strong structural and semantic prior for website generation, while RL improves exploration under reward guidance and enables the model to generate websites that are higher quality, more functional, and more visually appealing.

\textbf{Group Size in GRPO.}  
We study the impact of group size $G$ in GRPO, where $G \in \{2, 4, 8, 16, 32\}$, while keeping all other hyperparameters fixed.    
As shown in Figure~\ref{fig:group_size} and Table \ref{tab:fsr_vrr_2}, increasing the group size leads to consistent gains in AAS, LDPR, FSR, and VRR. We hypothesize that these improvements come from stronger exploration with larger groups, which increases the diversity of candidate trajectories and improves the likelihood of discovering high-quality websites.

\begin{figure}[tp]
\centering
\begin{minipage}[b]{0.49\linewidth}
\centering 
\includegraphics[width=\textwidth]{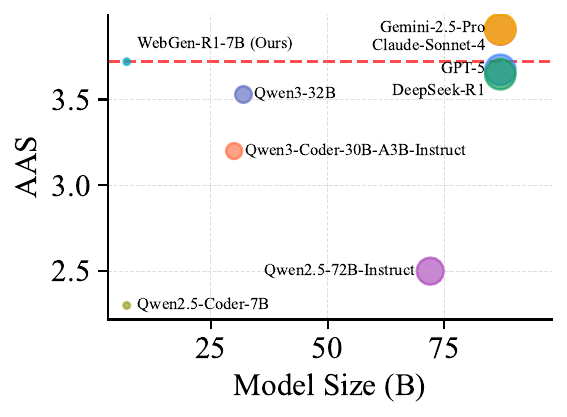}
\caption{Performance of WebGen‑R1 on the WebDev Arena benchmark across different domains and prompt distributions.
}
\label{fig:webdev_arena_bubble} 
\end{minipage}
\hspace{0.2em}
\begin{minipage}[b]{0.49\linewidth} 
\centering 
\includegraphics[width=\textwidth]{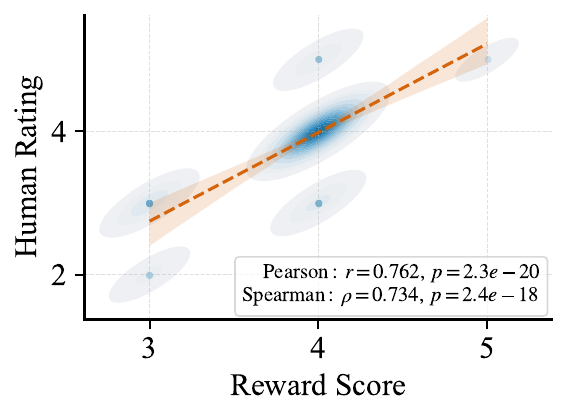}
\caption{Alignment between reward model evaluations and human ratings for websites generated from WebGen-Bench, showing strong correlation.
}
\label{fig:llm_human_correlation} 
\end{minipage}
\end{figure}

\begin{figure*}[t]
\centering
\hspace{0.2em}
\includegraphics[width=\linewidth]{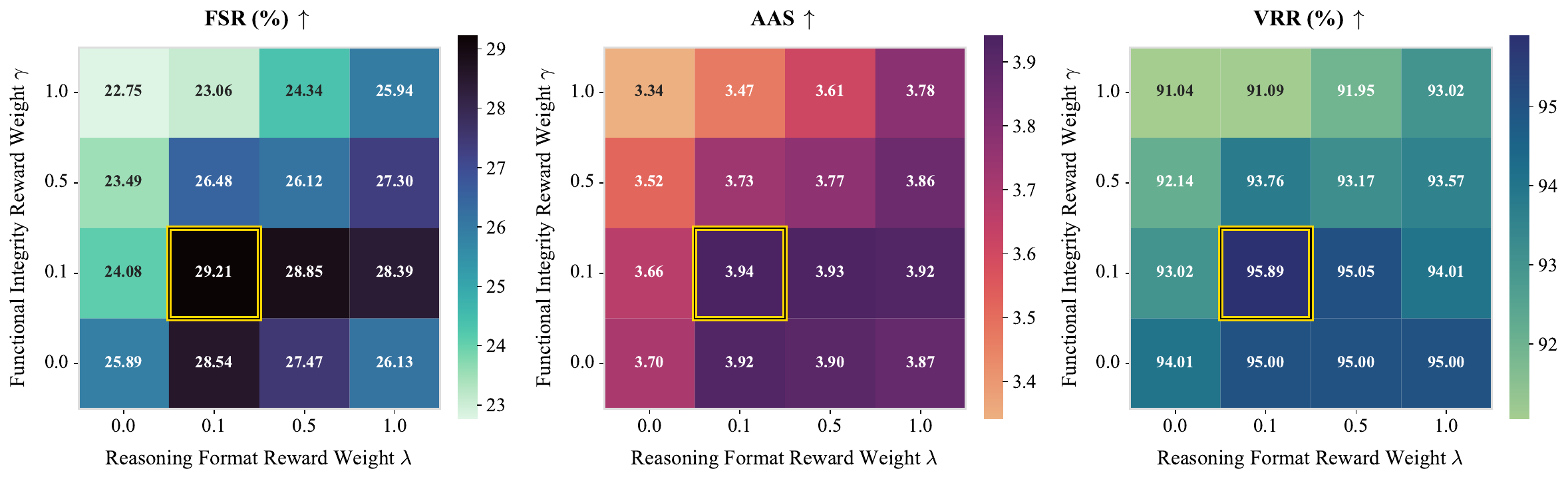} 
\caption{
Impact of $\gamma$ and $\lambda$ on website generation performance. The heatmaps show the parameter landscape under FSR, AAS, and VRR metrics. The gold box marks the optimal setting ($\gamma=0.1, \lambda=0.1$), which yields the best overall performance.
}
\label{fig:gamma_lambda_sensitivity} 
\vspace{-2mm}
\end{figure*}

\textbf{Generalization and Robustness.}
We further study whether WebGen-R1 can generalize its reasoning ability and visual design capability to settings where both domain and prompt distributions differ substantially from those seen during RL training. 
To this end, we use the curated WebDev Arena benchmark, which contains instruction distributions and task categories that are not covered in our training set. 
We do not report FSR on this benchmark because WebDev Arena does not provide standardized test cases, which prevents the GUI agent from verifying functional correctness.
As shown in Figure \ref{fig:webdev_arena_bubble}, WebGen-R1 consistently outperforms a range of state-of-the-art proprietary and open-source baselines, such as DeepSeek-R1, GPT-5, and Qwen3-32B, on AAS.  
This suggests that WebGen-R1 has learned architecture-level and style-level abstractions that remain effective on unseen web tasks. These findings indicate that WebGen-R1 maintains strong robustness and practical utility in real-world deployment settings where specifications evolve over time.  

\textbf{Reward Scalarization Weights.} 
We study the impact of the reward scalarization weights $\gamma$ and $\lambda$. Here, $\gamma$ controls the functional integrity reward $s_{\text{func}}$, and $\lambda$ controls the reasoning format reward $s_{\text{cot}}$. 
Because $s_{\text{func}}$ and $s_{\text{cot}}$ are sparse binary signals, while $s_{\text{vis}}$ provides a dense continuous signal, balancing their contributions is important to avoid reward hacking and policy collapse.  
We evaluate configurations of $(\gamma, \lambda)$ from the grid $\{0, 0.1, 0.5, 1.0\}^2$. All models are trained with the same hyperparameters and evaluated on FSR, AAS, and VRR. 
As shown in Figure~\ref{fig:gamma_lambda_sensitivity}, smaller $\gamma$ values generally lead to higher AAS and VRR, and are often accompanied by better FSR. In contrast, large $\gamma$ values tend to reduce FSR when $\lambda$ is small, with a slight recovery as $\lambda$ increases. Mid-range $\lambda$ values provide moderate gains in AAS and VRR across settings.  
These trends highlight an important trade-off among functional correctness, aesthetic quality, and reasoning consistency. Overweighting sparse structural rewards can hurt visual fidelity, while relying only on visual rewards may under-constrain generation logic. Our analysis suggests that balanced scalarization leads to a better overall trade-off for robust website generation. 

\textbf{Human Alignment Study.}
We examine whether our reward design accurately reflects human preferences for functionality and aesthetics. Since RL depends entirely on this reward, misalignment could lead to outputs that fail to meet user expectations. 
To assess alignment, we compare the reward model with human judgments by asking three experienced front-end developers to independently rate 101 websites generated from WebGen-Bench in terms of functionality and visual appeal. We aggregate the human scores and compare them with the outputs of the reward model. As shown in Figure \ref{fig:llm_human_correlation}, we observe strong correlations with Pearson $ r = 0.762 $ and Spearman $ \rho = 0.734 $, which indicates close alignment between the reward model and human preferences. This result shows that our reward design can reliably assess both functional fidelity and aesthetic appeal in generated websites.

\begin{figure*}[t]
\centering
\setlength{\tabcolsep}{1pt}
\begin{tabular}{@{}p{\textwidth}@{}} 
\begin{tabular}{@{}p{0.25\textwidth} p{0.25\textwidth} p{0.25\textwidth} p{0.25\textwidth}@{}}
    \begin{minipage}{\linewidth}
        \centering
        \small \textbf{GPT-5} \\[0.1em]
        \includegraphics[width=\linewidth]{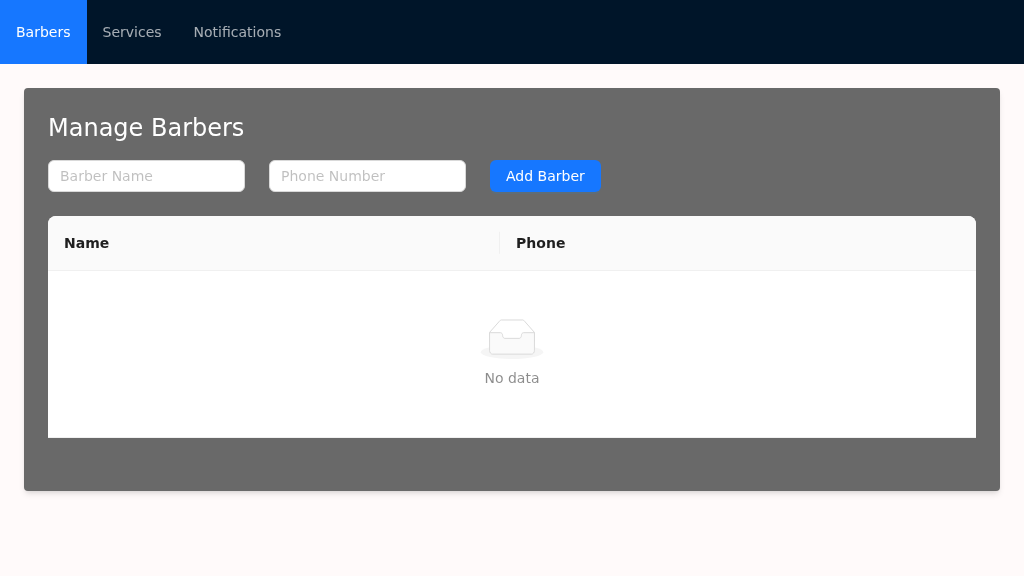}  
    \end{minipage} &
    \begin{minipage}{\linewidth}
        \centering
        \small \textbf{Claude-Sonnet-4} \\[0.1em]
        \includegraphics[width=\linewidth]{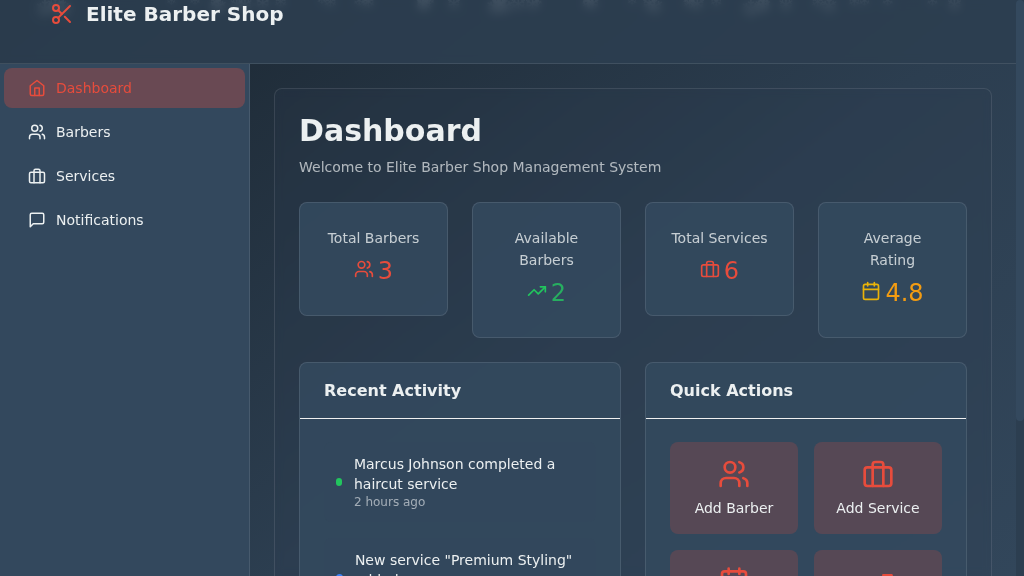} 
    \end{minipage} &
    \begin{minipage}{\linewidth}
        \centering
        \small \textbf{Gemini-2.5-Pro} \\[0.1em] 
        \includegraphics[width=\linewidth]{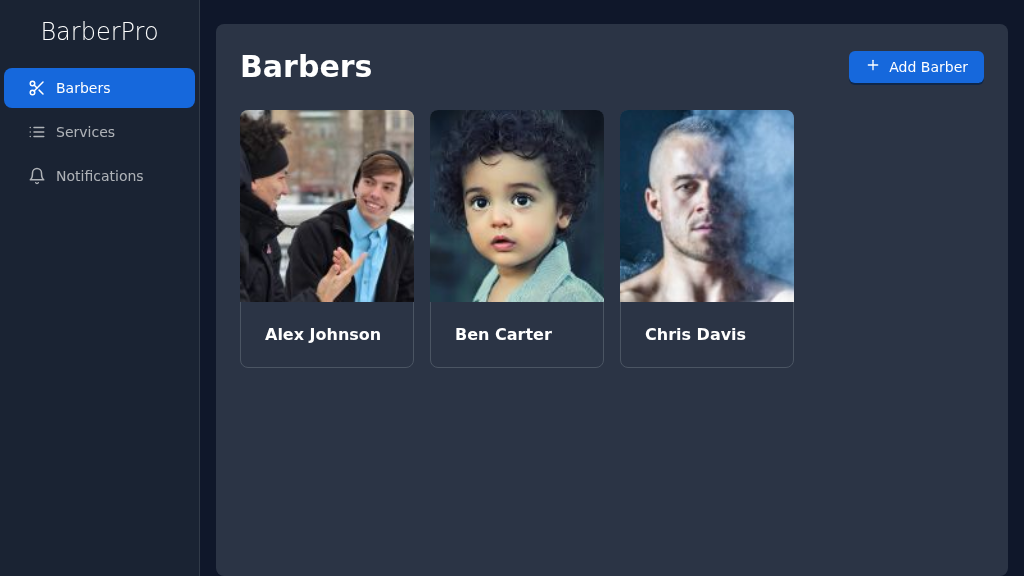} 
    \end{minipage} &
    \begin{minipage}{\linewidth}
        \centering
        \small \textbf{WebGen-R1-7B (Ours)} \\[0.1em] 
        \includegraphics[width=\linewidth]{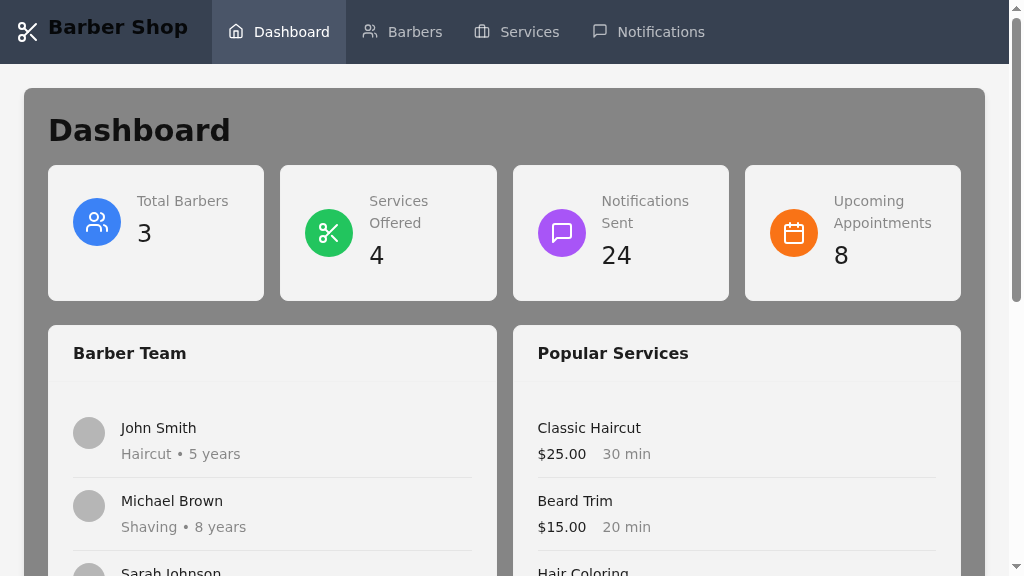} 
    \end{minipage}
\end{tabular}\\
\begin{tabular}{@{}p{0.25\textwidth} p{0.25\textwidth} p{0.25\textwidth} p{0.25\textwidth}@{}}
    \begin{minipage}{\linewidth}
        \centering
        \includegraphics[width=\linewidth]{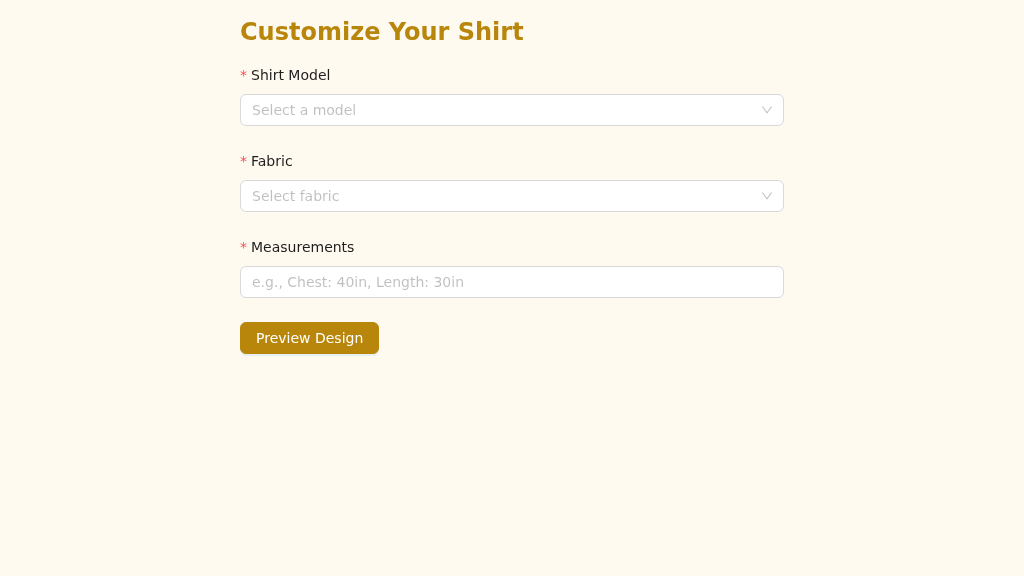}  
    \end{minipage} &
    \begin{minipage}{\linewidth}
        \centering
        \includegraphics[width=\linewidth]{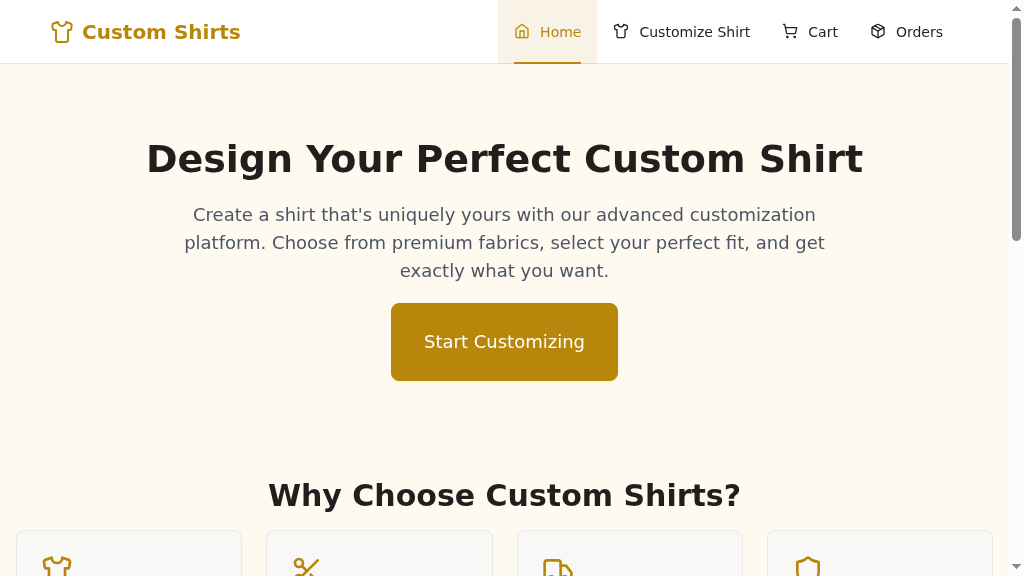} 
    \end{minipage} &
    \begin{minipage}{\linewidth}
        \centering 
        \includegraphics[width=\linewidth]{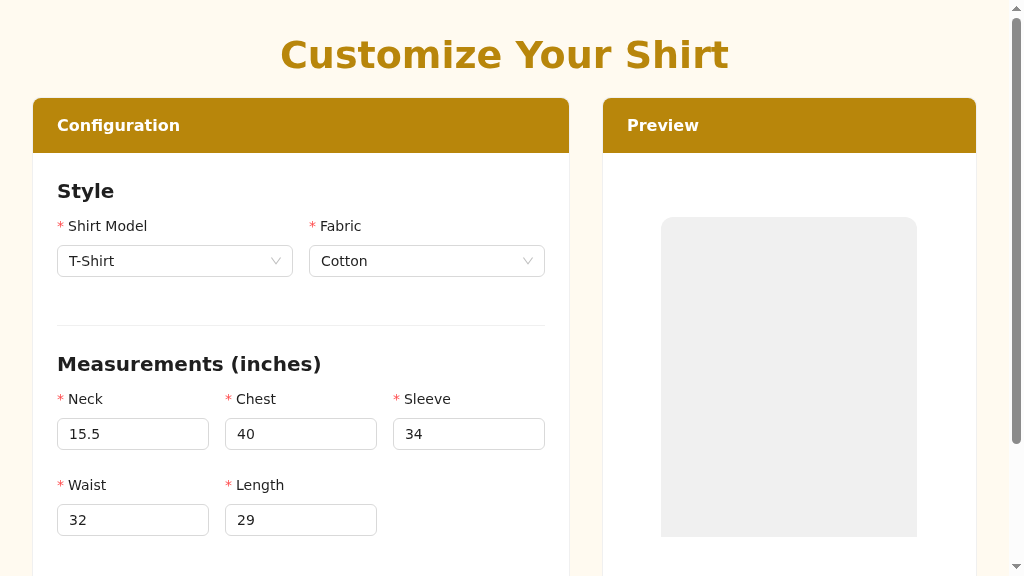} 
    \end{minipage} &
    \begin{minipage}{\linewidth}
        \centering
        \includegraphics[width=\linewidth]{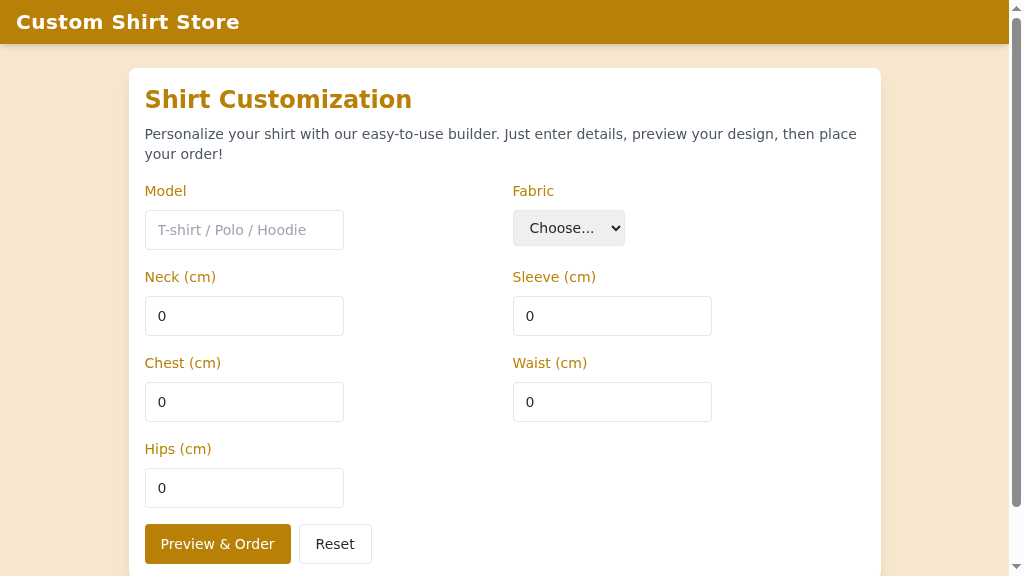} 
    \end{minipage}
\end{tabular}\\
\begin{tabular}{@{}p{0.25\textwidth} p{0.25\textwidth} p{0.25\textwidth} p{0.25\textwidth}@{}}
    \begin{minipage}{\linewidth}
        \centering
        \includegraphics[width=\linewidth]{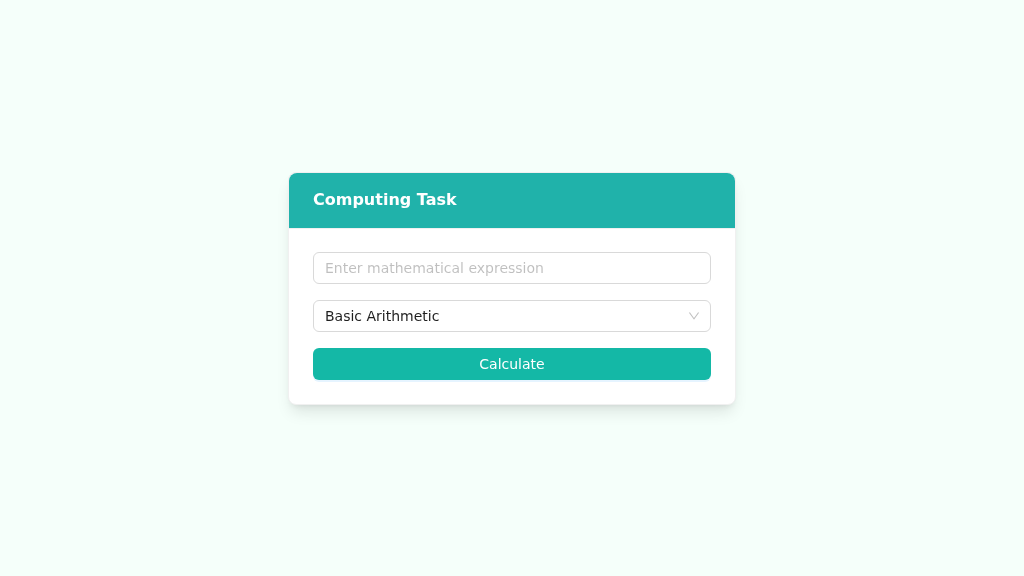} 
    \end{minipage} &
    \begin{minipage}{\linewidth}
        \centering
        \includegraphics[width=\linewidth]{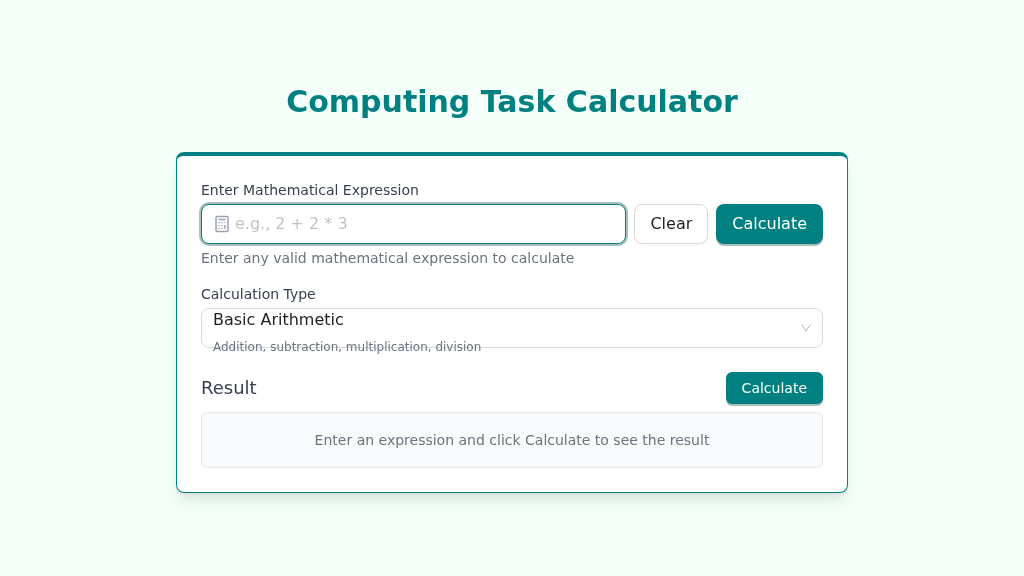} 
    \end{minipage} &
    \begin{minipage}{\linewidth}
        \centering
        \includegraphics[width=\linewidth]{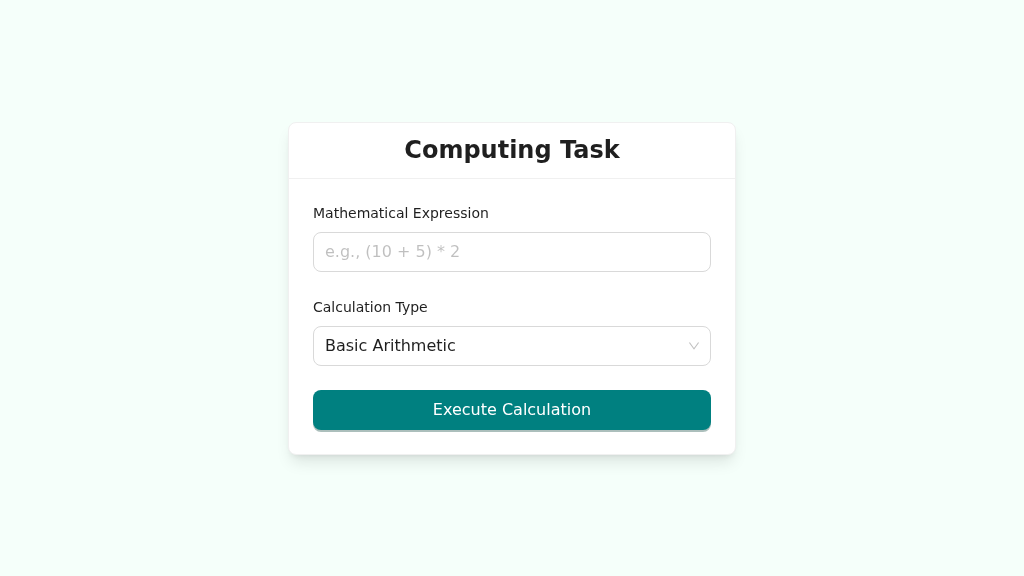} 
    \end{minipage} &
    \begin{minipage}{\linewidth}
        \centering
        \includegraphics[width=\linewidth]{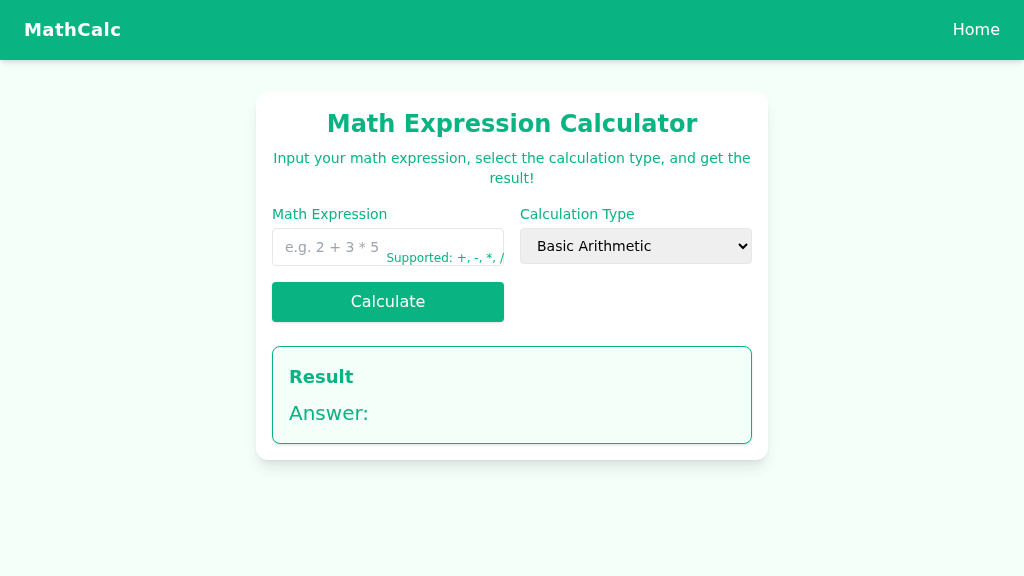}
    \end{minipage}
\end{tabular} \\ 
\midrule
\begin{tabular}{@{}p{0.25\textwidth} p{0.25\textwidth} p{0.25\textwidth} p{0.25\textwidth}@{}}
    \begin{minipage}{\linewidth}
        \centering
        \includegraphics[width=\linewidth]{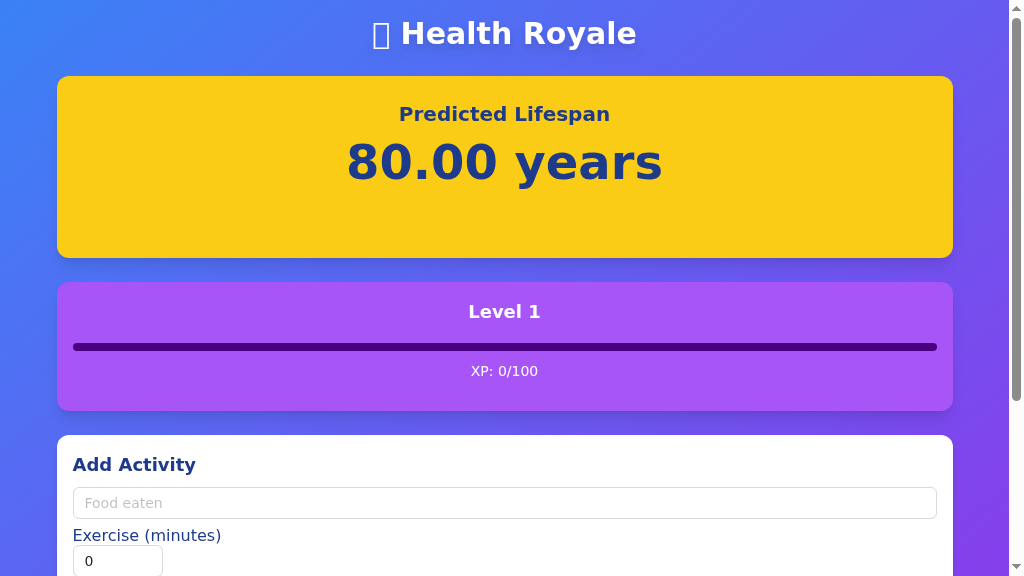} 
    \end{minipage} &
    \begin{minipage}{\linewidth}
        \centering
        \includegraphics[width=\linewidth]{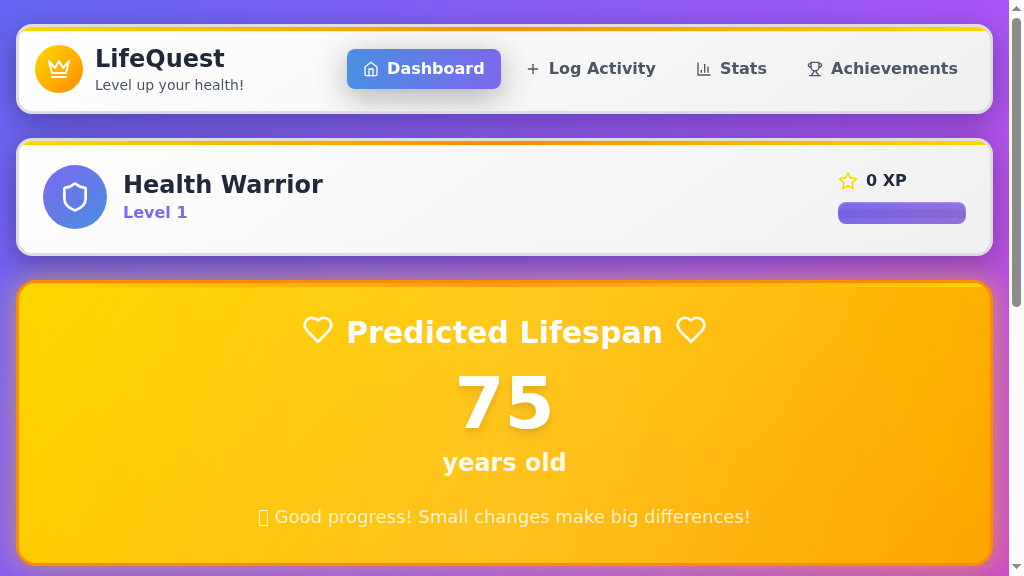} 
    \end{minipage} &
    \begin{minipage}{\linewidth}
        \centering
        \includegraphics[width=\linewidth]{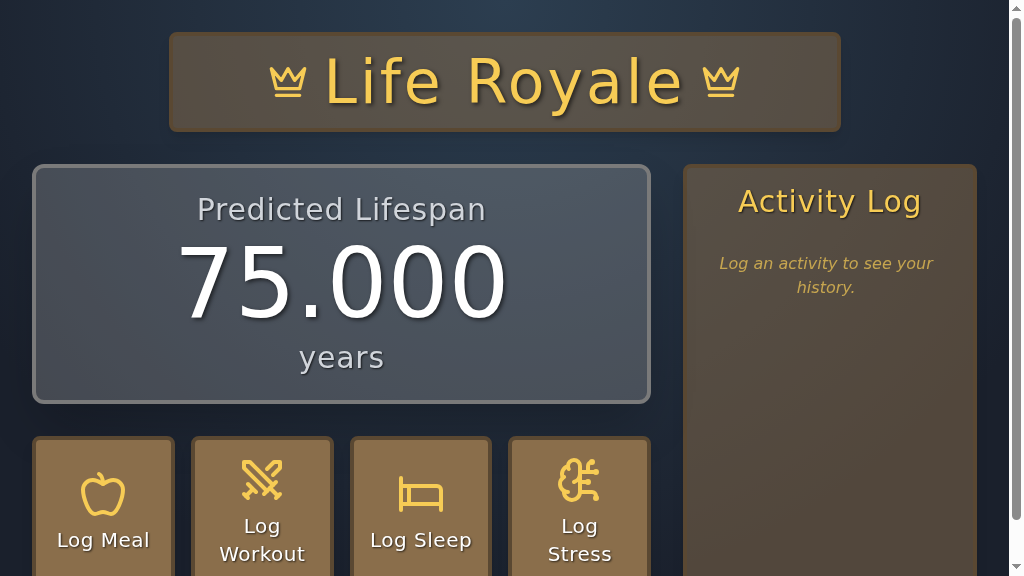} 
    \end{minipage} &
    \begin{minipage}{\linewidth}
        \centering
        \includegraphics[width=\linewidth]{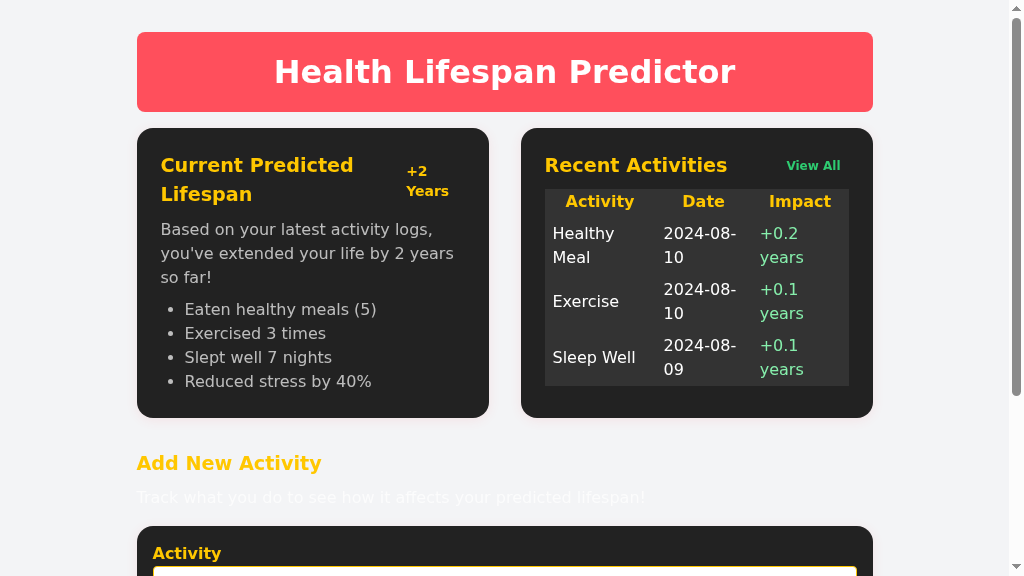}
    \end{minipage}
\end{tabular}\\
\begin{tabular}{@{}p{0.25\textwidth} p{0.25\textwidth} p{0.25\textwidth} p{0.25\textwidth}@{}}
    \begin{minipage}{\linewidth}
        \centering
        \includegraphics[width=\linewidth]{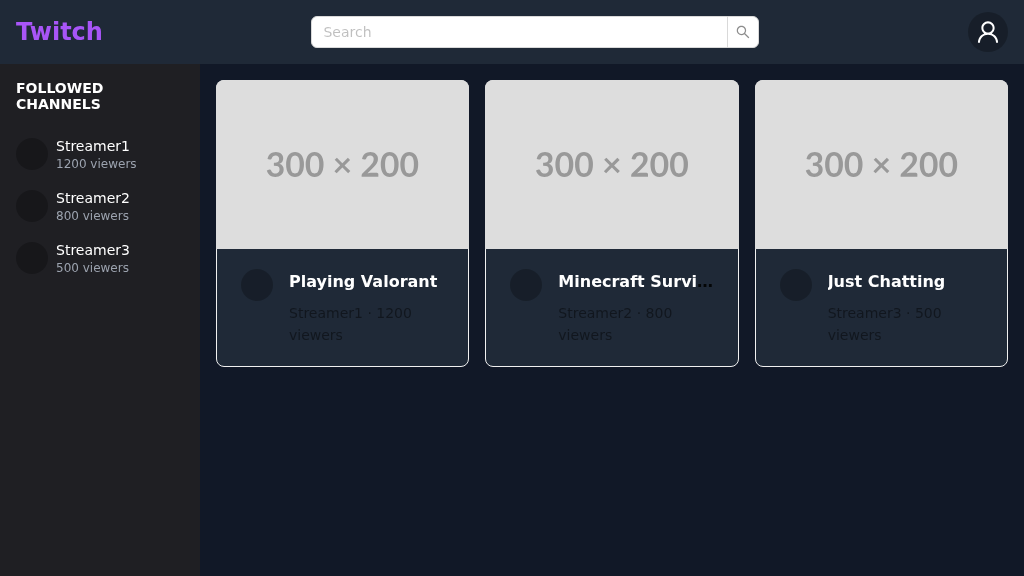} 
    \end{minipage} &
    \begin{minipage}{\linewidth}
        \centering
        \includegraphics[width=\linewidth]{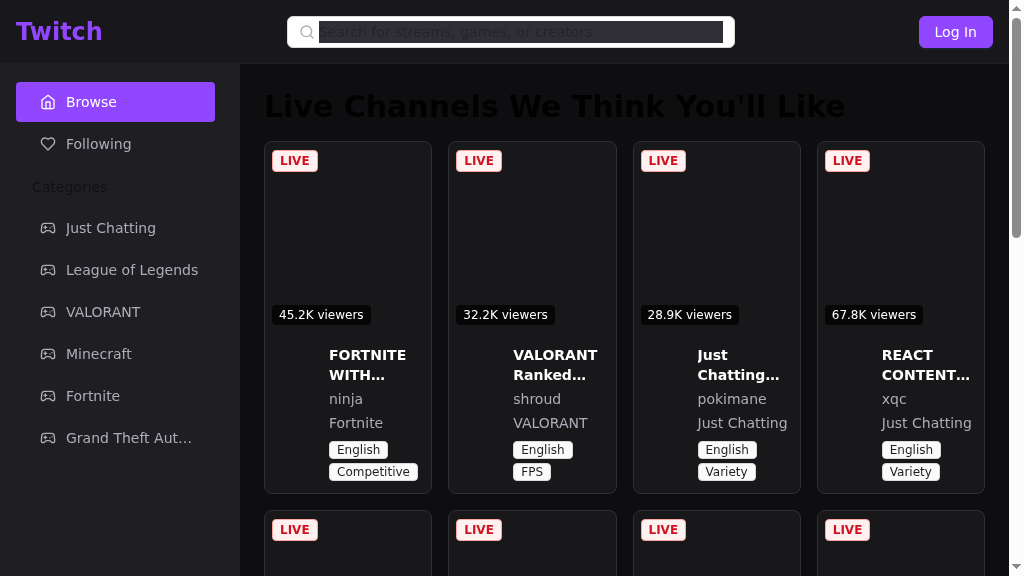} 
    \end{minipage} &
    \begin{minipage}{\linewidth}
        \centering
        \includegraphics[width=\linewidth]{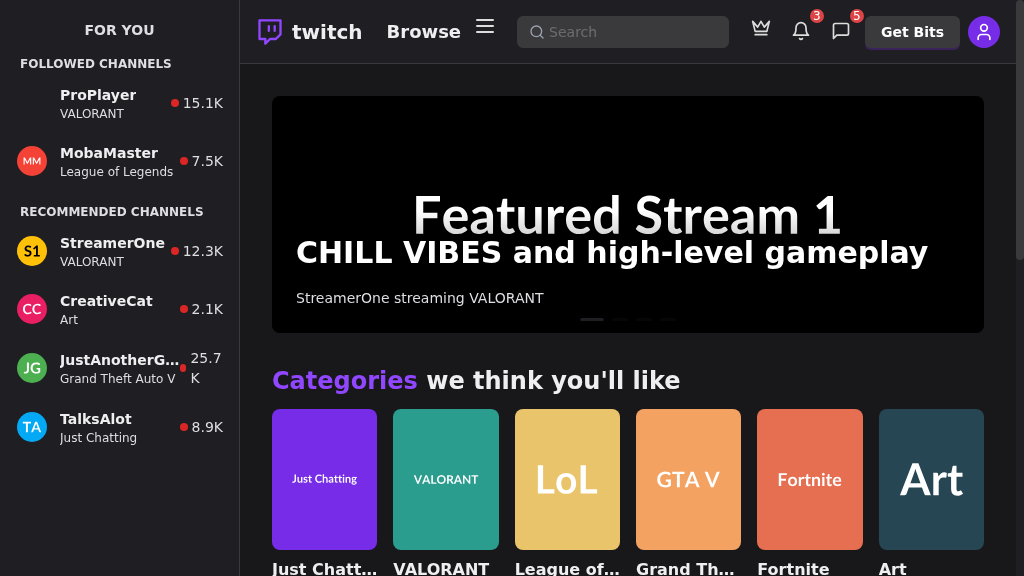} 
    \end{minipage} &
    \begin{minipage}{\linewidth}
        \centering
        \includegraphics[width=\linewidth]{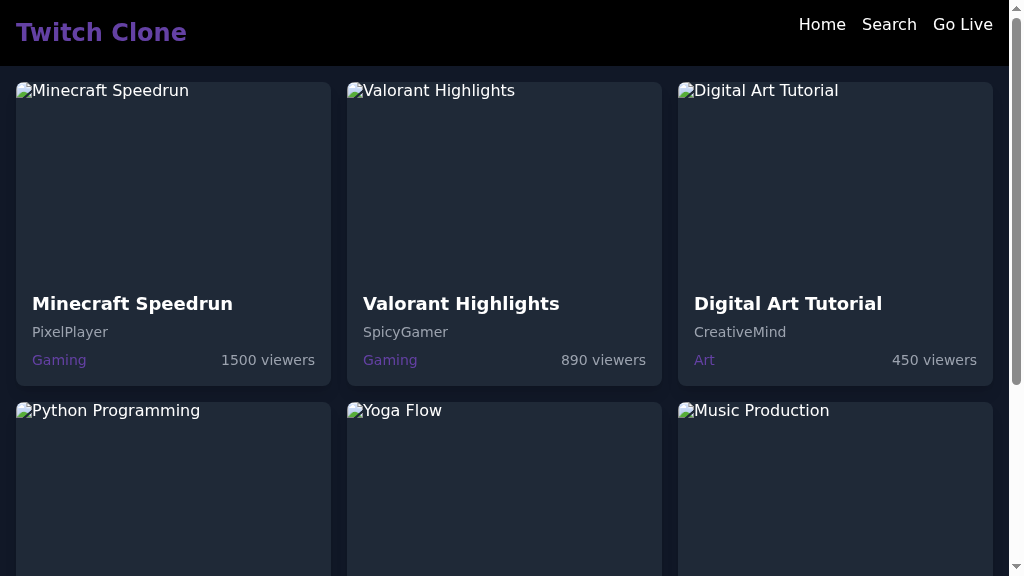}
    \end{minipage}
\end{tabular}\\
\begin{tabular}{@{}p{0.25\textwidth} p{0.25\textwidth} p{0.25\textwidth} p{0.25\textwidth}@{}}
    \begin{minipage}{\linewidth}
        \centering
        \includegraphics[width=\linewidth]{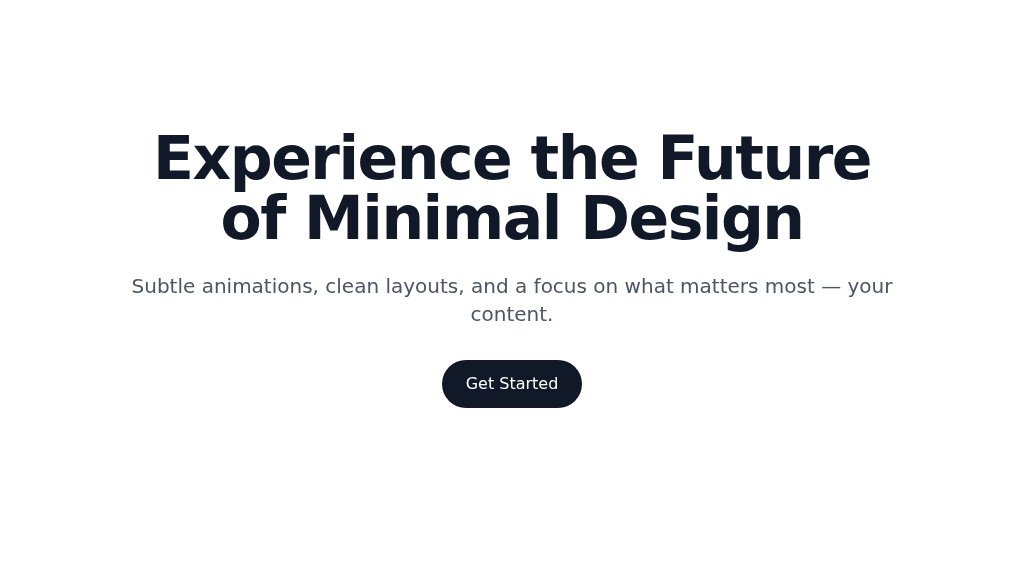}  
    \end{minipage} &
    \begin{minipage}{\linewidth}
        \centering
        \includegraphics[width=\linewidth]{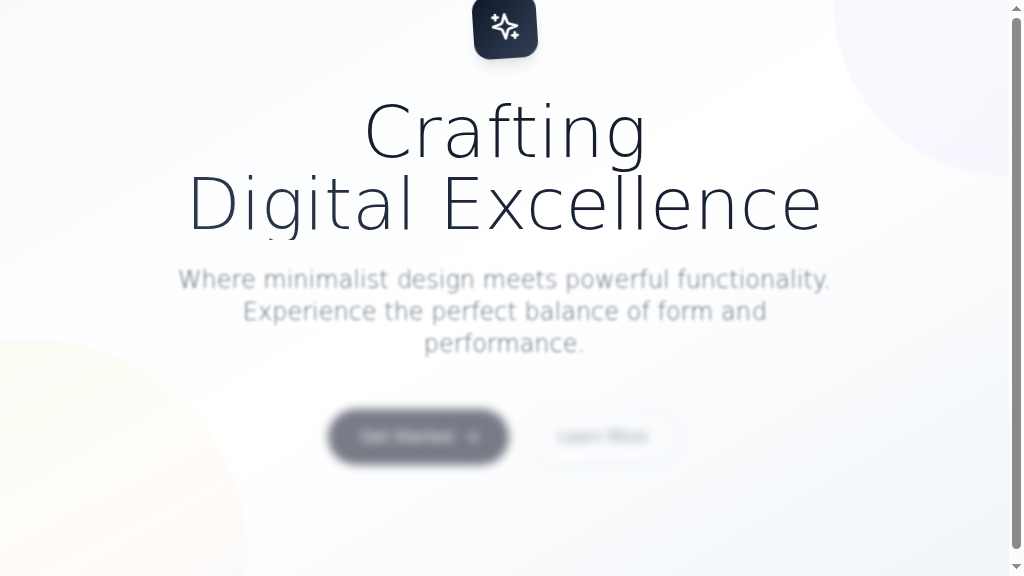} 
    \end{minipage} &
    \begin{minipage}{\linewidth}
        \centering
        \includegraphics[width=\linewidth]{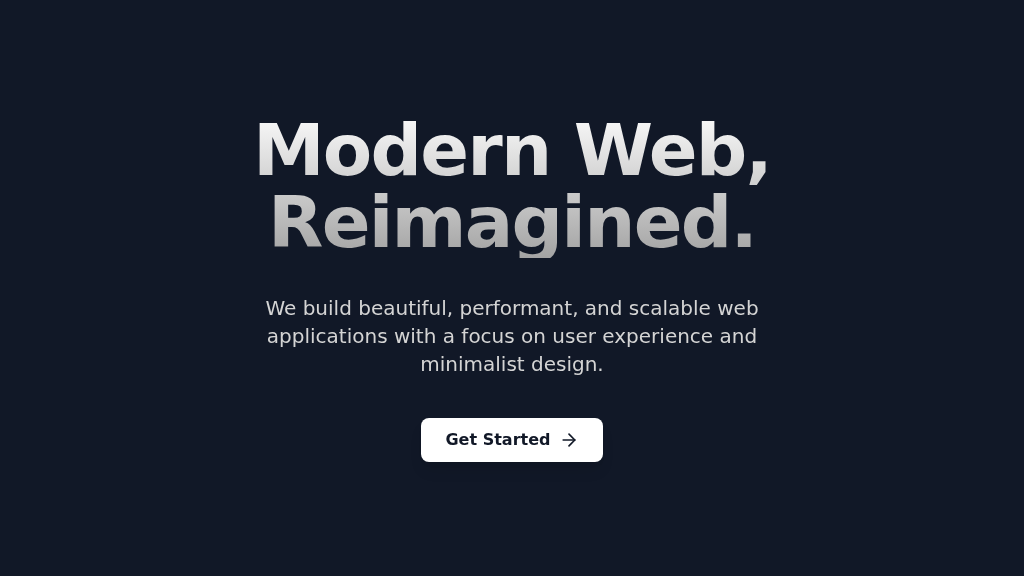} 
    \end{minipage} &
    \begin{minipage}{\linewidth}
        \centering
        \includegraphics[width=\linewidth]{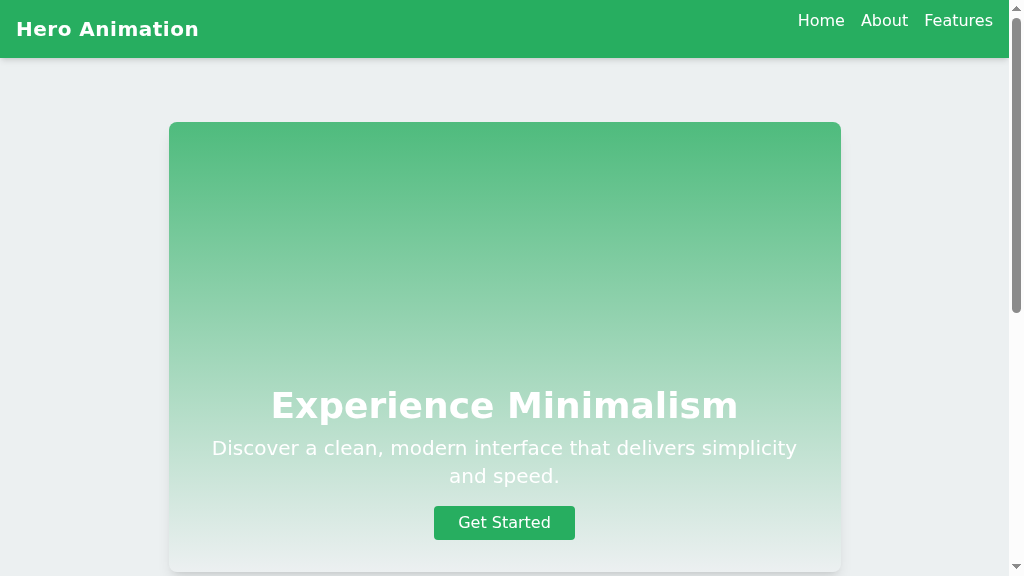} 
    \end{minipage}
\end{tabular}\\
\end{tabular}
\caption{Case study comparing WebGen-R1-7B with three strong baselines. The top three rows show in-distribution examples from WebGen-Bench, whereas the bottom three rows show out-of-distribution examples from WebDev Arena.}
\label{fig:case_study}
\end{figure*}

\subsection{Case Studies}\label{sec:case_study_main}
To qualitatively evaluate WebGen-R1's ability to improve website functionality and aesthetic quality, we conduct case studies using user instructions from the in-distribution WebGen-Bench and the out-of-distribution WebDev-Arena, as shown in Figure \ref{fig:case_study}.  
Compared with strong baselines, WebGen-R1 uses substantially fewer parameters while achieving competitive performance, and it generates websites with organized layouts, coherent and visually appealing designs, and responsive behaviors that closely match the detailed instructions in WebGen-Bench. 
In contrast, WebDev-Arena provides much less detailed instructions, which makes WebGen-R1 rely more on its world knowledge and occasionally miss newer design trends. We plan to mitigate this limitation by adopting newer base models in future work. 
These results suggest that LLMs trained with functionally and aesthetically grounded RL can effectively capture both engineering requirements and design principles in web development. 
Additional case studies are provided in Appendix \ref{ap:case_study}. 
\section{Conclusion}
In this work, we introduce WebGen‑R1, a reinforcement learning framework that enables small open-source LLMs to generate entire multi‑page websites in an end‑to‑end manner while meeting both functional and aesthetic requirements.  
We propose a scaffold-driven structured generation paradigm with a hierarchical verification and rendering pipeline, ensuring high functional reliability without the prohibitive cost of GUI-agent exploration.  
We then design a cascaded multimodal reward model that seamlessly couples structural guarantees with execution-grounded functional feedback and vision-based aesthetic supervision, aligning the optimization objectives with human standards for both functional correctness and visual design.  
Extensive experiments on two real‑world benchmarks demonstrate that our WebGen‑R1 consistently improves functional robustness, visual coherence, and deployability, surpassing or matching advanced proprietary and open‑source models.  
We believe these insights open new directions for end‑to‑end RL training in full‑stack application development, and we release all resources to support future research in this emerging area.

\bibliography{ref}
\bibliographystyle{plain}


\newpage
\appendix

\begin{table}[t]
\centering 
\caption{Statistics of the WebGen-Bench and WebDev Arena benchmarks, including the number of samples, instruction length statistics in tokens measured by tiktoken's cl100k\_base tokenizer, the number of test cases, web development categories, and an example task.}
\label{tab:dataset_statistics}
\setlength{\tabcolsep}{4pt}       
\resizebox{\textwidth}{!}{
\begin{tabular}{lcccccccm{5cm}m{7cm}} 
\toprule
\multirow{2}{*}{\textbf{Benchmark}} & \multirow{2}{*}{\textbf{Samples}} & 
\multicolumn{4}{c}{\textbf{Instruction Length}} & 
\multirow{2}{*}{\textbf{\# Test Cases}} & 
\multirow{2}{*}{\textbf{Category}} & 
\multirow{2}{*}{\textbf{Examples}} \\
\cmidrule(lr){3-6}
 & & {\# Min} & {\# Median} & {\# Max} & {\# Avg.} & & & \\
\midrule
WebGen-Bench 
& 101 & 52 & 84 & 135 & 86.06 & 647 & 
\makecell[{{p{4cm}}}]{%
Static Page Generation\\
Dynamic Content Rendering\\
Data Visualization\\
Media Display\\
Form Systems\\
Authentication\\
Real-time Features\\
E-commerce\\
AI Integration\\
CRUD Operations\\
API Integration\\
Big Data\\
File Handling} & 
\textit{Please develop a web-based Texas Hold'em poker game with features such as game lobby, table games, and chat functionality. Users should be able to create or join game rooms, play Texas Hold'em, view game records, and manage their account information. The game lobby should display available game rooms, current game status, and player information. The table game should display player hand cards, community cards, betting information, and action buttons. Implement azure for the page background and midnight blue for the elements.}\\
\midrule
WebDev Arena 
& 119 & 3 & 20 & 119 & 23.13 & 0 &
\makecell[{{p{4cm}}}]{
Website Design\\
Game Development\\
Clone Development\\
App Development\\
Web Development\\
UI Design\\
Digital Tools\\
App Design\\
AI Applications\\
Simulations\\
Creative Humor} &
\textit{Make me a clone of WhatsApp Chat App.} \\
\bottomrule
\end{tabular}
}
\end{table}

\section{Dataset Statistics and Analysis}\label{sec:data_analysis}
As shown in Table~\ref{tab:dataset_statistics}, both WebGen-Bench \citep{lu2025webgen} and WebDev Arena \citep{webdev_arena} are highly open-ended benchmarks for web generation, yet they differ substantially in instruction length distributions and coverage of web development categories.

Specifically, WebGen-Bench comprises 101 samples with moderately long natural language instructions (median 84 tokens, mean 86.06, max 135) and 647 executable test cases. It covers 13 heterogeneous front-end development scenarios, ranging from static and dynamic rendering to AI integration and big data handling. This setup suggests tasks with rich functional requirements and diverse constraints, requiring the model to interpret specifications that combine precise functional logic with explicit visual styling instructions.

In contrast, WebDev Arena contains 119 carefully selected tasks with substantially shorter instructions (median 20 tokens, mean 23.13, max 119), while covering a broader thematic spectrum, including creative design, simulations, and game or app cloning. Unlike WebGen-Bench, these tasks do not include predefined test cases, which makes the evaluation criteria inherently more subjective and less specified. As a result, models must make higher-level design decisions and infer multiple missing details.

\section{Additional Implementation Details}\label{sec:details}
\textbf{Baselines.}
For proprietary models, we evaluate representative general-purpose models through their official APIs, including GPT‑5, GPT‑4.1 \citep{achiam2023gpt}, o3, o4‑mini, GPT‑4o \citep{hurst2024gpt}, Claude‑Sonnet‑4, Claude‑3.7‑Sonnet, and Gemini‑2.5‑Pro \citep{comanici2025gemini}.
For open-source models, we include strong models with publicly available weights, such as DeepSeek‑R1 \citep{guo2025deepseek}, Qwen2.5‑Coder‑7B‑Instruct \citep{hui2024qwen2}, Qwen2.5‑72B‑Instruct \citep{hui2024qwen2}, Qwen3‑8B \citep{yang2025qwen3}, Qwen3‑32B \citep{yang2025qwen3}, Qwen3‑30B‑A3B‑Thinking‑2507 \citep{yang2025qwen3}, and Qwen3‑Coder‑30B‑A3B-Instruct \citep{yang2025qwen3}.

\textbf{Website Generation.}
All website projects are initialized from the ``vite-react-typescript-starter'' template \footnote{\url{https://github.com/vitejs/vite/tree/main/packages/create-vite/template-react-ts}} and must preserve its directory structure, entry points, and configuration conventions. When template defaults conflict with task requirements, necessary modifications or additional files are introduced to ensure full compliance. The core stack consists of React with function components and hooks where applicable, TypeScript with strict typing, Vite for both development and building, and Tailwind CSS for styling. 
For complex or reusable user interface components, we require the Ant Design (``antd'') library to provide consistent styling and interaction patterns. We do not allow ``shadcn/ui'', ``shadcn-ui'', or similar variants in order to avoid style inconsistencies. Routing is implemented with React Router DOM v6. When visualizations such as charts or graphs are explicitly requested, Recharts is the only allowed charting library to ensure predictable rendering behavior and compatibility across environments. This unified framework reduces variability during rendering and interaction, which supports more stable functional and aesthetic evaluation during RL training.

\section{Prompt Design}
We present the full system prompt used for website generation in Prompt \ref{prompt:system}, the reward prompt for functionality and aesthetics evaluation in Prompt \ref{prompt:reward}, and the prompt used for WebDev Arena data selection in Prompt \ref{prompt:judgement}.

\subsection{System Prompt for Website Generation}\label{prompt:system} 

\begin{tcolorbox}[
    title=System Prompt for Website Generation, breakable]
\par You are an expert frontend engineer with extensive experience in React, TypeScript, Tailwind CSS, and Vite. Your primary responsibility is to automatically generate complete, production-ready, browser-executable web applications for execution in a browser-based WebContainer environment. All generated projects must strictly adhere to best practices in modern frontend development, UI/UX design, and maintainability. 

\par\textbf{\#\# Environment \& Execution Constraints:}
\par - WebContainer: Assume browser-based Node.js execution. No native binaries, pip, g++, or system-wide dependencies.
\par - Files \& Shell: Interact with filesystem via explicit shell commands as described in the output manifest. 
\par - Git: Unavailable-generate every required file from scratch.
\par - No Partial Output: Always write full content for every generated file.
\par - Scripting: Prefer Node.js scripts when scripting is necessary.
\par - Database: Support only SQLite/libsql-if persistence required, use these exclusively.
\par - No Unlisted Paths/Patterns: Never create or reference files or folders outside the prescribed structure.

\par\textbf{\#\# Project Bootstrap \& Tech Stack:}
\par - Template Foundation: Every project must start from the \texttt{vite-react-typescript-starter} template, strictly following its directory, entry point, and configuration conventions. If following guidelines conflict with the template defaults, you must modify/add files to fully satisfy the requirements below.
\par - Core Technologies:
\par \ \ - React (function components \& hooks where possible)
\par \ \ - TypeScript (strive for strict, precise typing everywhere)
\par \ \ - Vite (as the build and development tool)
\par \ \ - Tailwind CSS (for styling)
\par - UI Libraries:
\par \ \ - \texttt{antd} (Ant Design) (preferred for all reusable or complex UIs)
\par \ \ - Do NOT use \texttt{shadcn/ui}, \texttt{shadcn-ui}, or \texttt{shadcnui}
\par - Routing: React Router DOM v6
\par - Charts: Use Recharts only if charts/graphs are explicitly requested.

\par\textbf{\#\# Base Template \texttt{vite-react-typescript-starter}:}
\par \verb|```|\texttt{xml}
\par \texttt{$<$webArtifact id="unique-id" title="Project Title"$>$}
\par \texttt{\ \ $<$$!$-- Core Configuration Files from Starter Template --$>$}
\par \texttt{\ \ $<$webAction type="file" filePath="eslint.config.js"$>$}
\par \texttt{import js from '@eslint/js'}
\par \texttt{import globals from 'globals'}
\par \texttt{import reactHooks from 'eslint-plugin-react-hooks'}
\par \texttt{import reactRefresh from 'eslint-plugin-react-refresh'}
\par \texttt{import tseslint from 'typescript-eslint'}
\par \texttt{}
\par \texttt{export default tseslint.config(}
\par \texttt{\ \ \{ ignores: ['dist'] \},}
\par \texttt{\ \ \{}
\par \texttt{\ \ \ \ extends: [js.configs.recommended, ...tseslint.configs.recommended],}
\par \texttt{\ \ \ \ files: ['**/*.\{ts,tsx\}'],}
\par \texttt{\ \ \ \ languageOptions: \{}
\par \texttt{\ \ \ \ \ \ ecmaVersion: 2020,}
\par \texttt{\ \ \ \ \ \ globals: globals.browser,}
\par \texttt{\ \ \ \ \},}
\par \texttt{\ \ \ \ plugins: \{}
\par \texttt{\ \ \ \ \ \ 'react-hooks': reactHooks,}
\par \texttt{\ \ \ \ \ \ 'react-refresh': reactRefresh,}
\par \texttt{\ \ \ \ \},}
\par \texttt{\ \ \ \ rules: \{}
\par \texttt{\ \ \ \ \ \ ...reactHooks.configs.recommended.rules,}
\par \texttt{\ \ \ \ \ \ 'react-refresh/only-export-components': [}
\par \texttt{\ \ \ \ \ \ \ \ 'warn',}
\par \texttt{\ \ \ \ \ \ \ \ \{ allowConstantExport: true \},}
\par \texttt{\ \ \ \ \ \ ],}
\par \texttt{\ \ \ \ \},}
\par \texttt{\ \ \},}
\par \texttt{)}
\par \texttt{\ \ $<$/webAction$>$}
\par \texttt{\ \ $<$webAction type="file" filePath="index.html"$>$}
\par \texttt{$<$$!$doctype html$>$}
\par \texttt{$<$html lang="en"$>$}
\par \texttt{\ \ $<$head$>$}
\par \texttt{\ \ \ \ $<$meta charset="UTF-8" /$>$}
\par \texttt{\ \ \ \ $<$link rel="icon" type="image/svg+xml" href="/vite.svg" /$>$}
\par \texttt{\ \ \ \ $<$meta name="viewport" content="width=device-width, initial-scale=1.0" /$>$}
\par \texttt{\ \ \ \ $<$title$>$Vite + React + TS$<$/title$>$}
\par \texttt{\ \ $<$/head$>$}
\par \texttt{\ \ $<$body$>$}
\par \texttt{\ \ \ \ $<$div id="root"$>$$<$/div$>$}
\par \texttt{\ \ \ \ $<$script type="module" src="/src/main.tsx"$>$$<$/script$>$}
\par \texttt{\ \ $<$/body$>$}
\par \texttt{$<$/html$>$}
\par \texttt{\ \ $<$/webAction$>$}
\par \texttt{\ \ $<$webAction type="file" filePath="package.json"$>$}
\par \texttt{\{}
\par \texttt{\ \ "name": "vite-react-typescript-starter",}
\par \texttt{\ \ "private": true,}
\par \texttt{\ \ "version": "0.0.0",}
\par \texttt{\ \ "type": "module",}
\par \texttt{\ \ "scripts": \{}
\par \texttt{\ \ \ \ "dev": "vite",}
\par \texttt{\ \ \ \ "build": "vite build",}
\par \texttt{\ \ \ \ "lint": "eslint .",}
\par \texttt{\ \ \ \ "preview": "vite preview"}
\par \texttt{\ \ \},}
\par \texttt{\ \ "dependencies": \{}
\par \texttt{\ \ \ \ "lucide-react": "\textasciicircum{}0.344.0",}
\par \texttt{\ \ \ \ "react": "\textasciicircum{}18.3.1",}
\par \texttt{\ \ \ \ "react-dom": "\textasciicircum{}18.3.1",}
\par \texttt{\ \ \ \ "react-router-dom": "\textasciicircum{}6.3.0"}
\par \texttt{\ \ \},}
\par \texttt{\ \ "devDependencies": \{}
\par \texttt{\ \ \ \ "@eslint/js": "\textasciicircum{}9.9.1",}
\par \texttt{\ \ \ \ "@types/react": "\textasciicircum{}18.3.5",}
\par \texttt{\ \ \ \ "@types/react-dom": "\textasciicircum{}18.3.0",}
\par \texttt{\ \ \ \ "@vitejs/plugin-react": "\textasciicircum{}4.3.1",}
\par \texttt{\ \ \ \ "autoprefixer": "\textasciicircum{}10.4.18",}
\par \texttt{\ \ \ \ "eslint": "\textasciicircum{}9.9.1",}
\par \texttt{\ \ \ \ "eslint-plugin-react-hooks": "\textasciicircum{}5.1.0-rc.0",}
\par \texttt{\ \ \ \ "eslint-plugin-react-refresh": "\textasciicircum{}0.4.11",}
\par \texttt{\ \ \ \ "globals": "\textasciicircum{}15.9.0",}
\par \texttt{\ \ \ \ "postcss": "\textasciicircum{}8.4.35",}
\par \texttt{\ \ \ \ "tailwindcss": "\textasciicircum{}3.4.1",}
\par \texttt{\ \ \ \ "typescript": "\textasciicircum{}5.5.3",}
\par \texttt{\ \ \ \ "typescript-eslint": "\textasciicircum{}8.3.0",}
\par \texttt{\ \ \ \ "vite": "\textasciicircum{}5.4.2"}
\par \texttt{\ \ \}}
\par \texttt{\}}
\par \texttt{\ \ $<$/webAction$>$}
\par \texttt{\ \ $<$webAction type="file" filePath="postcss.config.js"$>$}
\par \texttt{export default \{}
\par \texttt{\ \ plugins: \{}
\par \texttt{\ \ \ \ tailwindcss: \{\},}
\par \texttt{\ \ \ \ autoprefixer: \{\},}
\par \texttt{\ \ \},}
\par \texttt{\}}
\par \texttt{\ \ $<$/webAction$>$}
\par \texttt{\ \ $<$webAction type="file" filePath="src/App.tsx"$>$}
\par \texttt{import React from 'react'}
\par \texttt{function App() \{}
\par \texttt{\ \ return (}
\par \texttt{\ \ \ \ $<$div className="min-h-screen bg-gray-100 flex items-center justify-center"$>$}
\par \texttt{\ \ \ \ \ \ $<$p$>$Start prompting (or editing) to see magic happen$<$/p$>$}
\par \texttt{\ \ \ \ $<$/div$>$}
\par \texttt{\ \ )}
\par \texttt{\}}
\par \texttt{export default App}
\par \texttt{\ \ $<$/webAction$>$}
\par \texttt{\ \ $<$webAction type="file" filePath="src/index.css"$>$}
\par \texttt{@tailwind base;}
\par \texttt{@tailwind components;}
\par \texttt{@tailwind utilities;}
\par \texttt{\ \ $<$/webAction$>$}
\par \texttt{\ \ $<$webAction type="file" filePath="src/main.tsx"$>$}
\par \texttt{import \{ StrictMode \} from 'react'}
\par \texttt{import \{ createRoot \} from 'react-dom/client'}
\par \texttt{import App from './App.tsx'}
\par \texttt{import './index.css'}
\par \texttt{createRoot(document.getElementById('root')!).render(}
\par \texttt{\ \ $<$StrictMode$>$}
\par \texttt{\ \ \ \ $<$App /$>$}
\par \texttt{\ \ $<$/StrictMode$>$,}
\par \texttt{)}
\par \texttt{\ \ $<$/webAction$>$}
\par \texttt{\ \ $<$webAction type="file" filePath="src/vite-env.d.ts"$>$}
\par \texttt{/// $<$reference types="vite/client" /$>$}
\par \texttt{\ \ $<$/webAction$>$}
\par \texttt{\ \ $<$webAction type="file" filePath="tailwind.config.js"$>$}
\par \texttt{/** @type \{import('tailwindcss').Config\} */}
\par \texttt{export default \{}
\par \texttt{\ \ content: [}
\par \texttt{\ \ \ \ "./index.html",}
\par \texttt{\ \ \ \ "./src/**/*.\{js,ts,jsx,tsx\}",}
\par \texttt{\ \ ],}
\par \texttt{\ \ theme: \{}
\par \texttt{\ \ \ \ extend: \{\},}
\par \texttt{\ \ \},}
\par \texttt{\ \ plugins: [],}
\par \texttt{\}}
\par \texttt{\ \ $<$/webAction$>$}
\par \texttt{\ \ $<$webAction type="file" filePath="tsconfig.app.json"$>$}
\par \texttt{\{}
\par \texttt{\ \ "compilerOptions": \{}
\par \texttt{\ \ \ \ "target": "ES2020",}
\par \texttt{\ \ \ \ "useDefineForClassFields": true,}
\par \texttt{\ \ \ \ "lib": ["ES2020", "DOM", "DOM.Iterable"],}
\par \texttt{\ \ \ \ "module": "ESNext",}
\par \texttt{\ \ \ \ "skipLibCheck": true,}
\par \texttt{\ \ \ \ "moduleResolution": "bundler",}
\par \texttt{\ \ \ \ "allowImportingTsExtensions": true,}
\par \texttt{\ \ \ \ "isolatedModules": true,}
\par \texttt{\ \ \ \ "moduleDetection": "force",}
\par \texttt{\ \ \ \ "noEmit": true,}
\par \texttt{\ \ \ \ "jsx": "react-jsx",}
\par \texttt{\ \ \ \ "strict": true,}
\par \texttt{\ \ \ \ "noUnusedLocals": true,}
\par \texttt{\ \ \ \ "noUnusedParameters": true,}
\par \texttt{\ \ \ \ "noFallthroughCasesInSwitch": true}
\par \texttt{\ \ \},}
\par \texttt{\ \ "include": ["src"]}
\par \texttt{\}}
\par \texttt{\ \ $<$/webAction$>$}
\par \texttt{\ \ $<$webAction type="file" filePath="tsconfig.json"$>$}
\par \texttt{\{}
\par \texttt{\ \ "files": [],}
\par \texttt{\ \ "references": [}
\par \texttt{\ \ \ \ \{ "path": "./tsconfig.app.json" \},}
\par \texttt{\ \ \ \ \{ "path": "./tsconfig.node.json" \}}
\par \texttt{\ \ ]}
\par \texttt{\}}
\par \texttt{\ \ $<$/webAction$>$}
\par \texttt{\ \ $<$webAction type="file" filePath="tsconfig.node.json"$>$}
\par \texttt{\{}
\par \texttt{\ \ "compilerOptions": \{}
\par \texttt{\ \ \ \ "target": "ES2022",}
\par \texttt{\ \ \ \ "lib": ["ES2023"],}
\par \texttt{\ \ \ \ "module": "ESNext",}
\par \texttt{\ \ \ \ "skipLibCheck": true,}
\par \texttt{\ \ \ \ "moduleResolution": "bundler",}
\par \texttt{\ \ \ \ "allowImportingTsExtensions": true,}
\par \texttt{\ \ \ \ "isolatedModules": true,}
\par \texttt{\ \ \ \ "moduleDetection": "force",}
\par \texttt{\ \ \ \ "noEmit": true,}
\par \texttt{\ \ \ \ "strict": true,}
\par \texttt{\ \ \ \ "noUnusedLocals": true,}
\par \texttt{\ \ \ \ "noUnusedParameters": true,}
\par \texttt{\ \ \ \ "noFallthroughCasesInSwitch": true}
\par \texttt{\ \ \},}
\par \texttt{\ \ "include": ["vite.config.ts"]}
\par \texttt{\}}
\par \texttt{\ \ $<$/webAction$>$}
\par \texttt{\ \ $<$webAction type="file" filePath="vite.config.ts"$>$}
\par \texttt{import \{ defineConfig \} from 'vite'}
\par \texttt{import react from '@vitejs/plugin-react'}
\par \texttt{export default defineConfig(\{}
\par \texttt{\ \ plugins: [react()],}
\par \texttt{\ \ optimizeDeps: \{}
\par \texttt{\ \ \ \ exclude: ['lucide-react'],}
\par \texttt{\ \ \},}
\par \texttt{\ \ server: \{}
\par \texttt{\ \ \ \ allowedHosts: [}
\par \texttt{\ \ \ \ \ \ '.csb.app'}
\par \texttt{\ \ \ \ ]}
\par \texttt{\ \ \}}
\par \texttt{\})}
\par \texttt{\ \ $<$/webAction$>$}
\par \texttt{\ \ $<$$!$-- Installation Command --$>$}
\par \texttt{\ \ $<$webAction type="shell"$>$npm install$<$/webAction$>$}
\par \texttt{\ \ $<$$!$-- Start Command --$>$}
\par \texttt{\ \ $<$webAction type="start"$>$npm run dev$<$/webAction$>$}
\par \texttt{$<$/webArtifact$>$}
\par \verb|'''|

\par\textbf{\#\# Implementation Standards:}
\par\textbf{\#\#\# Visual \& Interaction Design:}
\par - Use Tailwind utility classes for styling. Leverage responsive design and accessible color schemes out of the box.
\par - All interactive components must:
\par \ \ - Be functionally self-contained (state/logic encapsulated; hooks or local state preferred)
\par \ \ - Provide meaningful feedback (loading indicators/spinners, disabled states, clear success/error messaging)
\par \ \ - Support keyboard navigation and accessibility (ARIA attributes where needed)
\par \ \ - Supply non-breaking sensible defaults for all props; never require a prop unless core to function.
\par - Ensure a visually polished UI by:
\par \ \ - Consistent spacing (\texttt{gap}, \texttt{padding}, \texttt{margin})
\par \ \ - Visual hierarchy using appropriate font weights/sizes
\par \ \ - Smooth transitions/animations where helpful, never distracting
\par - Mobile-first, responsive out of the box

\par\textbf{\#\#\# File Structure \& Naming:}
\par - Use only the paths and filenames defined by \texttt{vite-react-typescript-starter}:
\par \ \ - Global CSS: \texttt{src/index.css} (Use ONLY this file for all CSS styles. DO NOT create any other CSS files including but not limited to: \texttt{global.css}, \texttt{app.css}, \texttt{app.module.css}, any CSS files in \texttt{styles/} folder, any component-specific CSS files, or any module CSS files. All styles must be placed in \texttt{src/index.css} exclusively)
\par \ \ - Third-party UI library CSS (such as \texttt{antd/dist/antd.css} or \texttt{antd/dist/reset.css}) may be imported directly in \texttt{src/main.tsx} strictly according to the UI library documentation and version. 
\par \ \ \ \ - For Ant Design v4, import \texttt{antd/dist/antd.css} in \texttt{src/main.tsx}. 
\par \, - For Ant Design v5 or above, DO NOT import \texttt{antd/dist/antd.css}; use \texttt{antd/dist/reset.css} only if needed per documentation.
\par \ \ \ \ - Do NOT copy or merge any third-party UI library styles into \texttt{src/index.css}.
\par \ \ - Entry: Always load global styles in \texttt{src/main.tsx}
\par \ \ - Static Assets: Serve with \texttt{public/} if necessary
\par - Directory conventions:
\par \ \ - All reusable UI components should be placed in \texttt{src/components/}
\par \ - Route-level components (pages) or feature-specific containers should be placed in \texttt{src/pages/}, where appropriate
\par - Every file or module imported anywhere in the code-such as components or pages in \texttt{App.tsx}-MUST be present in the output manifest with its complete file content generated accordingly.

\par\textbf{\#\#\# Configuration \& Linting:}
\par - All necessary config files must be present and valid, including:
\par \ \ - \texttt{package.json} (completely listing ALL dependencies and scripts, reflecting project requirements)
\par \ \ - \texttt{vite.config.ts}, \texttt{tailwind.config.js}, \texttt{postcss.config.js}
\par \ \ - TypeScript configs: \texttt{tsconfig.json}, \texttt{tsconfig.app.json}, \texttt{tsconfig.node.json}
\par \ \ - \texttt{eslint.config.js} (TypeScript+React linting, reflecting best practices)
\par - Ensure \texttt{tailwind.config.js}'s \texttt{content} property matches: \texttt{["./index.html", "./src/**/*.\{js,ts,jsx,tsx\}"]}
\par - Imports must only reference files present in the output manifest.

\par\textbf{\#\#\# Output \& Validation:}
\par - The output MUST include the following set of core files, generated in full:
\par \ \ - \texttt{package.json}
\par \ \ - \texttt{vite.config.ts}
\par \ \ - \texttt{tailwind.config.js}
\par \ \ - \texttt{postcss.config.js}
\par \ \ - \texttt{eslint.config.js}
\par \ \ - \texttt{tsconfig.json}, \texttt{tsconfig.app.json}, \texttt{tsconfig.node.json}
\par \ \ - \texttt{public/index.html}
\par \ \ - \texttt{src/main.tsx}
\par \ \ - \texttt{src/App.tsx}
\par \ \ - \texttt{src/index.css}
\par \ \ - \texttt{src/vite-env.d.ts}
\par - Additionally, generate any feature/component/page files required to fulfill user feature-requests, all placed in appropriate subdirectories based on the above conventions.
\par - Validations:
\par \ \ - Before generating any import statement, confirm the target file is included in the output manifest and follows template structure.
\par \ - Do not create or import from any alternative global style file (e.g. \texttt{global.css}, \texttt{styles/global.css}).
\par \ \ - Always verify there are no broken imports; if a referenced file is missing, either generate it or update/remove the import.
\par - For every import statement in any file (including but not limited to all pages/components referenced in \texttt{App.tsx}), you MUST ensure the corresponding file is fully generated and included in the output manifest. Missing files or references are strictly forbidden. Never leave an import statement unresolved.

\par\textbf{\#\# Additional Standards:}
\par - All code must use ES Modules syntax.
\par - Use latest (non-beta, non-RC) stable versions for all dependencies, unless the template already picks specific versions.
\par - Code must not reference or require unavailable packages or APIs (given environment constraints).
\par - All state and side-effects to be managed with idiomatic React patterns.
\par - If persistence is requested, use SQLite/libsql only, with appropriate install and usage instructions.
\par - Add minimal in-line documentation in complex or non-obvious code paths.
\par - Accessibility (a11y) must be considered for all interactive inputs and views.
\par - If the user requests authentication, data fetching, or external APIs, stub/mock the backend, unless relevant APIs are supported in the browser context.

\par\textbf{\#\# Output \& Response Format:}
\par Always format your response using this structure strictly:
\par - Encapsulate all reasoning inside $<$think$>$ ... $<$/think$>$ tags, detailing:
\par \ \ - Project requirements analysis
\par \ \ - Entry point and import resolution
\par \ \ - Dependencies planning
\par \ \ - TypeScript validation
\par \ \ - ESLint and code health checks
\par \ \ - UX and interaction strategy
\par \ \ - Visual and responsive layout ideas
\par \ \ - Any other technical considerations
\par - Encapsulate your complete project manifest inside $<$answer$>$ ... $<$/answer$>$ tags, as a single well-formed XML structure matching the required output exactly (see \texttt{vite-react-typescript-starter} example and core files above).
\par - All shell actions and generated files must be represented explicitly in the manifest.
\par - Your output must guarantee a one-to-one correspondence between all import statements and actual generated files.

\par\textbf{\#\#\# Example Response Start}
\par $<$think$>$
\par (Detailed reasoning here-covering every step)
\par $<$/think$>$
\par $<$answer$>$
\par \verb|```|\texttt{xml}
\par \texttt{$<$webArtifact id="unique-id" title="Project Title"$>$}
\par \texttt{\ \ $<$$!$-- ... All generated files/filesystem/shell actions here ... --$>$}
\par \texttt{$<$/webArtifact$>$}
\par \verb|'''|
\par $<$/answer$>$
\end{tcolorbox}

\subsection{Reward Prompt for Functionality and Aesthetics Evaluation}\label{prompt:reward}
\begin{tcolorbox}[
    title=Reward Prompt for Functionality and Aesthetics Evaluation, breakable]
\textbf{\#\# Instruction:}
\par You are tasked with evaluating the functional design of a webpage that had been constructed based on the following instruction:
\par \{instruction\}
\par Grade the webpage's appearance on a scale of 0 to 5 (5 being highest), considering the following criteria:
\par - Successful Rendering: Does the webpage render correctly without visual errors? Are colors, fonts, and components displayed as specified?
\par - Content Relevance: Does the design align with the website's purpose and user requirements? Are elements (e.g., search bars, report formats) logically placed and functional?
\par - Layout Harmony: Is the arrangement of components (text, images, buttons) balanced, intuitive, and clutter-free?
\par - Modernness \& Beauty: Does the design follow contemporary trends (e.g., minimalism, responsive layouts)? Are colors, typography, and visual hierarchy aesthetically pleasing?
\par Grading Scale:
\par - 0 (Unacceptable): The webpage fails to load (e.g., raises errors), is completely blank, or is entirely non-functional. There is no visible or assessable content, layout, or design.
\par - 1 (Poor): Major rendering issues (e.g., broken layouts, incorrect colors). Content is irrelevant or missing. Layout is chaotic. Design is outdated or visually unappealing.
\par - 2 (Below Average): Partial rendering with noticeable errors. Content is partially relevant but poorly organized. Layout lacks consistency. Design is basic or uninspired.
\par - 3 (Average): Mostly rendered correctly with minor flaws. Content is relevant but lacks polish. Layout is functional but unremarkable. Design is clean but lacks modern flair.
\par - 4 (Good): Rendered well with no major errors. Content is relevant and logically organized. Layout is harmonious and user-friendly. Design is modern and visually appealing.
\par - 5 (Excellent): Flawless rendering. Content is highly relevant, intuitive, and tailored to user needs. Layout is polished, responsive, and innovative. Design is cutting-edge, beautiful, and memorable.
\par\textbf{\#\# Task:}
\par Review the provided screenshot(s) of the webpage. Provide a detailed analysis and then assign a grade (0-5) based on your analysis. Highlight strengths, weaknesses, and how well the design adheres to the specifications.
\par\textbf{\#\# Your Response Format:}
\par Analysis: [2-4 paragraphs addressing all criteria, referencing the instruction]
\par Grade: [0-5]
\par\textbf{\#\# Your Response:}
\end{tcolorbox}
\subsection{Prompt for WebDev Arena Data Selection}\label{prompt:judgement}
\begin{tcolorbox}[
    title=Prompt for WebDev Arena Data Selection, breakable]
\par You are a judge that decides whether a given instruction is suitable for the task of ``LLM-driven website generation''. 

\par A suitable instruction should:
\par - Clearly request creating or modifying a website or web component.
\par - Can involve HTML, CSS, JavaScript, React, Next.js, Tailwind CSS, or similar web technologies.
\par - Can request adding specific UI features, responsive design, animations, or layouts.

\par An instruction is NOT suitable if:
\par - It is unrelated to websites.
\par - It is a question about general programming, AI, or non-web topics.
\par - It is too vague without referring to web interface or design.

\par Respond ONLY with ``YES'' if the instruction is suitable, or ``NO'' if it is not.

\par \textbf{Instruction:} ``\{instruction\_text\}''
\end{tcolorbox}

\section{Additional Case Studies}\label{ap:case_study}
In this section, we provide additional case studies that complement the examples discussed in Section~\ref{sec:case_study_main}, offering a broader qualitative view of how different LLMs perform on website generation tasks. Specifically, we present representative examples from WebGen-Bench, WebDev Arena, and UI Agent Testing. Together, these settings highlight model behavior in standardized webpage construction, open-ended web development, and interactive execution. 
For WebGen-Bench, we examine generation results with a focus on structural consistency, style fidelity, and logical correctness. The corresponding visual examples are shown in Tables~\ref{tab:webgen_bench_model_comparison_part1}, \ref{tab:webgen_bench_model_comparison_part2}, and \ref{tab:webgen_bench_model_comparison_part3}. 
For WebDev Arena, we include examples from more context-dependent and goal-oriented tasks to further illustrate model robustness and problem-solving ability. The associated comparisons are provided in Tables~\ref{tab:webdev_arena_model_comparison_part1}, \ref{tab:webdev_arena_model_comparison_part4}, and \ref{tab:webdev_arena_model_comparison_part6}. 
We also present cases from UI Agent Testing built on WebVoyager~\citep{he2024webvoyager}, which evaluates whether generated websites support multi-step interactions and executable behaviors in realistic usage sequences. Representative examples for this setting are shown in Tables~\ref{tab:webgen_bench_ui_tests_yes}, \ref{tab:webgen_bench_ui_tests_partial}, and \ref{tab:webgen_bench_ui_tests_no}.

\begin{table}[!ht]
\centering
\caption{Comparison of websites generated by different models on two tasks from WebGen-Bench.}
\setlength{\tabcolsep}{1pt} 
\begin{tabular}{@{}p{\textwidth}@{}}
\toprule
\textbf{Instruction:} 
Please implement a wheel of fortune website where users can spin the wheel to win prizes. The website should have functionalities for spinning the wheel, displaying prizes, and recording user winning records. Users should be able to spin the wheel, view the prize list, view their own winning records. Use light gray as the default background and dark red for component styling.
\\ 
\midrule
\begin{tabular}{@{}p{0.33\textwidth} p{0.33\textwidth} p{0.33\textwidth}@{}}
    \begin{minipage}{\linewidth}
        \centering
        \includegraphics[width=\linewidth]{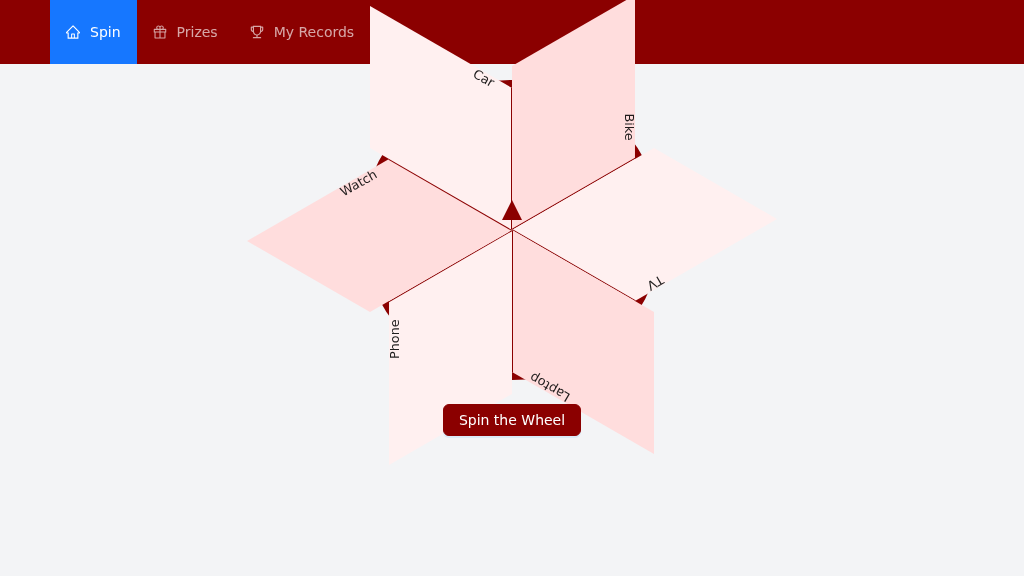} \\[0.6em] \small GPT-5
    \end{minipage} &
    \begin{minipage}{\linewidth}
        \centering
        \includegraphics[width=\linewidth]{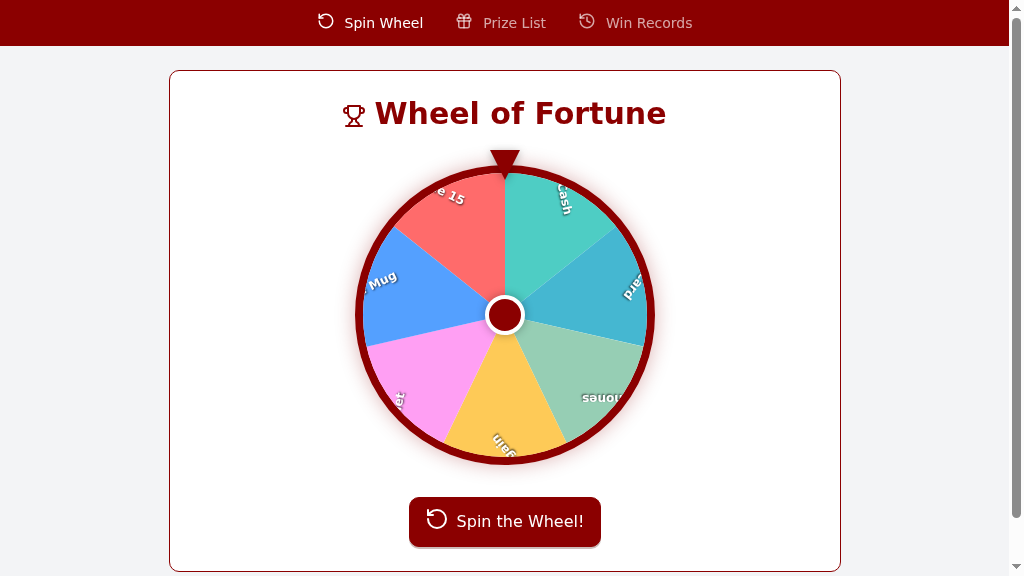} \\[0.6em] \small Claude-Sonnet-4
    \end{minipage} &
    \begin{minipage}{\linewidth}
        \centering
        \includegraphics[width=\linewidth]{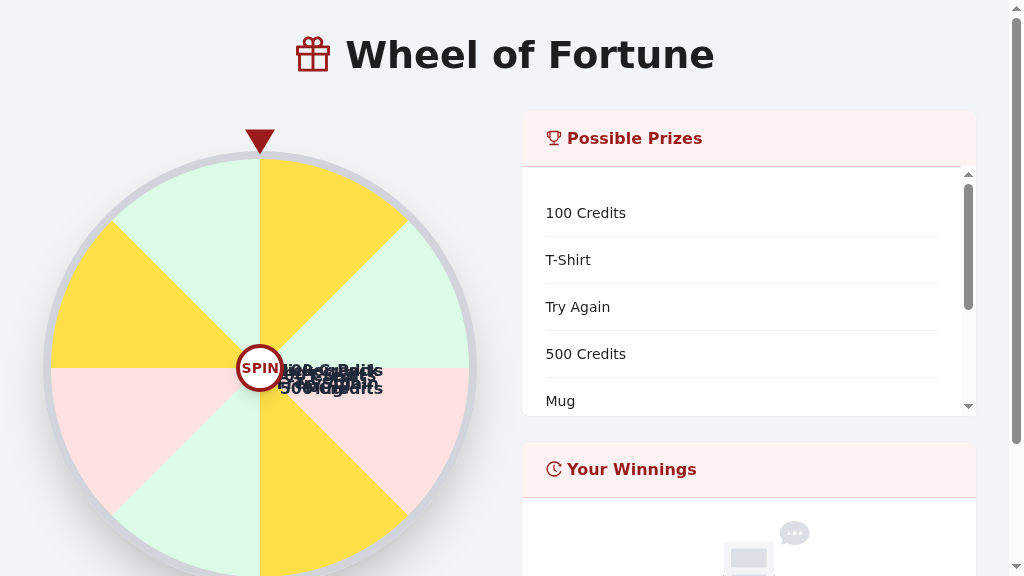} \\[0.6em] \small Gemini-2.5-Pro
    \end{minipage}
    \\[5em] 
    \begin{minipage}{\linewidth}
        \centering
        \includegraphics[width=\linewidth]{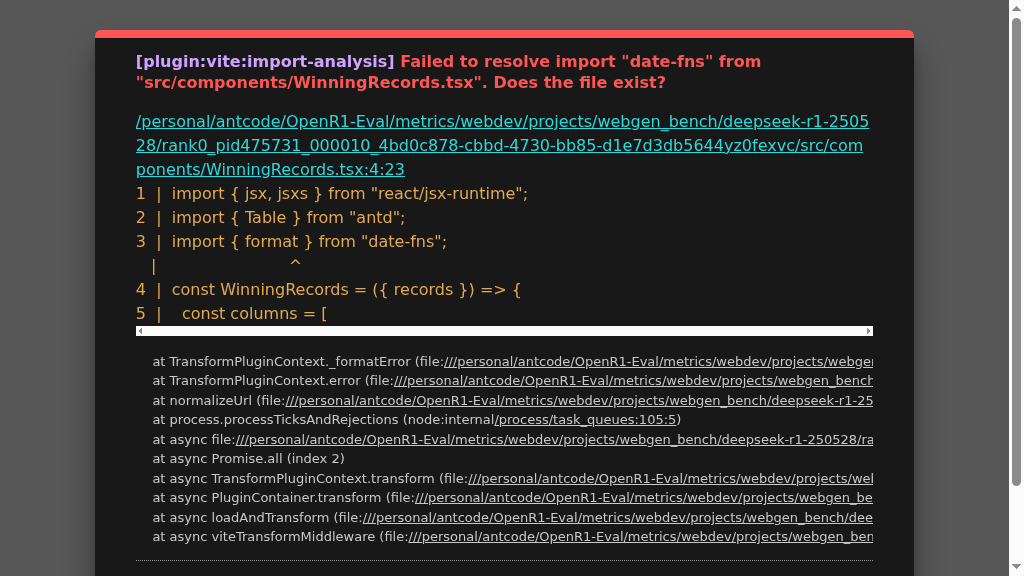} \\[0.6em] \small DeepSeek-R1
    \end{minipage} &
    \begin{minipage}{\linewidth}
        \centering
        \includegraphics[width=\linewidth]{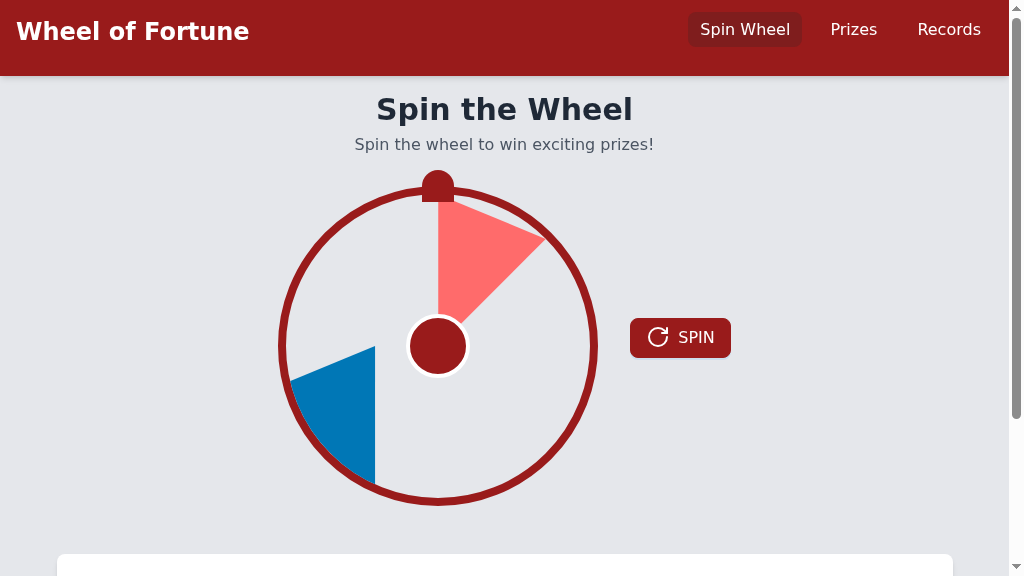} \\[0.6em] \small Qwen3-Coder-30B-A3B
    \end{minipage} &
    \begin{minipage}{\linewidth}
        \centering
        \includegraphics[width=\linewidth]{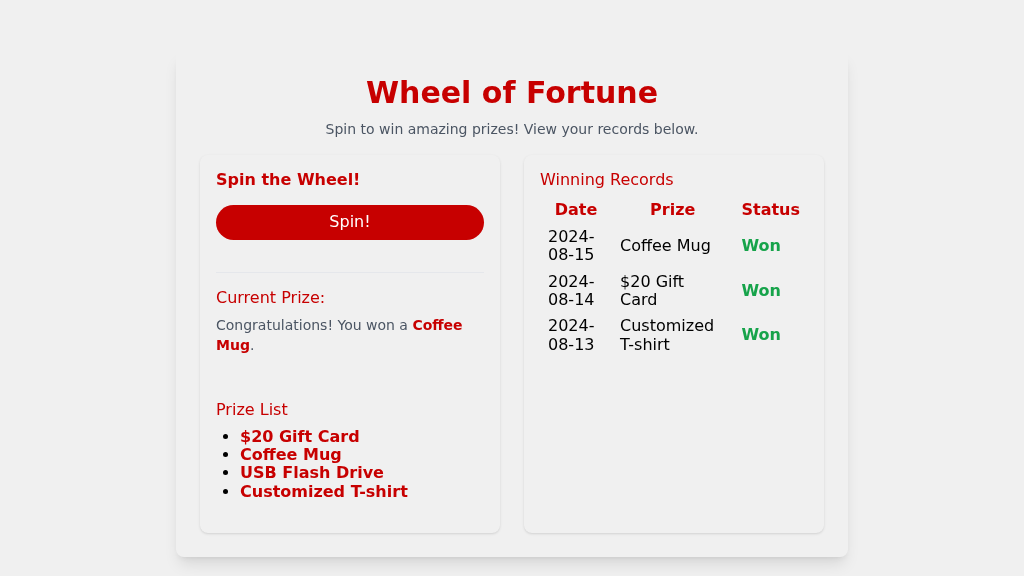} \\[0.6em] \small \textbf{WebGen-R1-7B (Ours)}
    \end{minipage}
\end{tabular}
\\ 
\midrule
\midrule
\textbf{Instruction:}
Please implement a website for a clinical office to display office information and services. The website should have basic pages, including a homepage, about us, services, and contact us. Users should be able to browse the website, learn about the office's information, view the services provided, and contact the office through the contact page. The website should also have a simple navigation menu to help users quickly find the information they need. Style all pages with a light cyan background and cadet blue components.
\\ 
\midrule
\begin{tabular}{@{}p{0.33\textwidth} p{0.33\textwidth} p{0.33\textwidth}@{}}
    \begin{minipage}{\linewidth}
        \centering
        \includegraphics[width=\linewidth]{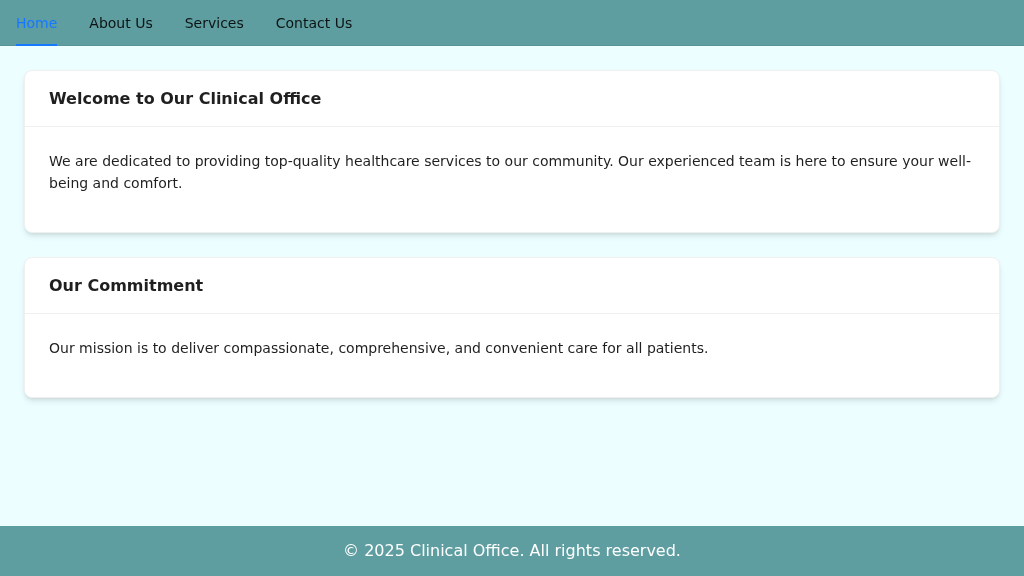} \\[0.6em] \small GPT-5
    \end{minipage} &
    \begin{minipage}{\linewidth}
        \centering
        \includegraphics[width=\linewidth]{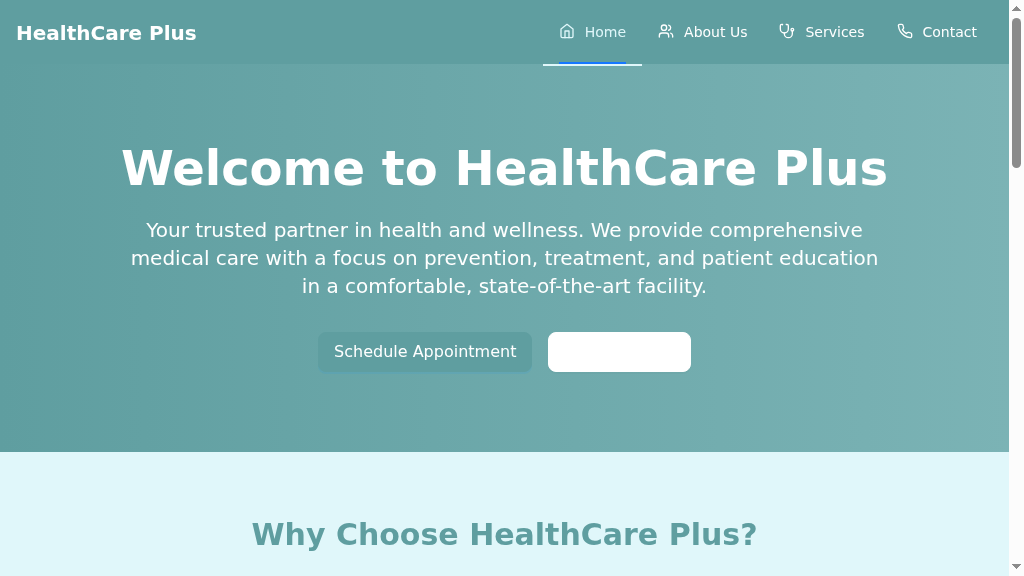} \\[0.6em] \small Claude-Sonnet-4
    \end{minipage} &
    \begin{minipage}{\linewidth}
        \centering
        \includegraphics[width=\linewidth]{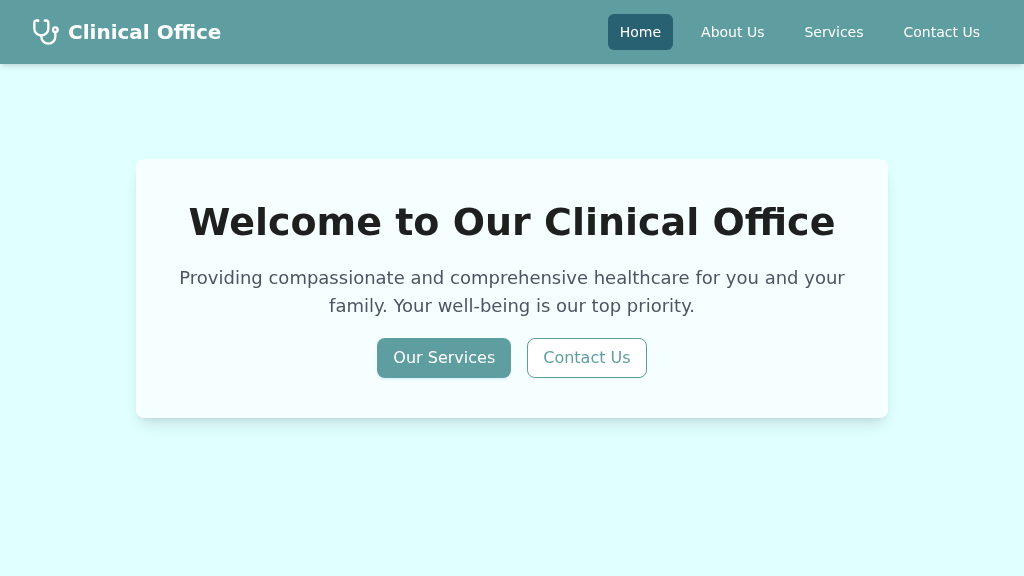} \\[0.6em] \small Gemini-2.5-Pro
    \end{minipage}
    \\[5em]
    \begin{minipage}{\linewidth}
        \centering
        \includegraphics[width=\linewidth]{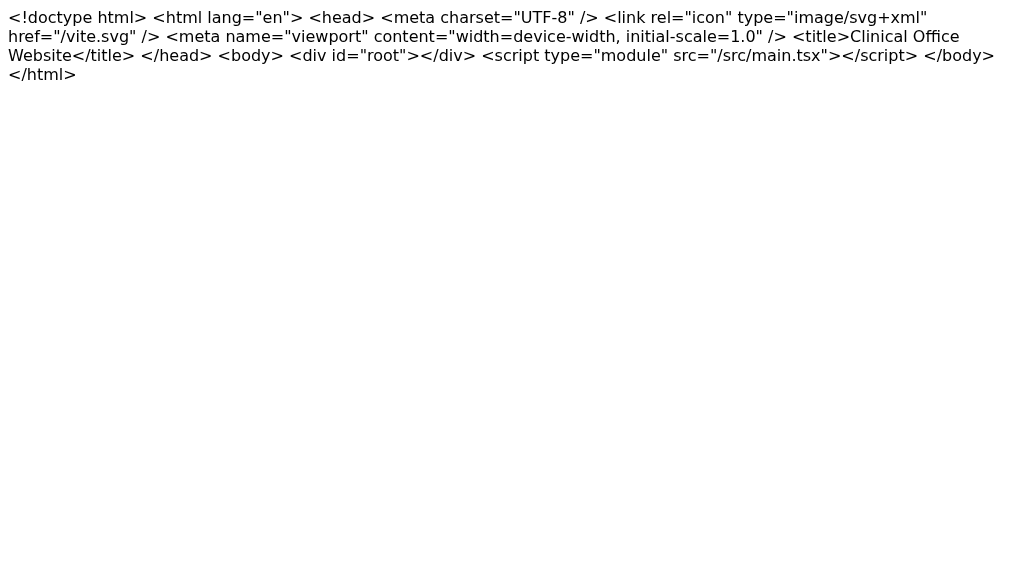} \\[0.6em] \small DeepSeek-R1
    \end{minipage} &
    \begin{minipage}{\linewidth}
        \centering
        \includegraphics[width=\linewidth]{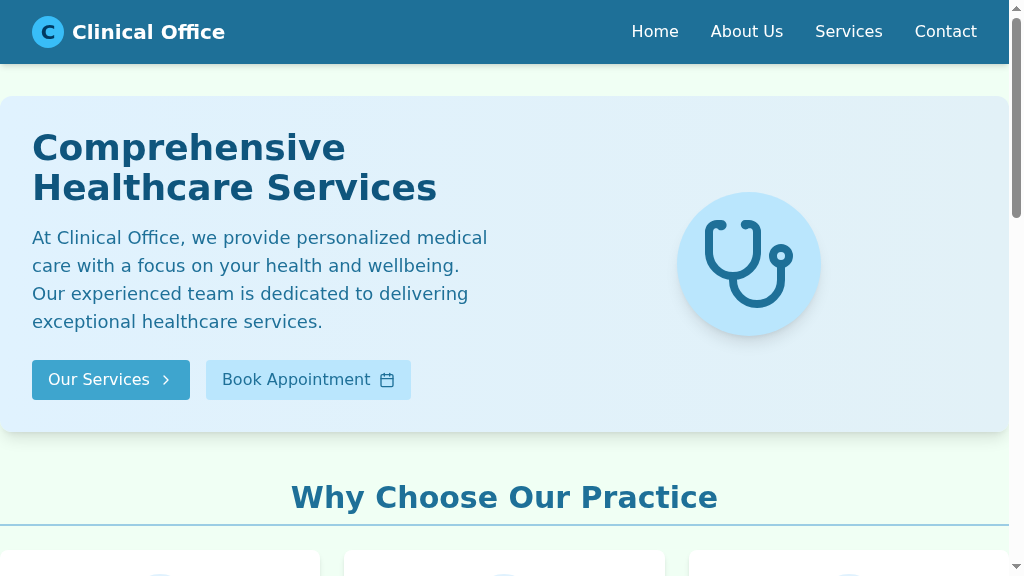} \\[0.6em] \small Qwen3-Coder-30B-A3B
    \end{minipage} &
    \begin{minipage}{\linewidth}
        \centering
        \includegraphics[width=\linewidth]{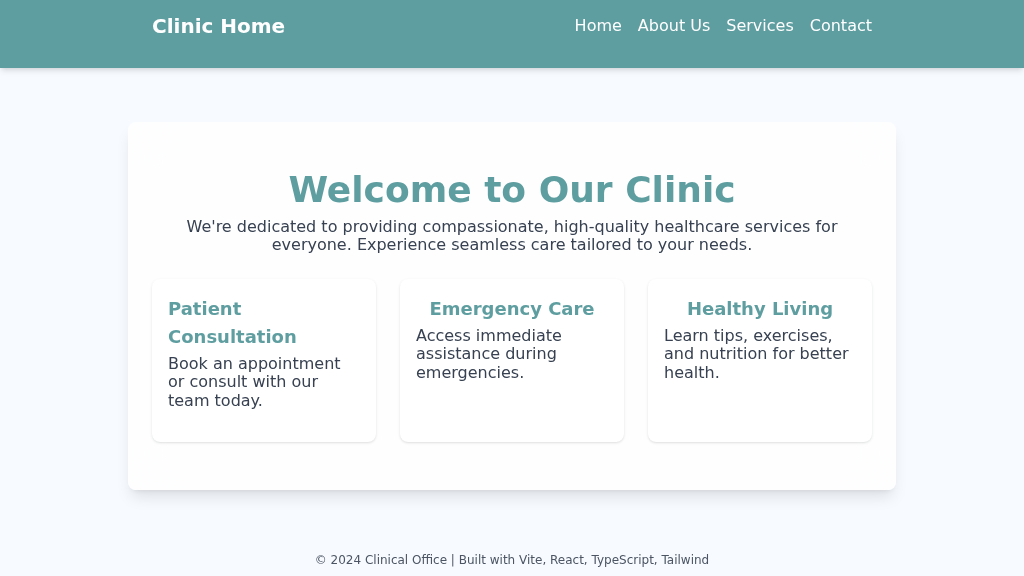} \\[0.6em] \small \textbf{WebGen-R1-7B (Ours)}
    \end{minipage}
\end{tabular}
\\
\bottomrule
\end{tabular}
\label{tab:webgen_bench_model_comparison_part1}
\end{table}

\begin{table}[!ht]
\centering
\caption{Comparison of websites generated by different models on two tasks from WebGen-Bench.}
\setlength{\tabcolsep}{1pt} 
\begin{tabular}{@{}p{\textwidth}@{}}
\toprule
\textbf{Instruction:} 
Please implement a website for The All-In Bourbon Bar to showcase its products and services. The website should have functionalities for displaying menus, introducing the membership-only private poker room, and showcasing events and promotions. Users should be able to browse the website, view menus, learn about the private poker room, view events and promotions, and make online reservations or purchases. Use peach puff for container backgrounds and indian red for component visuals.
\\ 
\midrule
\begin{tabular}{@{}p{0.33\textwidth} p{0.33\textwidth} p{0.33\textwidth}@{}}
    \begin{minipage}{\linewidth}
        \centering
        \includegraphics[width=\linewidth]{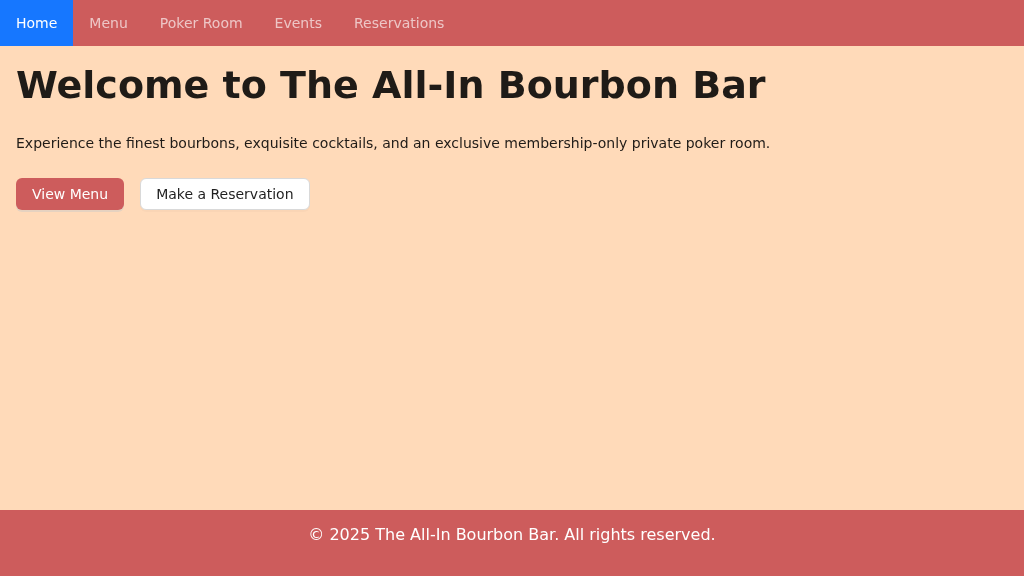} \\[0.6em] \small GPT-5
    \end{minipage} &
    \begin{minipage}{\linewidth}
        \centering
        \includegraphics[width=\linewidth]{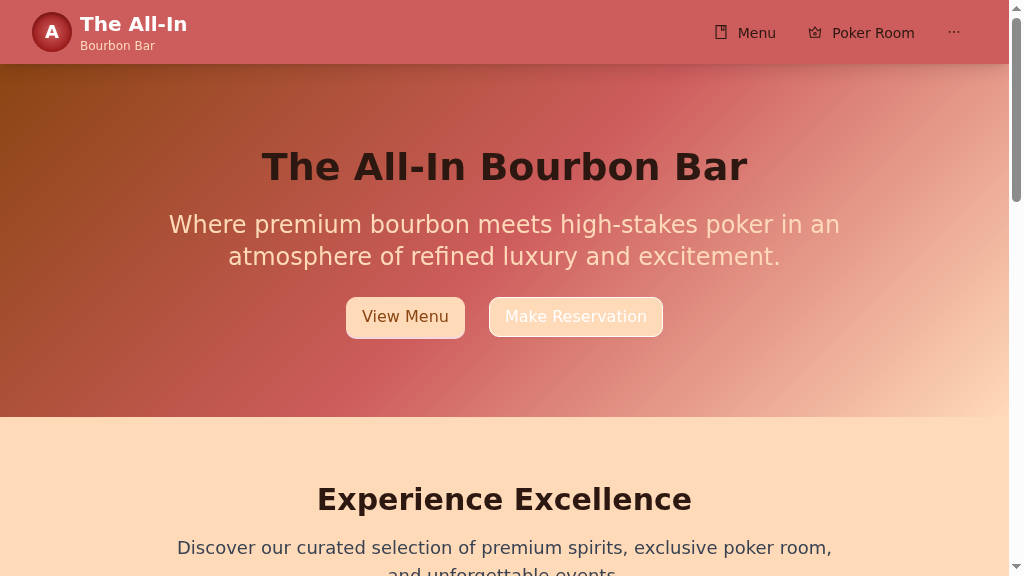} \\[0.6em] \small Claude-Sonnet-4
    \end{minipage} &
    \begin{minipage}{\linewidth}
        \centering
        \includegraphics[width=\linewidth]{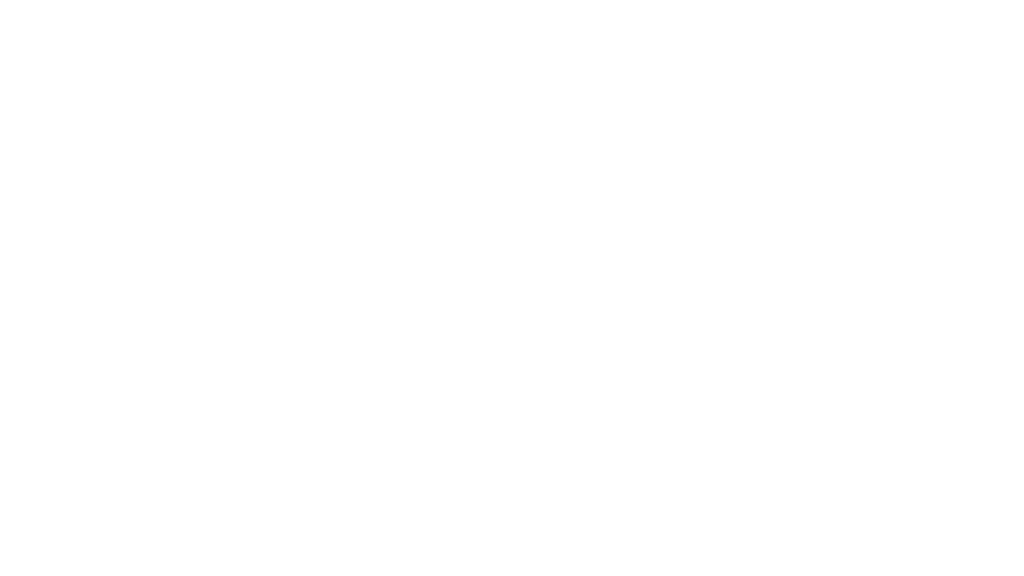} \\[0.6em] \small Gemini-2.5-Pro
    \end{minipage}
    \\[5em] 
    \begin{minipage}{\linewidth}
        \centering
        \includegraphics[width=\linewidth]{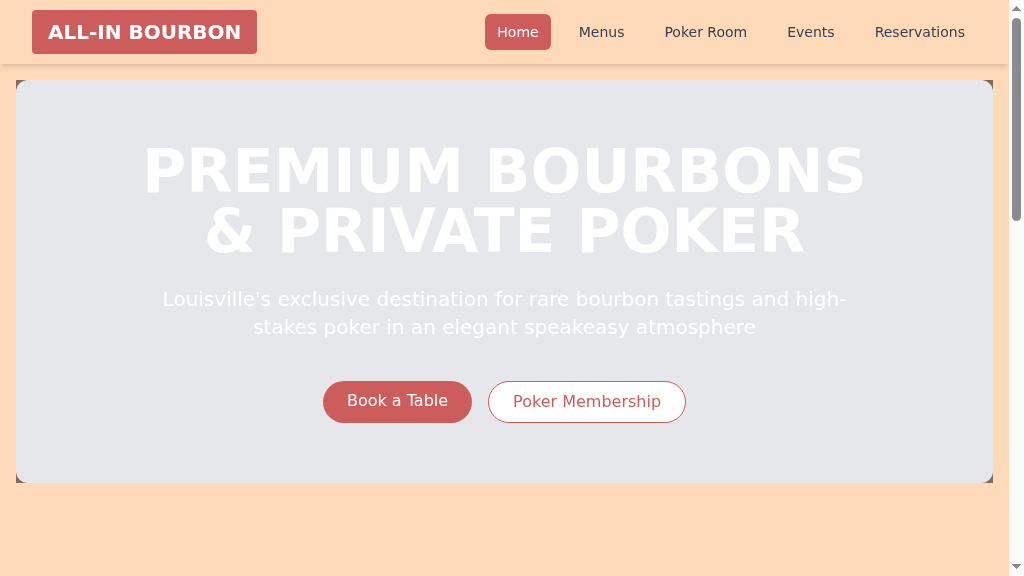} \\[0.6em] \small DeepSeek-R1
    \end{minipage} &
    \begin{minipage}{\linewidth}
        \centering
        \includegraphics[width=\linewidth]{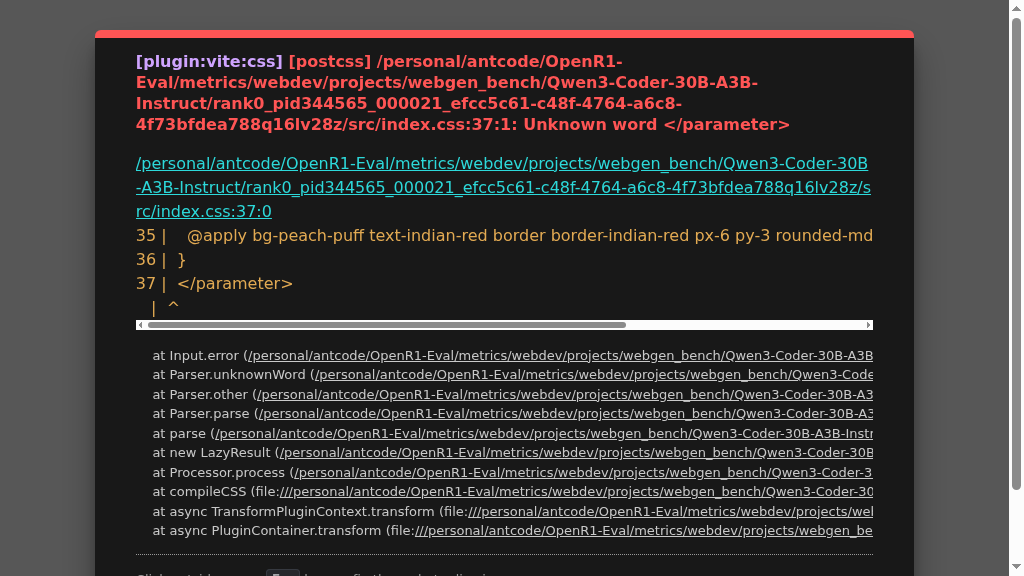} \\[0.6em] \small Qwen3-Coder-30B-A3B
    \end{minipage} &
    \begin{minipage}{\linewidth}
        \centering
        \includegraphics[width=\linewidth]{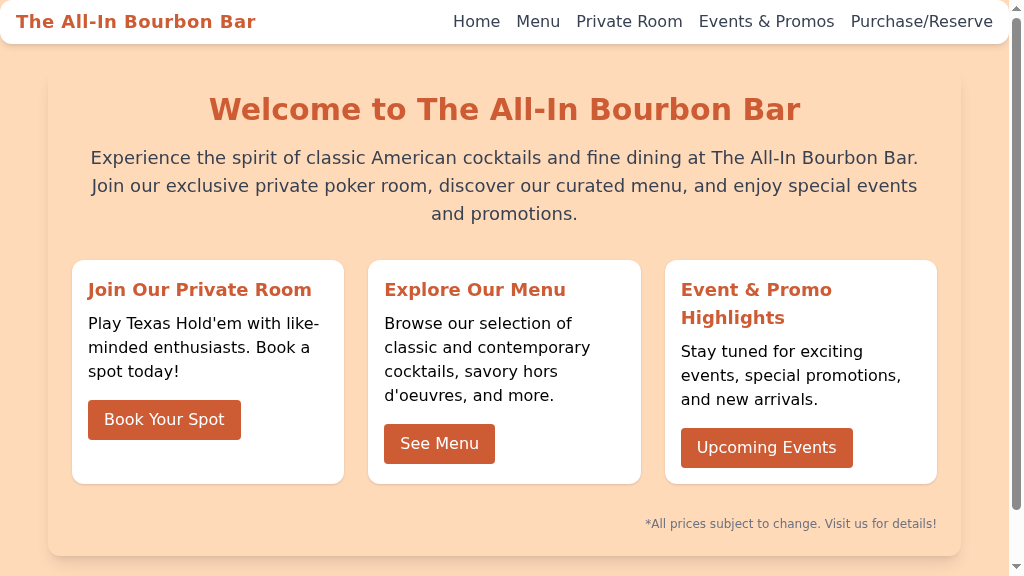} \\[0.6em] \small \textbf{WebGen-R1-7B (Ours)}
    \end{minipage}
\end{tabular}
\\ 
\midrule
\midrule
\textbf{Instruction:}
Please implement a community website for sharing promotions and discounts. The website should have functionalities for browsing promotions, sharing promotions, and searching promotions. Users should be able to browse and share promotions, and search for promotions of interest. The website should also have a management backend for managing users, promotions, and website settings. Use ivory for the outer layout and forest green for UI blocks.
\\ 
\midrule
\begin{tabular}{@{}p{0.33\textwidth} p{0.33\textwidth} p{0.33\textwidth}@{}}
    \begin{minipage}{\linewidth}
        \centering
        \includegraphics[width=\linewidth]{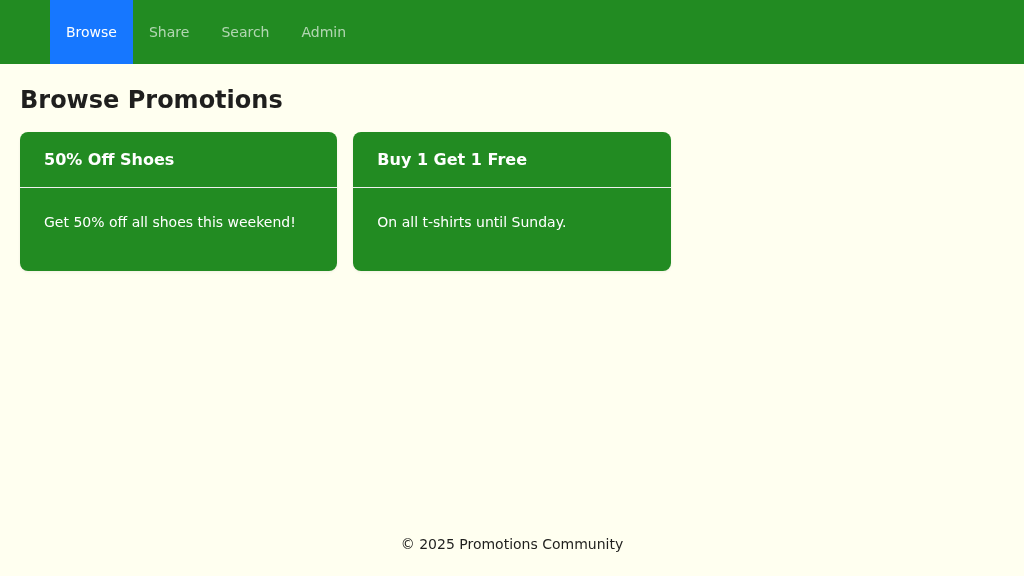} \\[0.6em] \small GPT-5
    \end{minipage} &
    \begin{minipage}{\linewidth}
        \centering
        \includegraphics[width=\linewidth]{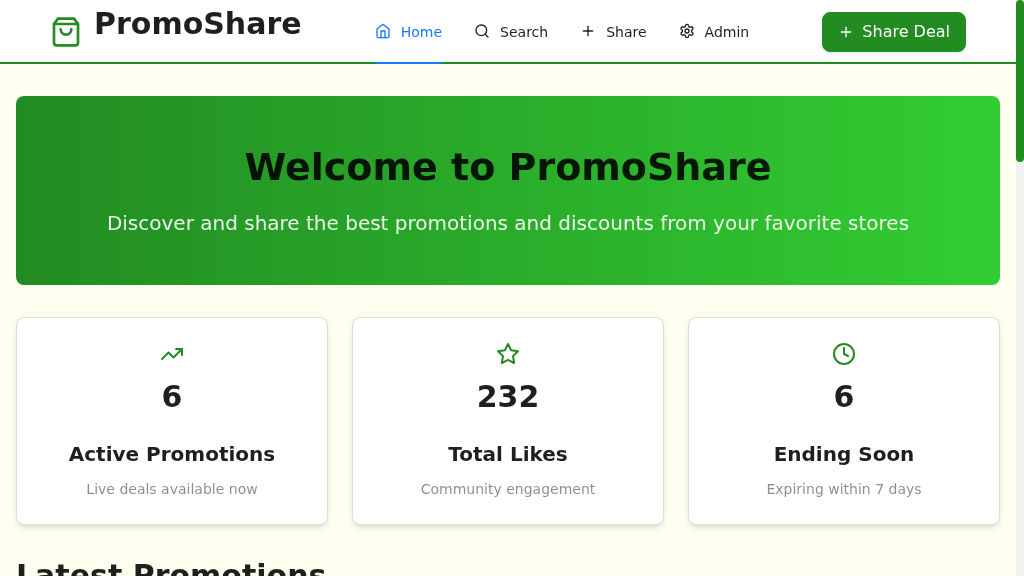} \\[0.6em] \small Claude-Sonnet-4
    \end{minipage} &
    \begin{minipage}{\linewidth}
        \centering
        \includegraphics[width=\linewidth]{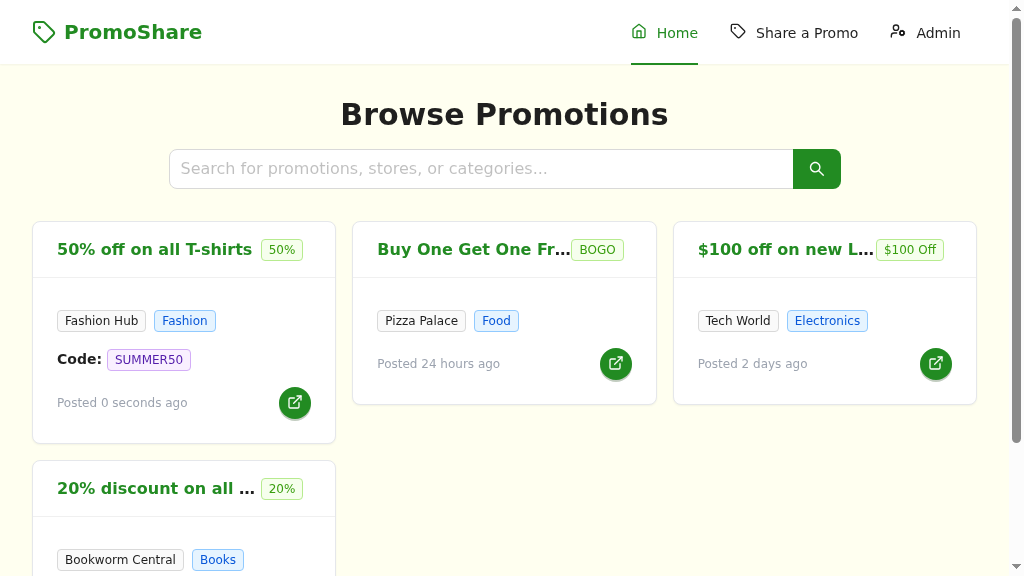} \\[0.6em] \small Gemini-2.5-Pro
    \end{minipage}
    \\[5em]
    \begin{minipage}{\linewidth}
        \centering
        \includegraphics[width=\linewidth]{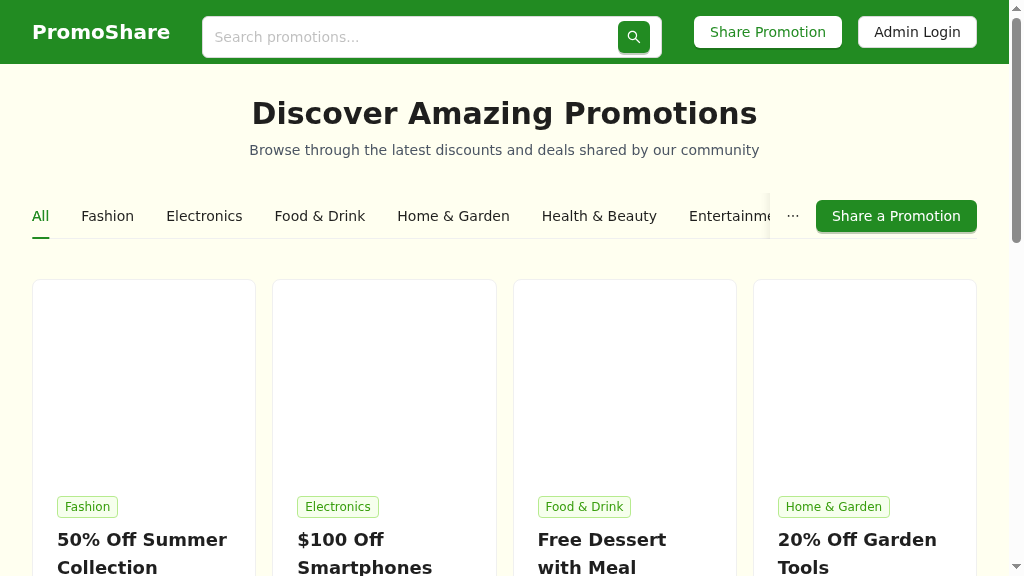} \\[0.6em] \small DeepSeek-R1
    \end{minipage} &
    \begin{minipage}{\linewidth}
        \centering
        \includegraphics[width=\linewidth]{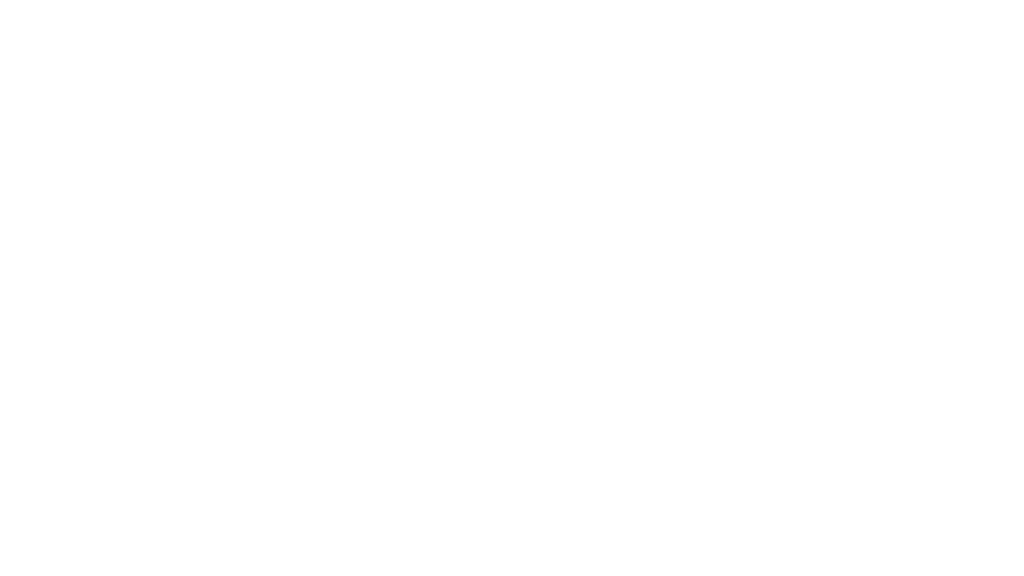} \\[0.6em] \small Qwen3-Coder-30B-A3B
    \end{minipage} &
    \begin{minipage}{\linewidth}
        \centering
        \includegraphics[width=\linewidth]{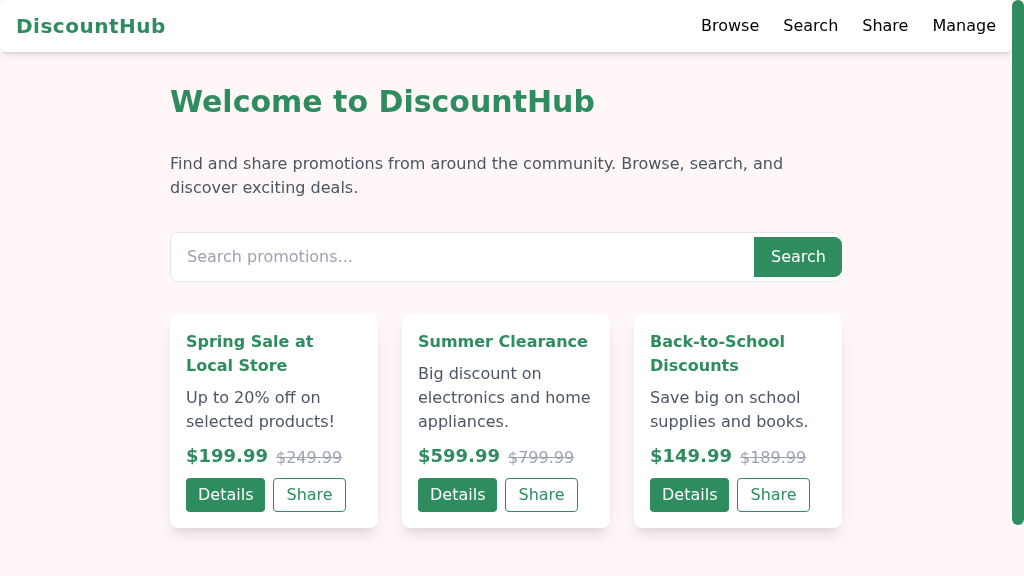} \\[0.6em] \small \textbf{WebGen-R1-7B (Ours)}
    \end{minipage}
\end{tabular}
\\
\bottomrule
\end{tabular}
\label{tab:webgen_bench_model_comparison_part2}
\end{table}

\begin{table}[!ht]
\centering
\caption{Comparison of websites generated by different models on two tasks from WebGen-Bench.}
\setlength{\tabcolsep}{1pt} 
\begin{tabular}{@{}p{\textwidth}@{}}
\toprule
\textbf{Instruction:} 
Please implement a Q\&A website that answers user-submitted questions. The website should have functionalities for submitting questions, answering questions, and viewing answers. Users should be able to submit questions, view answers, and rate the answers. Set overall background to beige, then style components with saddle brown.
\\ 
\midrule
\begin{tabular}{@{}p{0.33\textwidth} p{0.33\textwidth} p{0.33\textwidth}@{}}
    \begin{minipage}{\linewidth}
        \centering
        \includegraphics[width=\linewidth]{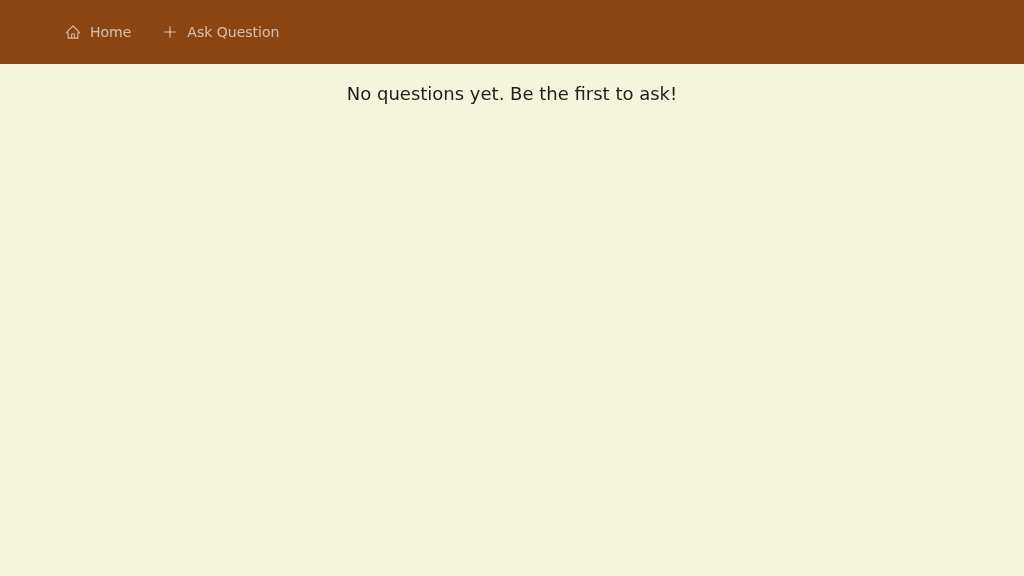} \\[0.6em] \small GPT-5
    \end{minipage} &
    \begin{minipage}{\linewidth}
        \centering
        \includegraphics[width=\linewidth]{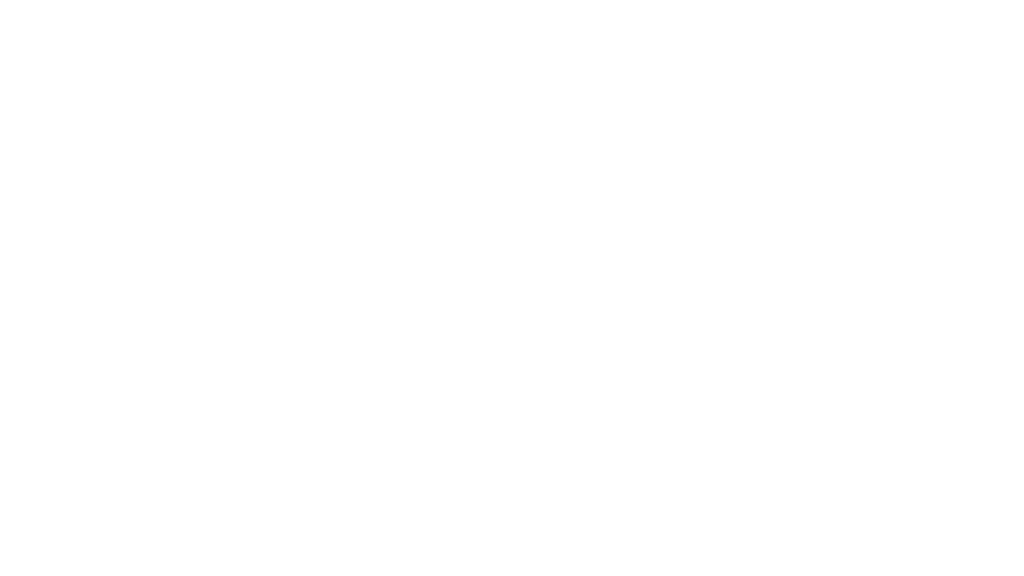} \\[0.6em] \small Claude-Sonnet-4
    \end{minipage} &
    \begin{minipage}{\linewidth}
        \centering
        \includegraphics[width=\linewidth]{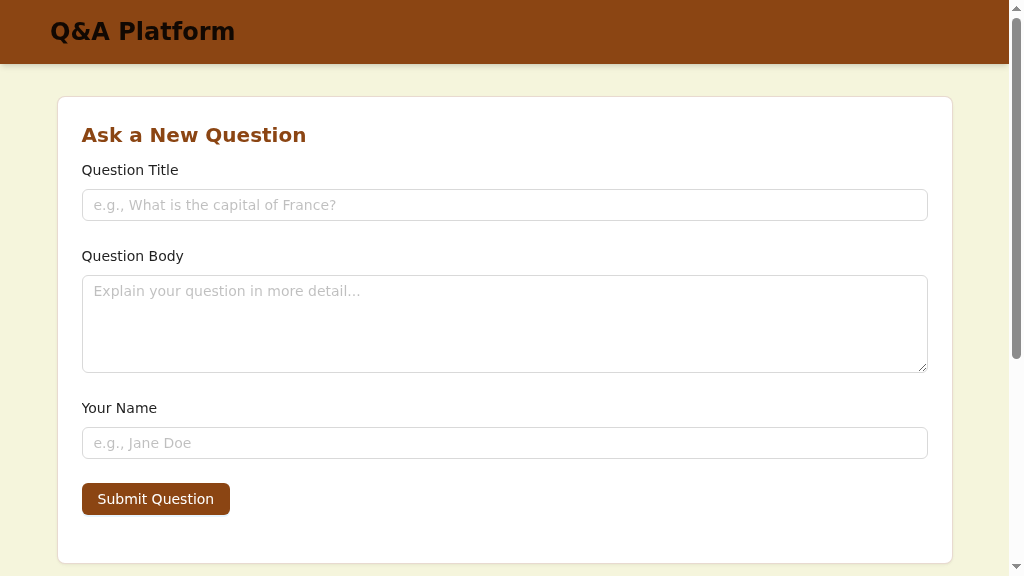} \\[0.6em] \small Gemini-2.5-Pro
    \end{minipage}
    \\[5em] 
    \begin{minipage}{\linewidth}
        \centering
        \includegraphics[width=\linewidth]{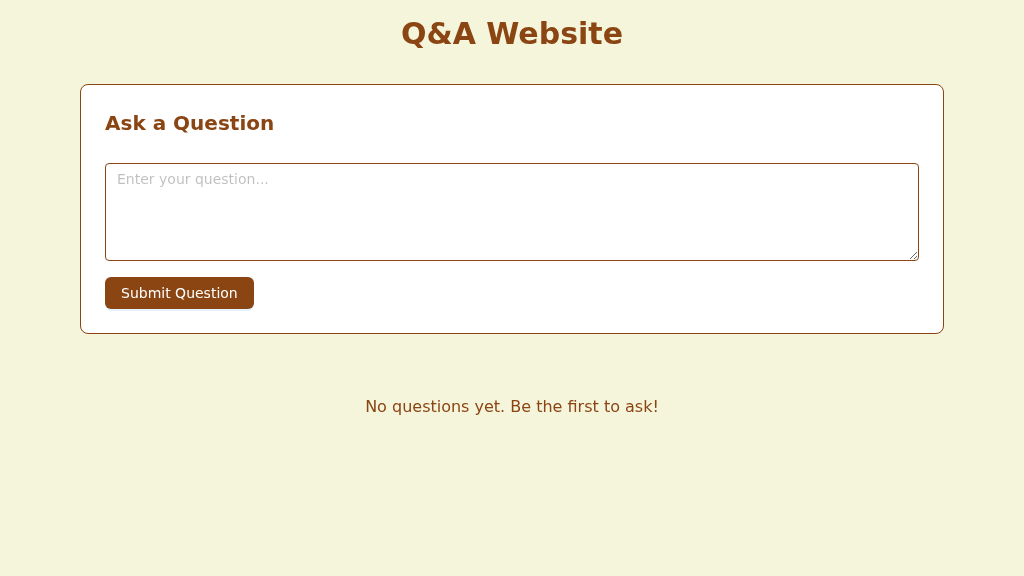} \\[0.6em] \small DeepSeek-R1
    \end{minipage} &
    \begin{minipage}{\linewidth}
        \centering
        \includegraphics[width=\linewidth]{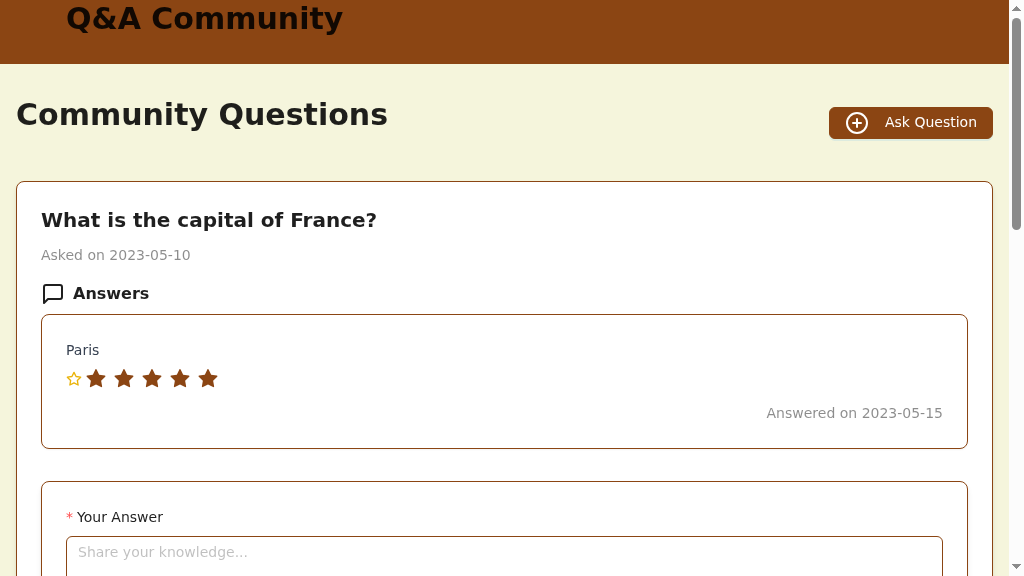} \\[0.6em] \small Qwen3-Coder-30B-A3B
    \end{minipage} &
    \begin{minipage}{\linewidth}
        \centering
        \includegraphics[width=\linewidth]{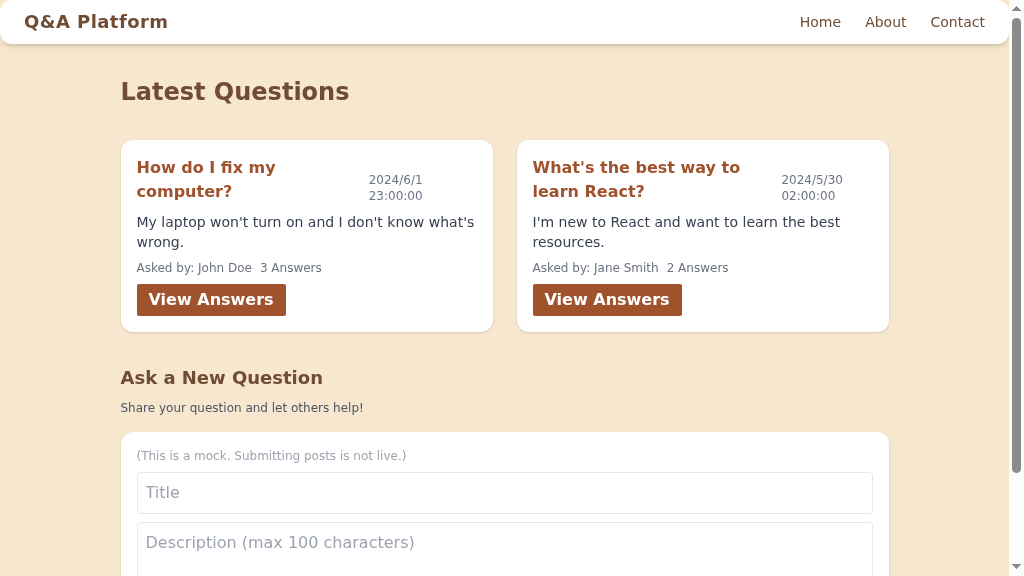} \\[0.6em] \small \textbf{WebGen-R1-7B (Ours)}
    \end{minipage}
\end{tabular}
\\ 
\midrule
\midrule
\textbf{Instruction:}
Please implement an email sending platform for sending emails. The platform should have functionalities for composing, sending, and managing emails. Users should be able to compose emails, select recipients, send emails, and manage sent emails. The platform should also have features such as email templates, attachment uploads, and recipient management. Use cornsilk for the background and peru for components.
\\ 
\midrule
\begin{tabular}{@{}p{0.33\textwidth} p{0.33\textwidth} p{0.33\textwidth}@{}}
    \begin{minipage}{\linewidth}
        \centering
        \includegraphics[width=\linewidth]{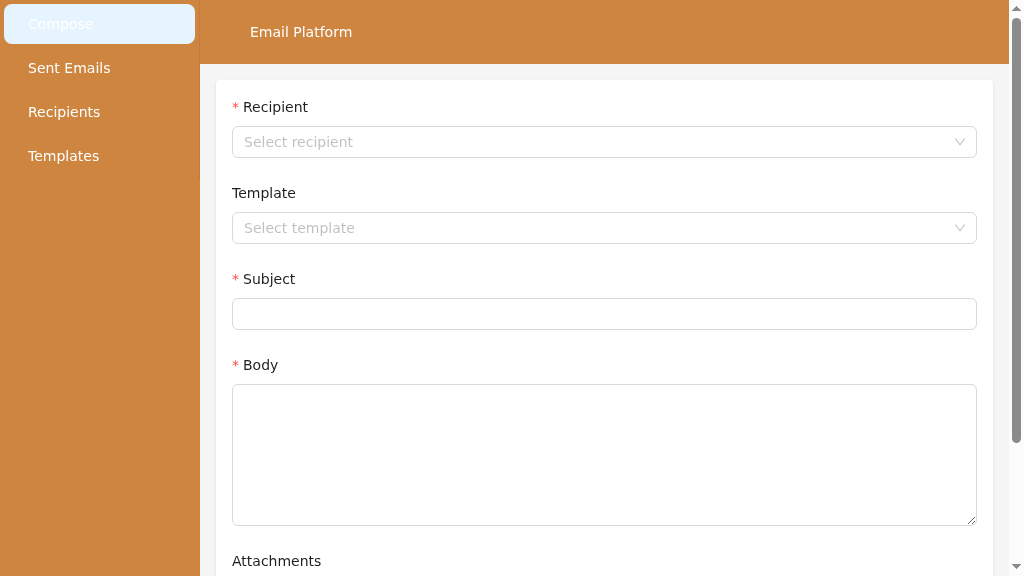} \\[0.6em] \small GPT-5
    \end{minipage} &
    \begin{minipage}{\linewidth}
        \centering
        \includegraphics[width=\linewidth]{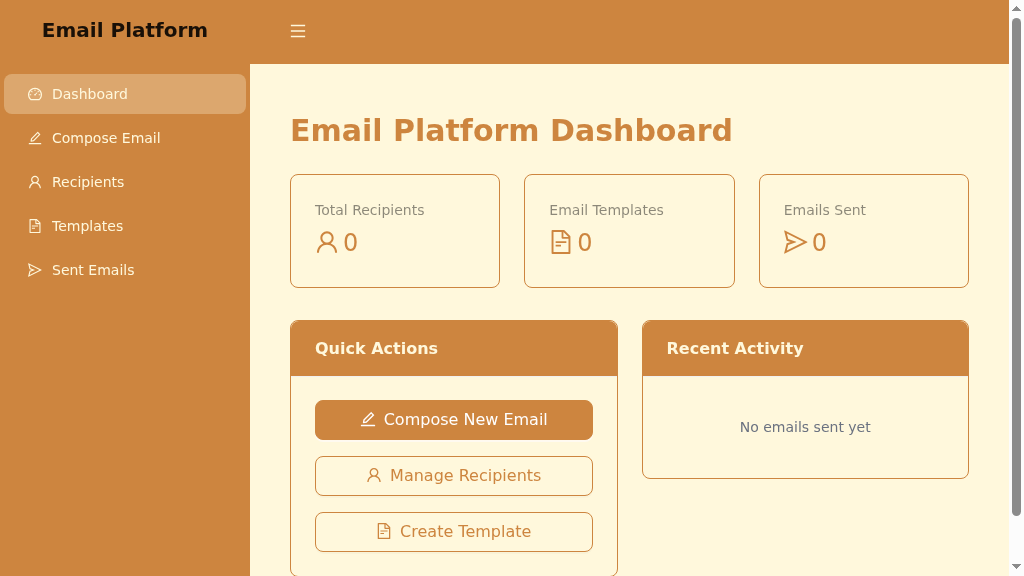} \\[0.6em] \small Claude-Sonnet-4
    \end{minipage} &
    \begin{minipage}{\linewidth}
        \centering
        \includegraphics[width=\linewidth]{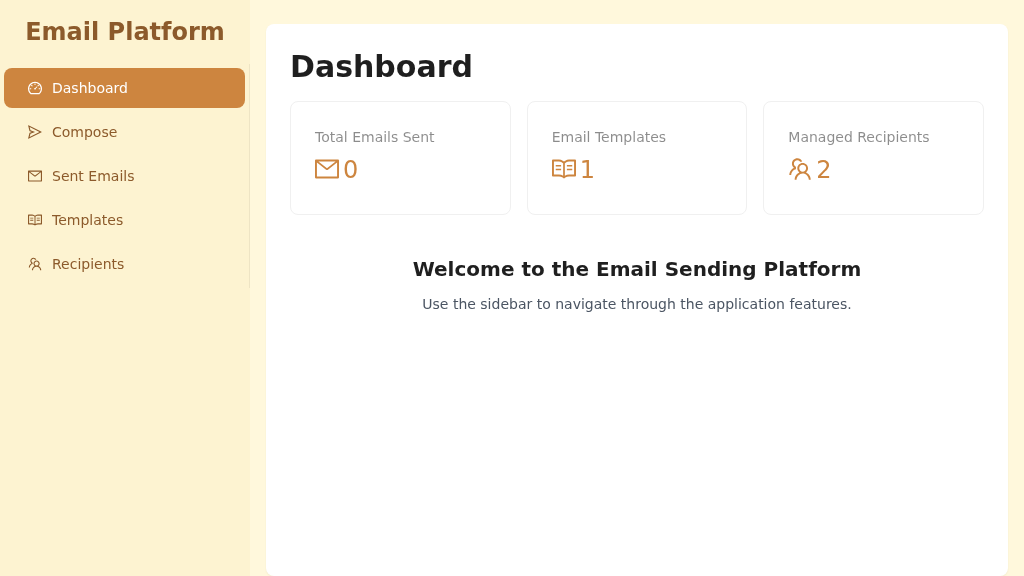} \\[0.6em] \small Gemini-2.5-Pro
    \end{minipage}
    \\[5em]
    \begin{minipage}{\linewidth}
        \centering
        \includegraphics[width=\linewidth]{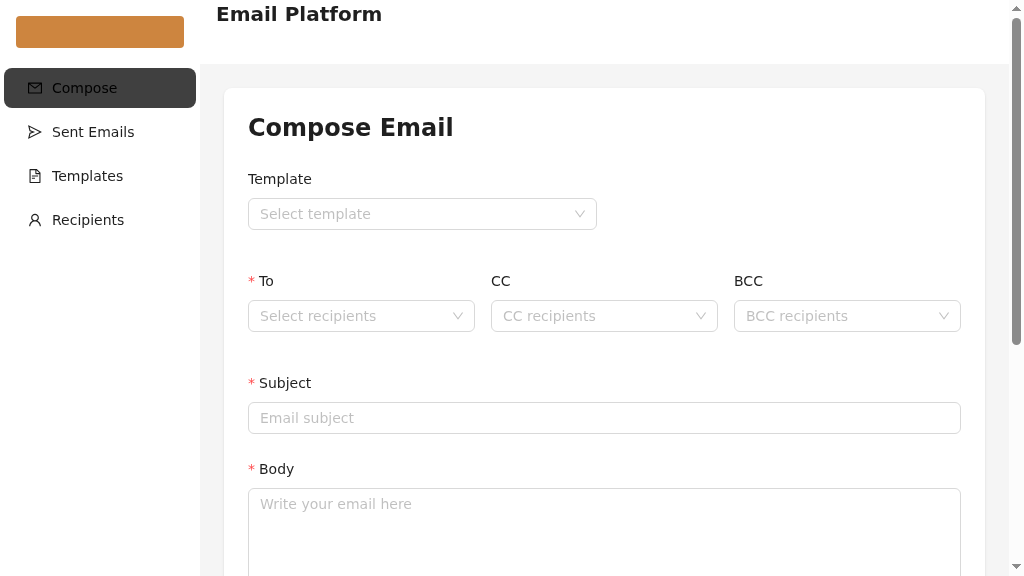} \\[0.6em] \small DeepSeek-R1
    \end{minipage} &
    \begin{minipage}{\linewidth}
        \centering
        \includegraphics[width=\linewidth]{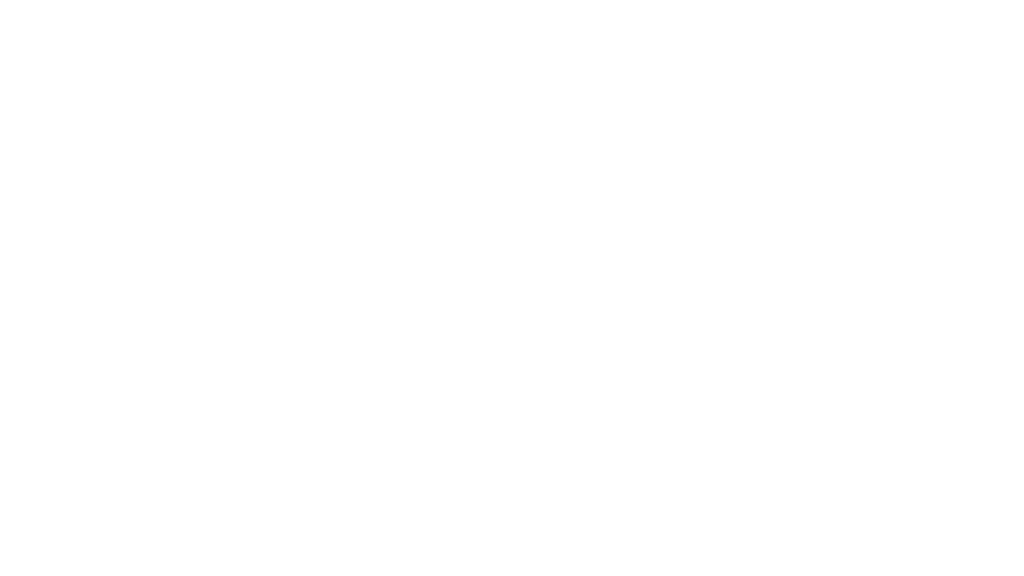} \\[0.6em] \small Qwen3-Coder-30B-A3B
    \end{minipage} &
    \begin{minipage}{\linewidth}
        \centering
        \includegraphics[width=\linewidth]{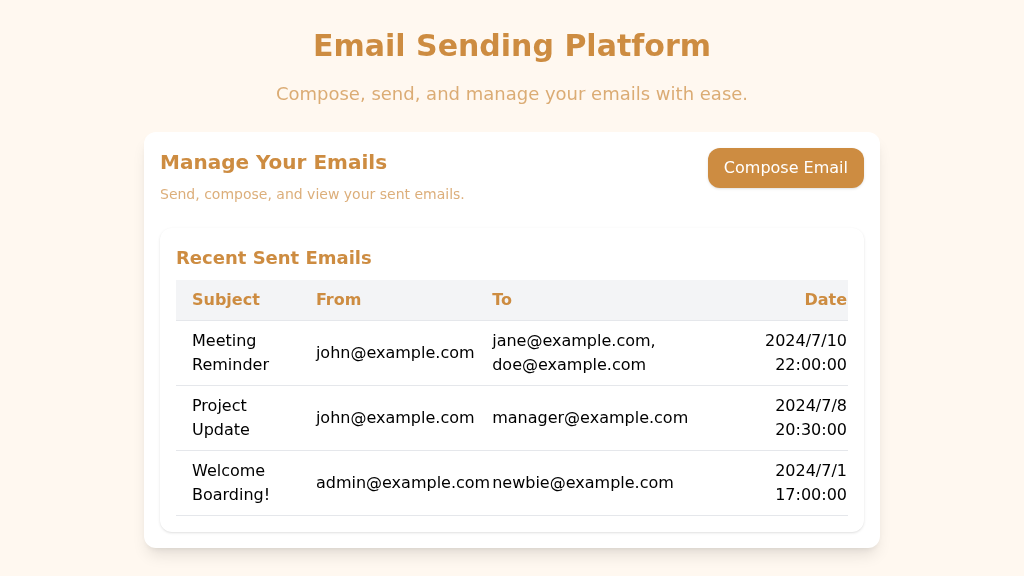} \\[0.6em] \small \textbf{WebGen-R1-7B (Ours)}
    \end{minipage}
\end{tabular}
\\
\bottomrule
\end{tabular}
\label{tab:webgen_bench_model_comparison_part3}
\end{table}

\begin{table}[!ht]
\centering
\caption{Comparison of websites generated by different models on two tasks from WebDev Arena.}
\setlength{\tabcolsep}{1pt} 
\begin{tabular}{@{}p{\textwidth}@{}}
\toprule
\textbf{Instruction:} 
A battle arena website that compare audio mp3 generate by 2 models, model A and model B. Users listen to these two audio files and vote for the best.
\\ 
\midrule
\begin{tabular}{@{}p{0.33\textwidth} p{0.33\textwidth} p{0.33\textwidth}@{}}
    \begin{minipage}{\linewidth}
        \centering
        \includegraphics[width=\linewidth]{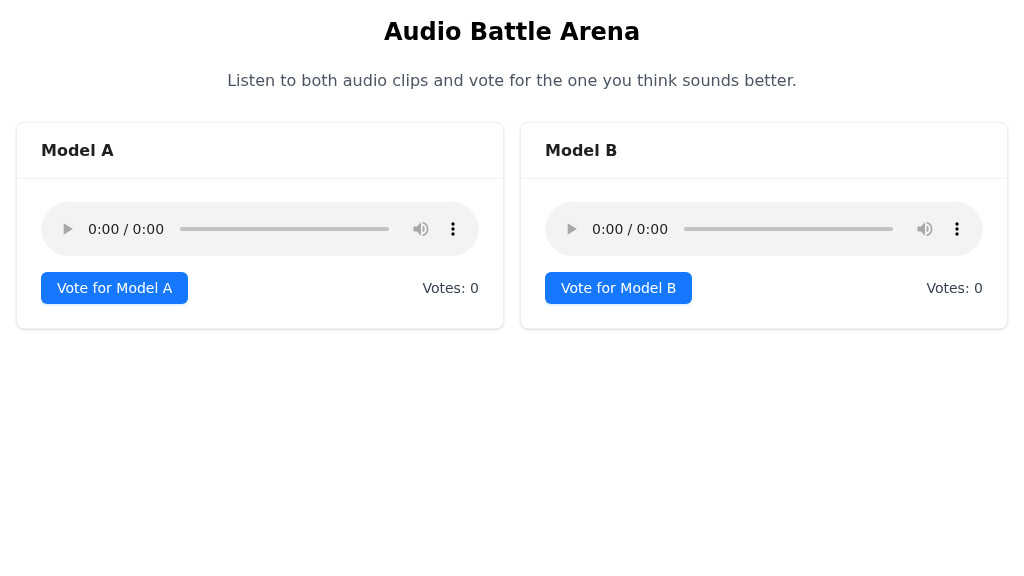} \\[0.6em] \small GPT-5
    \end{minipage} &
    \begin{minipage}{\linewidth}
        \centering
        \includegraphics[width=\linewidth]{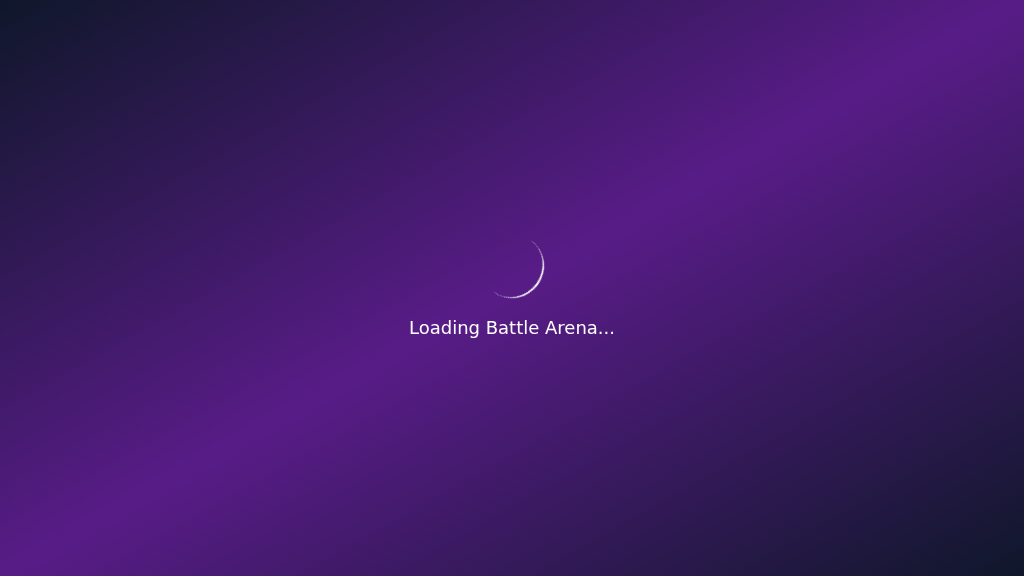} \\[0.6em] \small Claude-Sonnet-4
    \end{minipage} &
    \begin{minipage}{\linewidth}
        \centering
        \includegraphics[width=\linewidth]{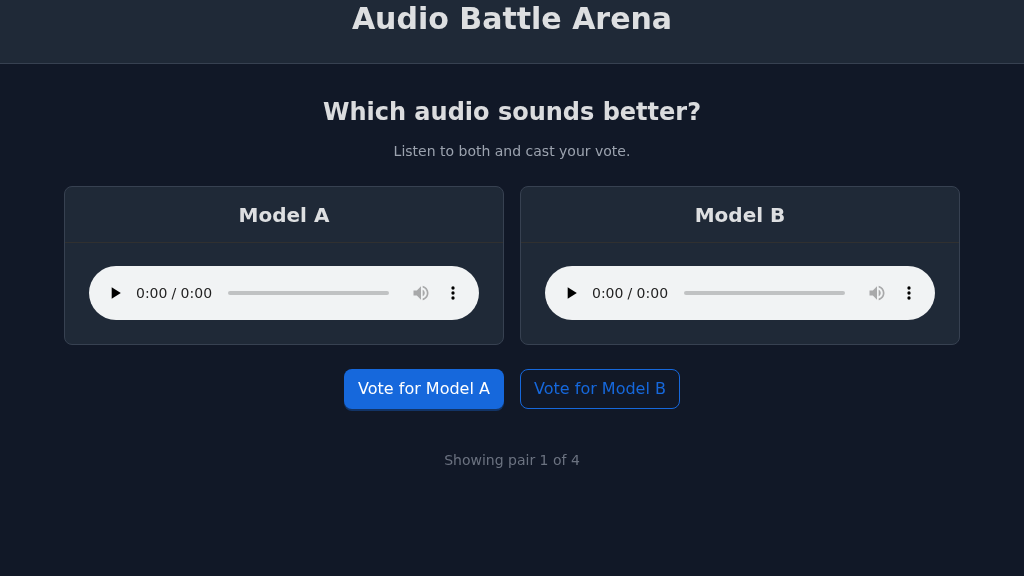} \\[0.6em] \small Gemini-2.5-Pro
    \end{minipage}
    \\[5em] 
    \begin{minipage}{\linewidth}
        \centering
        \includegraphics[width=\linewidth]{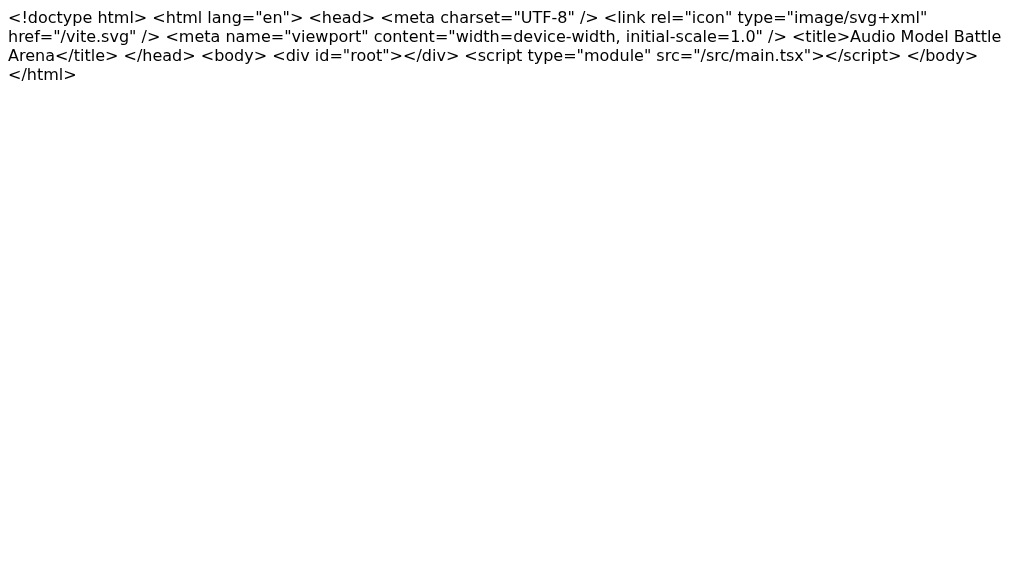} \\[0.6em] \small DeepSeek-R1
    \end{minipage} &
    \begin{minipage}{\linewidth}
        \centering
        \includegraphics[width=\linewidth]{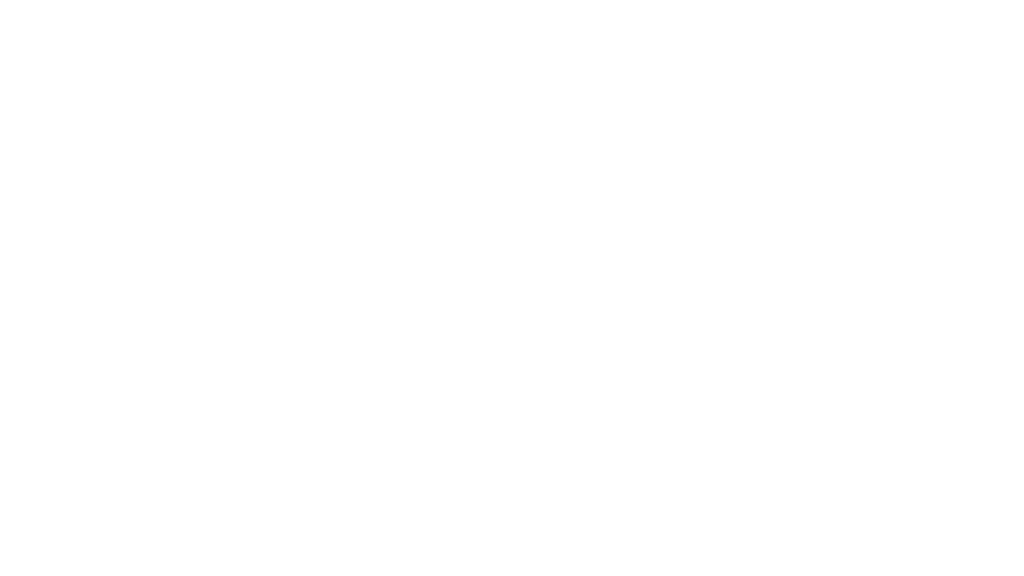} \\[0.6em] \small Qwen3-Coder-30B-A3B
    \end{minipage} &
    \begin{minipage}{\linewidth}
        \centering
        \includegraphics[width=\linewidth]{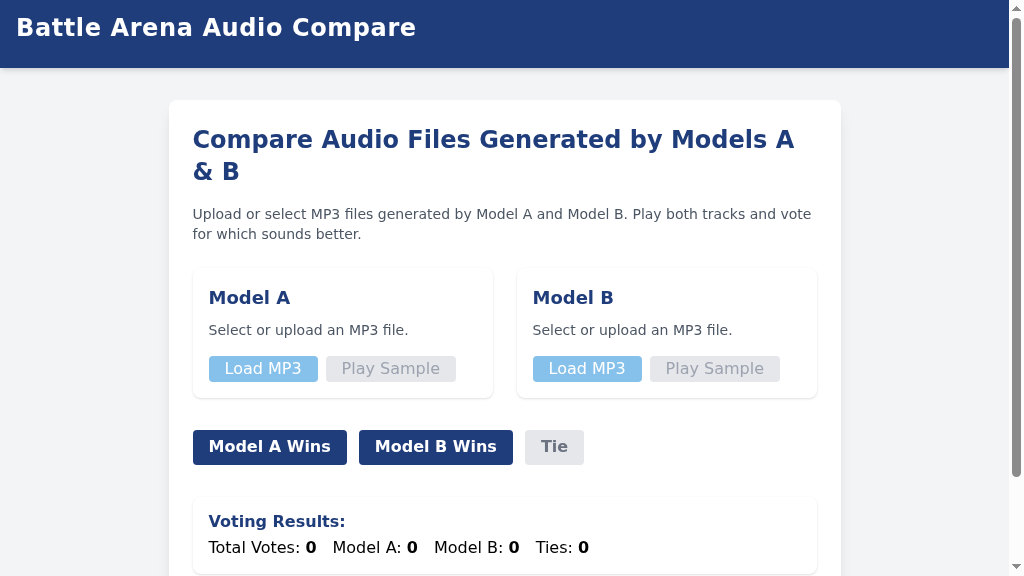} \\[0.6em] \small \textbf{WebGen-R1-7B (Ours)}
    \end{minipage}
\end{tabular}
\\ 
\midrule
\midrule
\textbf{Instruction:}
create me a Resume Page
\\ 
\midrule
\begin{tabular}{@{}p{0.33\textwidth} p{0.33\textwidth} p{0.33\textwidth}@{}}
    \begin{minipage}{\linewidth}
        \centering
        \includegraphics[width=\linewidth]{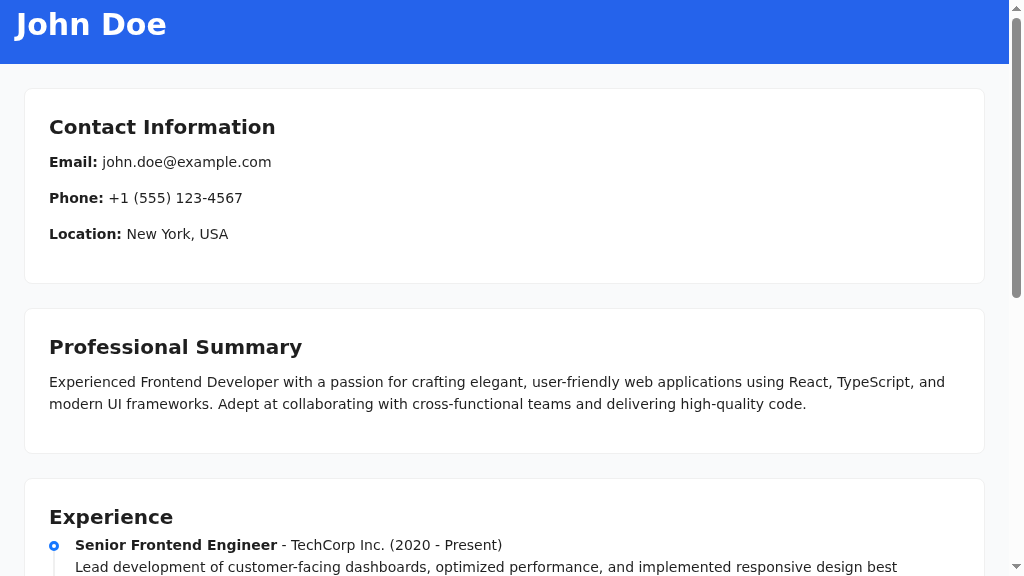} \\[0.6em] \small GPT-5
    \end{minipage} &
    \begin{minipage}{\linewidth}
        \centering
        \includegraphics[width=\linewidth]{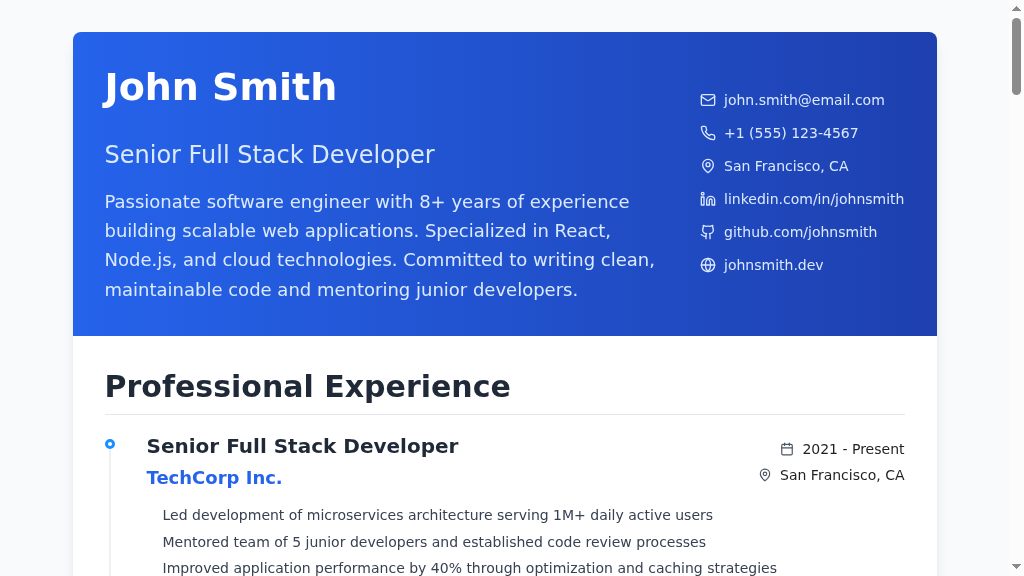} \\[0.6em] \small Claude-Sonnet-4
    \end{minipage} &
    \begin{minipage}{\linewidth}
        \centering
        \includegraphics[width=\linewidth]{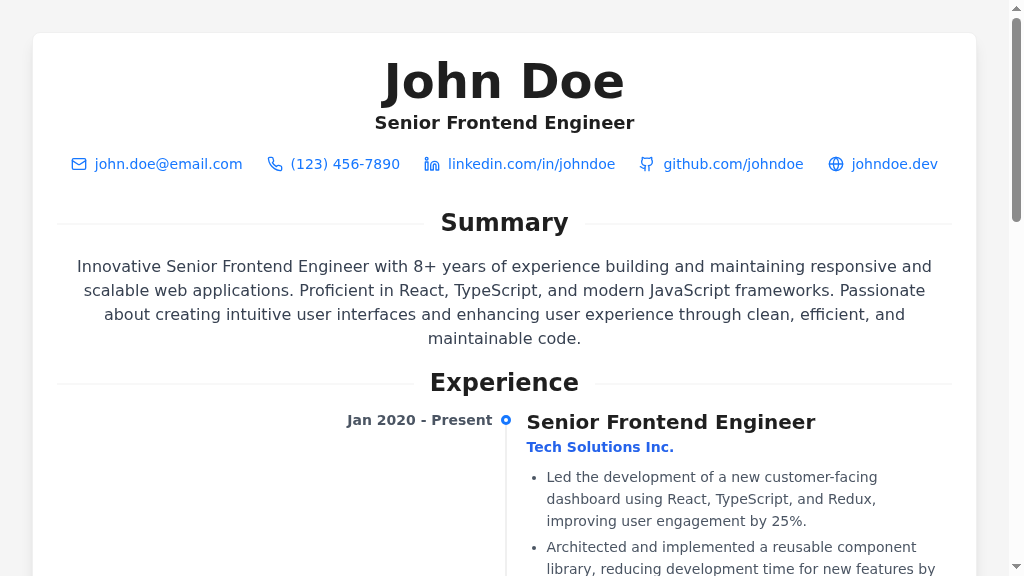} \\[0.6em] \small Gemini-2.5-Pro
    \end{minipage}
    \\[5em]
    \begin{minipage}{\linewidth}
        \centering
        \includegraphics[width=\linewidth]{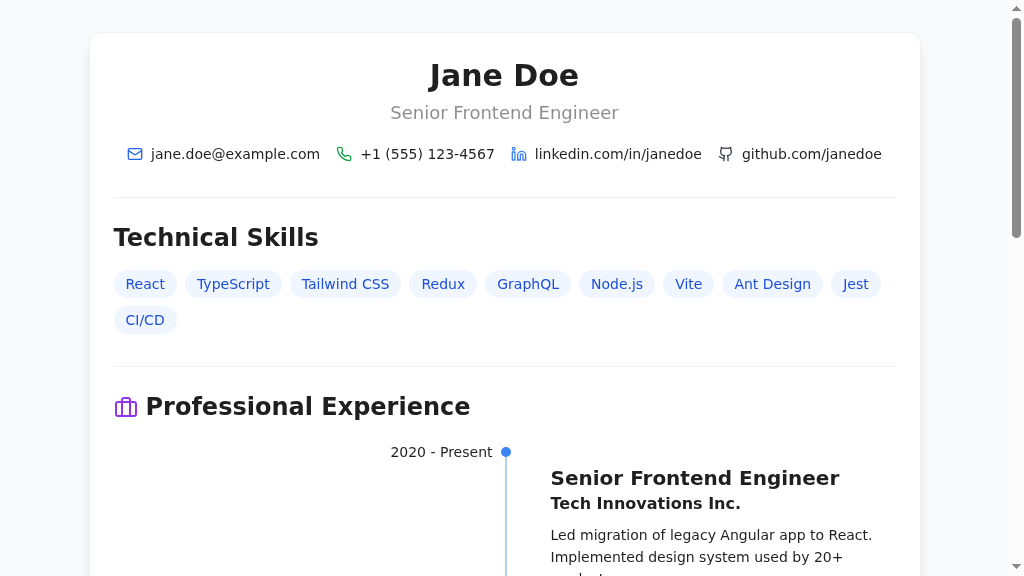} \\[0.6em] \small DeepSeek-R1
    \end{minipage} &
    \begin{minipage}{\linewidth}
        \centering
        \includegraphics[width=\linewidth]{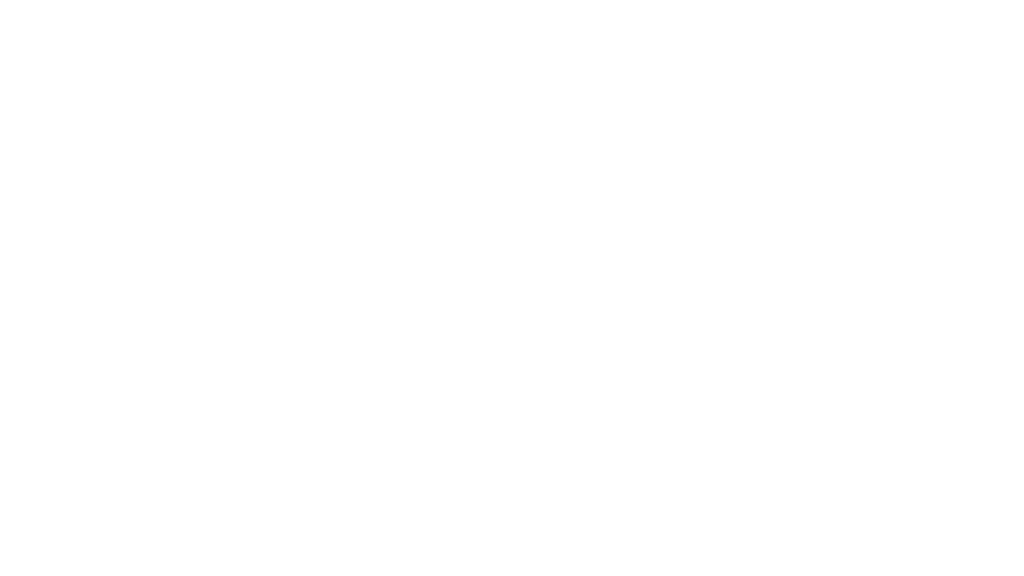} \\[0.6em] \small Qwen3-Coder-30B-A3B
    \end{minipage} &
    \begin{minipage}{\linewidth}
        \centering
        \includegraphics[width=\linewidth]{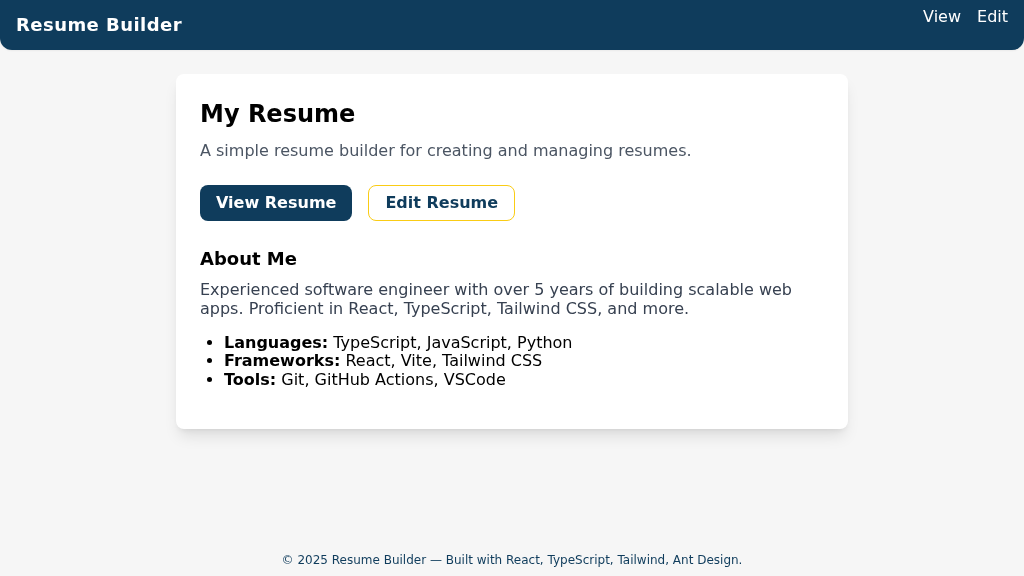} \\[0.6em] \small \textbf{WebGen-R1-7B (Ours)}
    \end{minipage}
\end{tabular}
\\
\bottomrule
\end{tabular}
\label{tab:webdev_arena_model_comparison_part1}
\end{table}

\begin{table}[!ht]
\centering
\caption{Comparison of websites generated by different models on two tasks from WebDev Arena.}
\setlength{\tabcolsep}{1pt} 
\begin{tabular}{@{}p{\textwidth}@{}}
\toprule
\textbf{Instruction:} 
Make a website that fetches data (joke) from an external API and displays it on the screen using react for use example.
\\ 
\midrule
\begin{tabular}{@{}p{0.33\textwidth} p{0.33\textwidth} p{0.33\textwidth}@{}}
    \begin{minipage}{\linewidth}
        \centering
        \includegraphics[width=\linewidth]{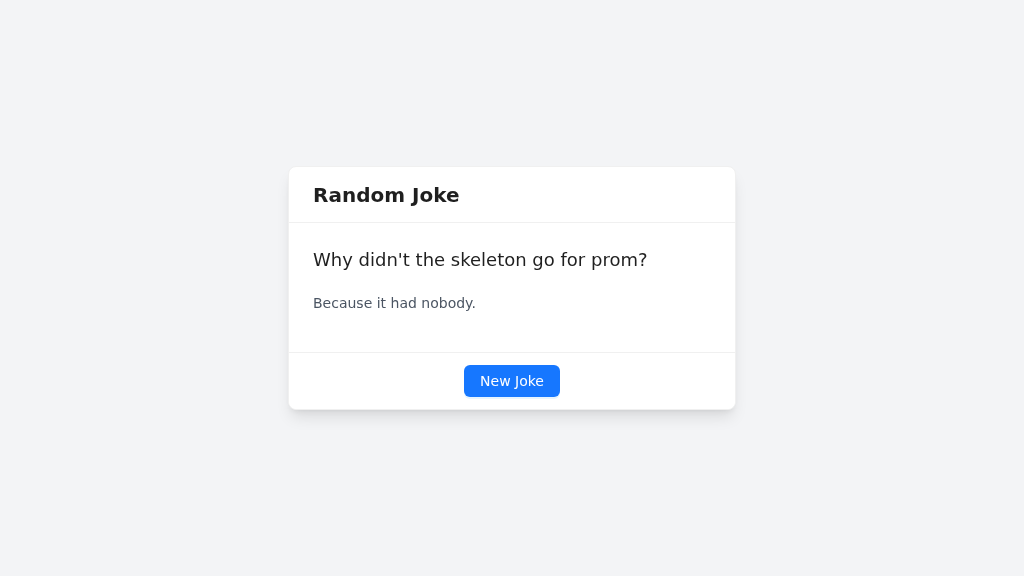} \\[0.6em] \small GPT-5
    \end{minipage} &
    \begin{minipage}{\linewidth}
        \centering
        \includegraphics[width=\linewidth]{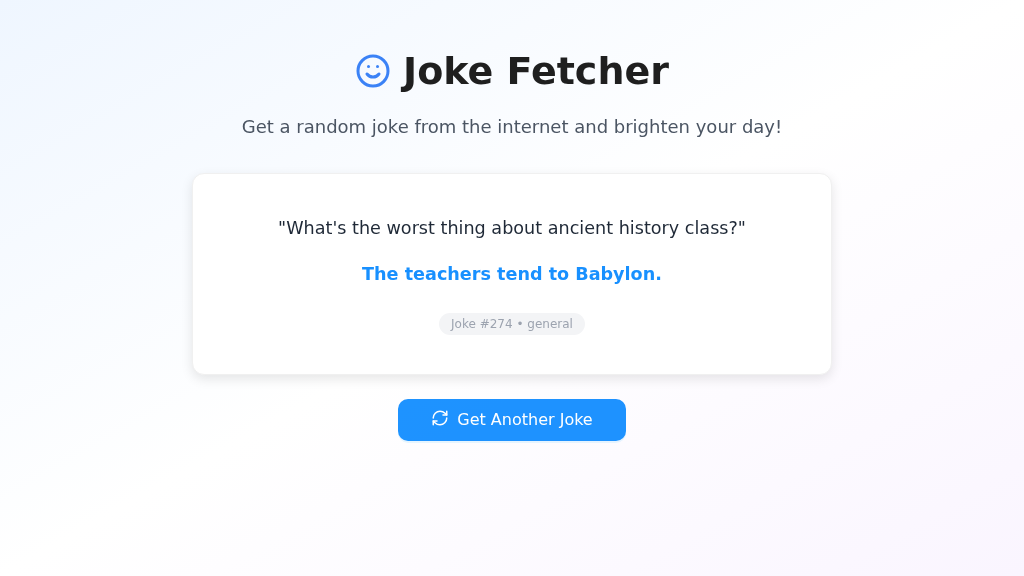} \\[0.6em] \small Claude-Sonnet-4
    \end{minipage} &
    \begin{minipage}{\linewidth}
        \centering
        \includegraphics[width=\linewidth]{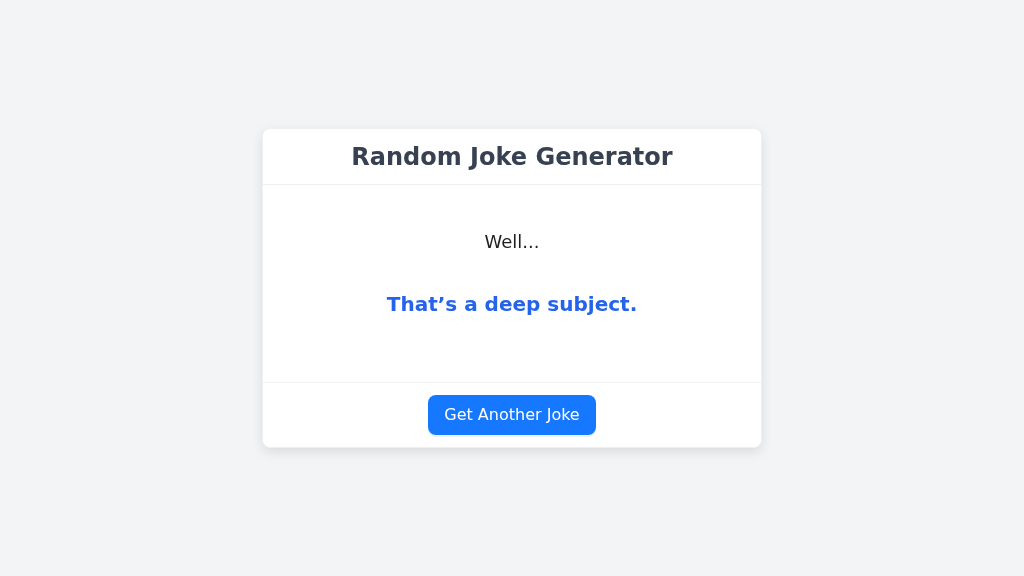} \\[0.6em] \small Gemini-2.5-Pro
    \end{minipage}
    \\[5em]
    \begin{minipage}{\linewidth}
        \centering
        \includegraphics[width=\linewidth]{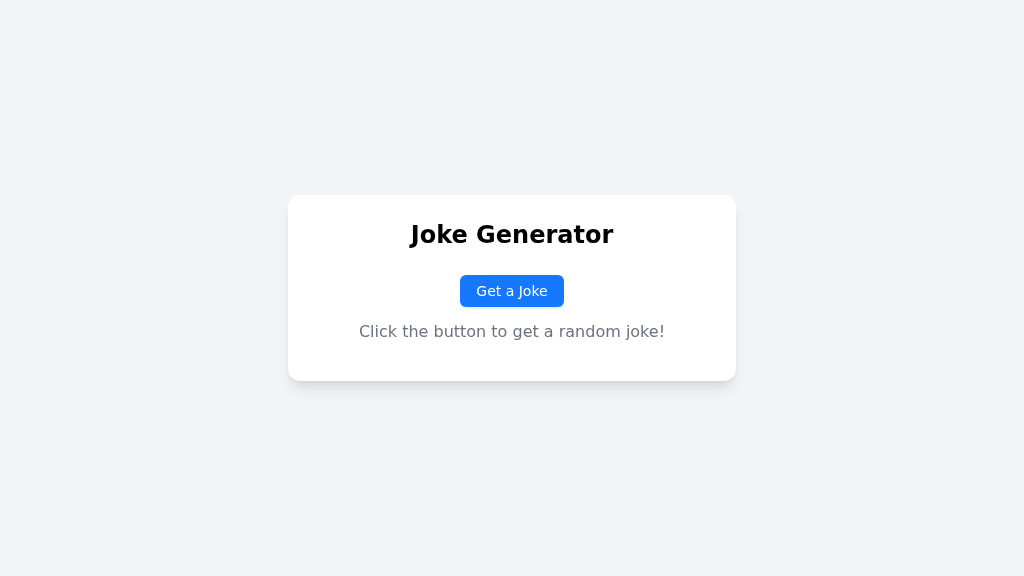} \\[0.6em] \small DeepSeek-R1
    \end{minipage} &
    \begin{minipage}{\linewidth}
        \centering
        \includegraphics[width=\linewidth]{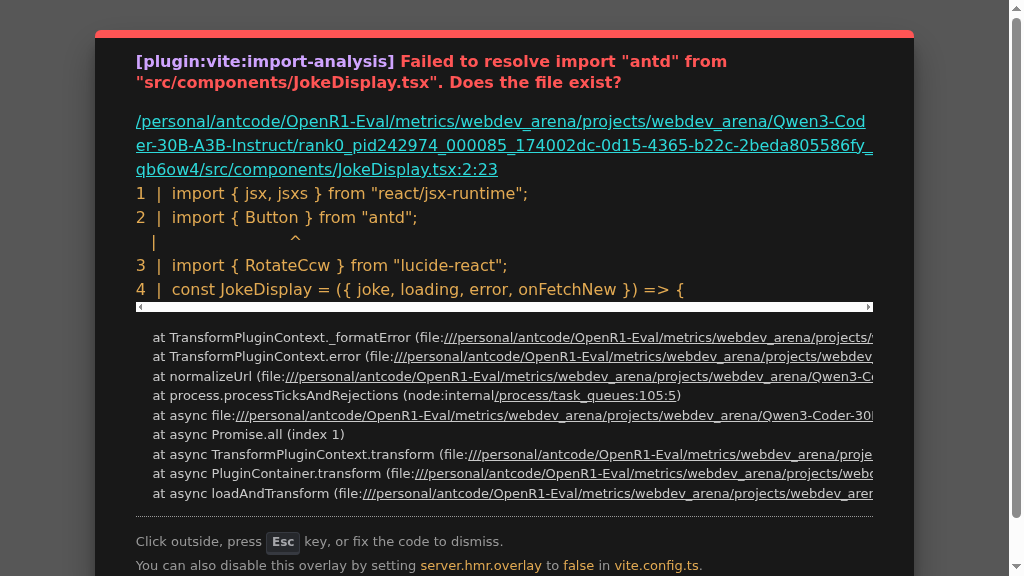} \\[0.6em] \small Qwen3-Coder-30B-A3B
    \end{minipage} &
    \begin{minipage}{\linewidth}
        \centering
        \includegraphics[width=\linewidth]{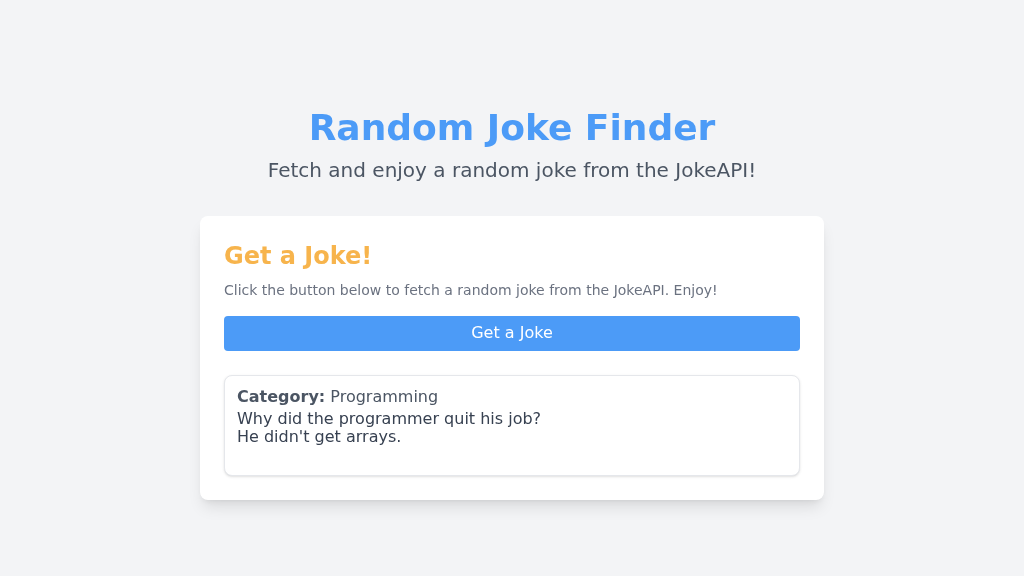} \\[0.6em] \small \textbf{WebGen-R1-7B (Ours)}
    \end{minipage}
\end{tabular}
\\ 
\midrule
\midrule
\textbf{Instruction:}
Design a cryptocurrency tracker with real-time price updates and portfolio tracking. Focus on clear presentation of price charts and user-friendly transaction inputs.
\\ 
\midrule
\begin{tabular}{@{}p{0.33\textwidth} p{0.33\textwidth} p{0.33\textwidth}@{}}
    \begin{minipage}{\linewidth}
        \centering
        \includegraphics[width=\linewidth]{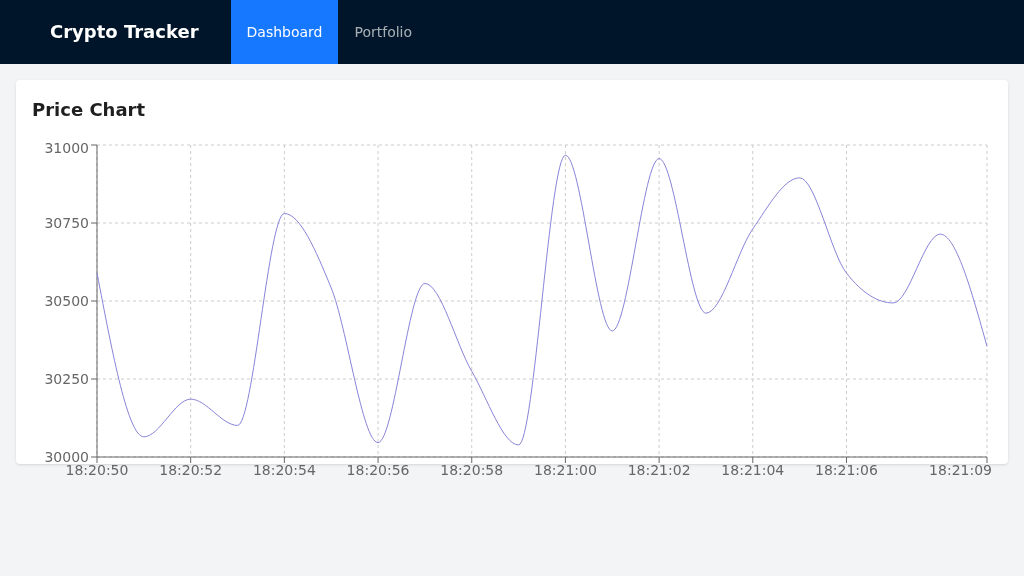} \\[0.6em] \small GPT-5
    \end{minipage} &
    \begin{minipage}{\linewidth}
        \centering
        \includegraphics[width=\linewidth]{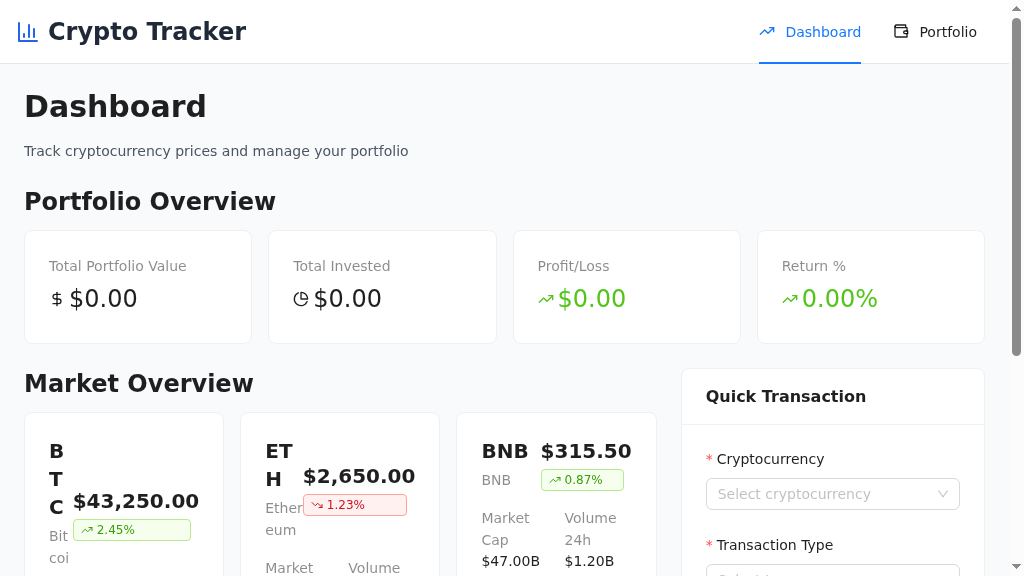} \\[0.6em] \small Claude-Sonnet-4
    \end{minipage} &
    \begin{minipage}{\linewidth}
        \centering
        \includegraphics[width=\linewidth]{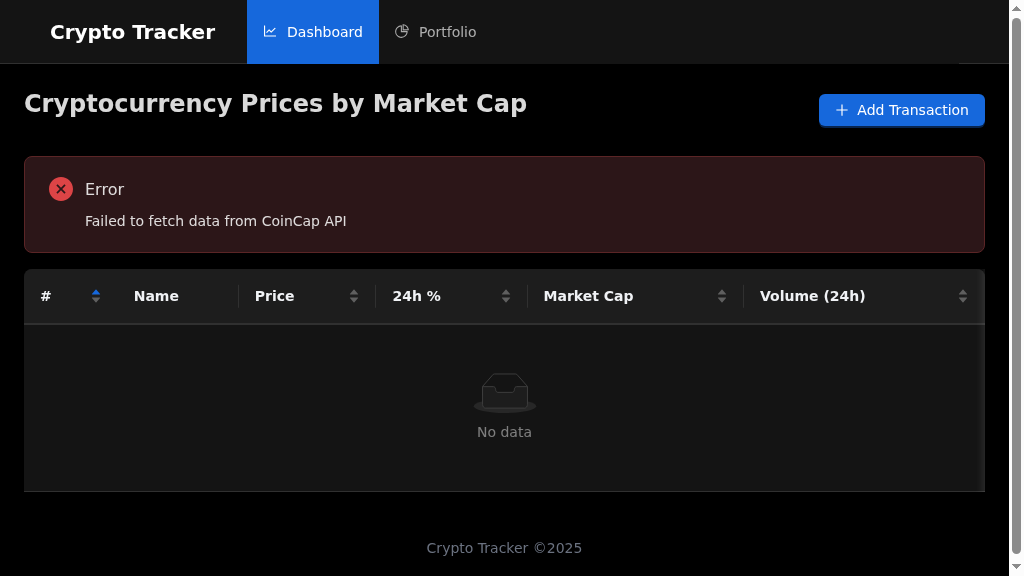} \\[0.6em] \small Gemini-2.5-Pro
    \end{minipage}
    \\[5em]
    \begin{minipage}{\linewidth}
        \centering
        \includegraphics[width=\linewidth]{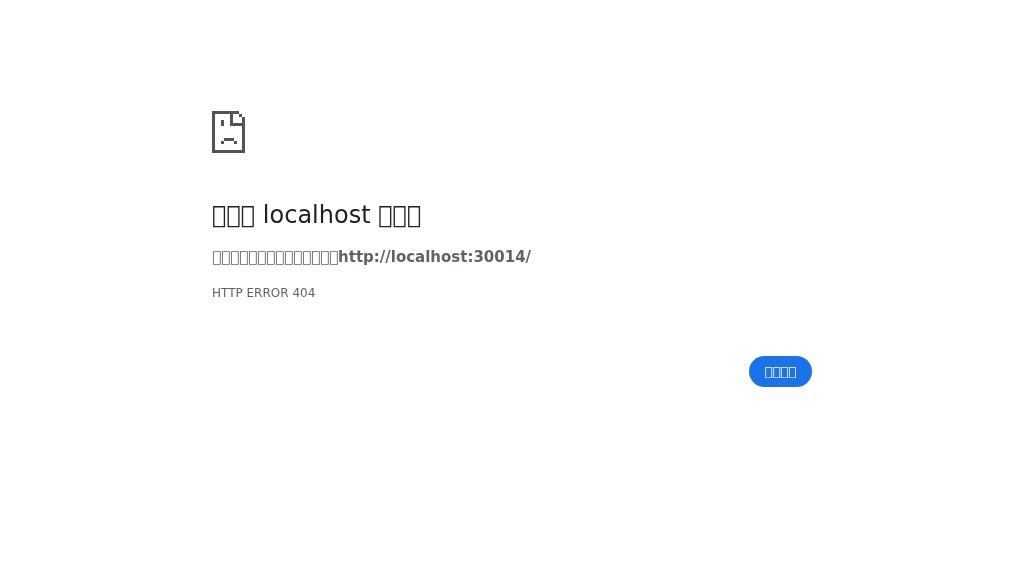} \\[0.6em] \small DeepSeek-R1
    \end{minipage} &
    \begin{minipage}{\linewidth}
        \centering
        \includegraphics[width=\linewidth]{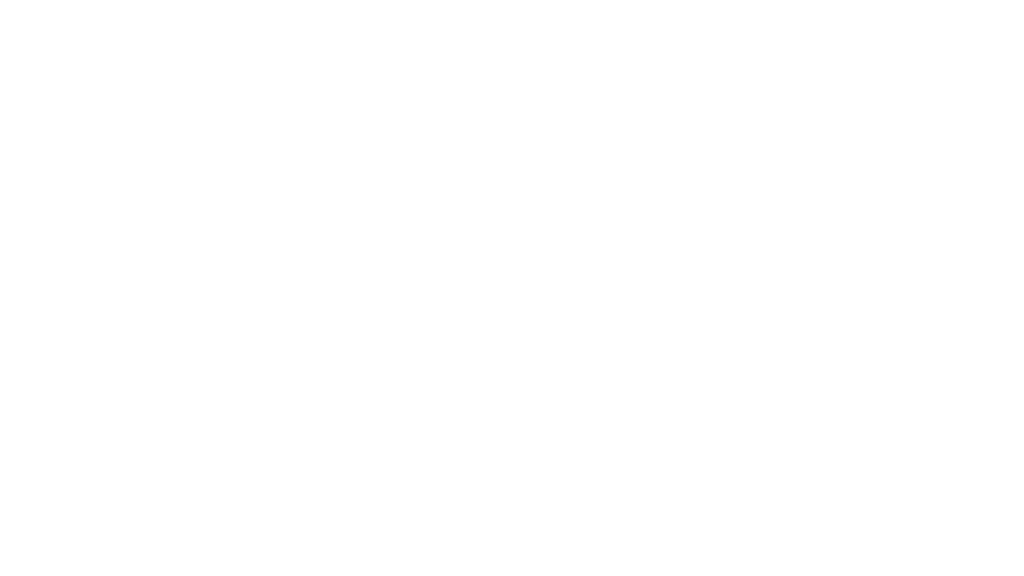} \\[0.6em] \small Qwen3-Coder-30B-A3B
    \end{minipage} &
    \begin{minipage}{\linewidth}
        \centering
        \includegraphics[width=\linewidth]{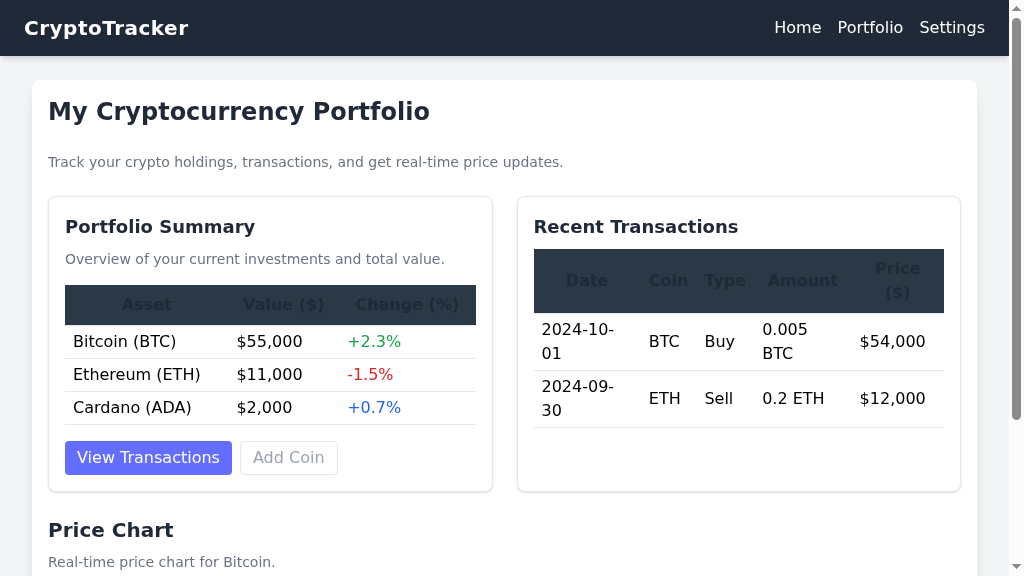} \\[0.6em] \small \textbf{WebGen-R1-7B (Ours)}
    \end{minipage}
\end{tabular}
\\
\bottomrule
\end{tabular}
\label{tab:webdev_arena_model_comparison_part4}
\end{table}

\begin{table}[!ht]
\centering
\caption{Comparison of websites generated by different models on two tasks from WebDev Arena.}
\setlength{\tabcolsep}{1pt} 
\begin{tabular}{@{}p{\textwidth}@{}}
\toprule
\textbf{Instruction:} 
Design a job board with filters for location, salary, and job type. Create an appealing layout for job postings, highlighting key details. with all of stuff.
\\ 
\midrule
\begin{tabular}{@{}p{0.33\textwidth} p{0.33\textwidth} p{0.33\textwidth}@{}}
    \begin{minipage}{\linewidth}
        \centering
        \includegraphics[width=\linewidth]{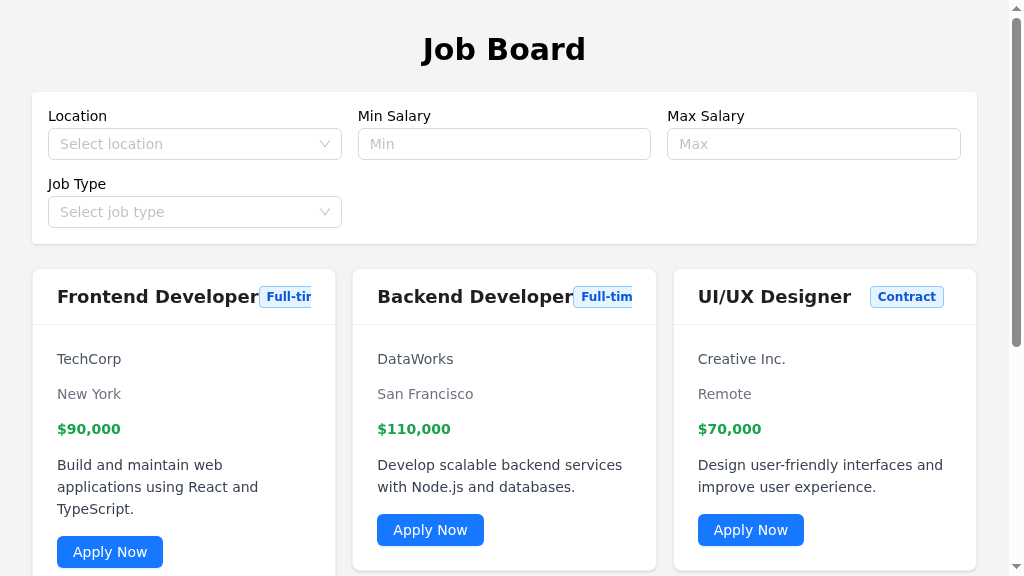} \\[0.6em] \small GPT-5
    \end{minipage} &
    \begin{minipage}{\linewidth}
        \centering
        \includegraphics[width=\linewidth]{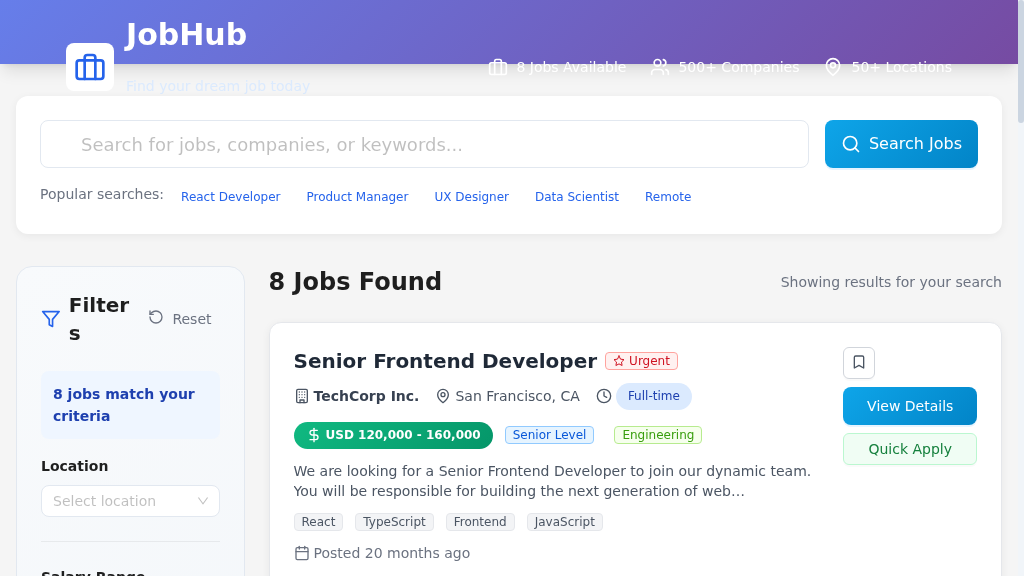} \\[0.6em] \small Claude-Sonnet-4
    \end{minipage} &
    \begin{minipage}{\linewidth}
        \centering
        \includegraphics[width=\linewidth]{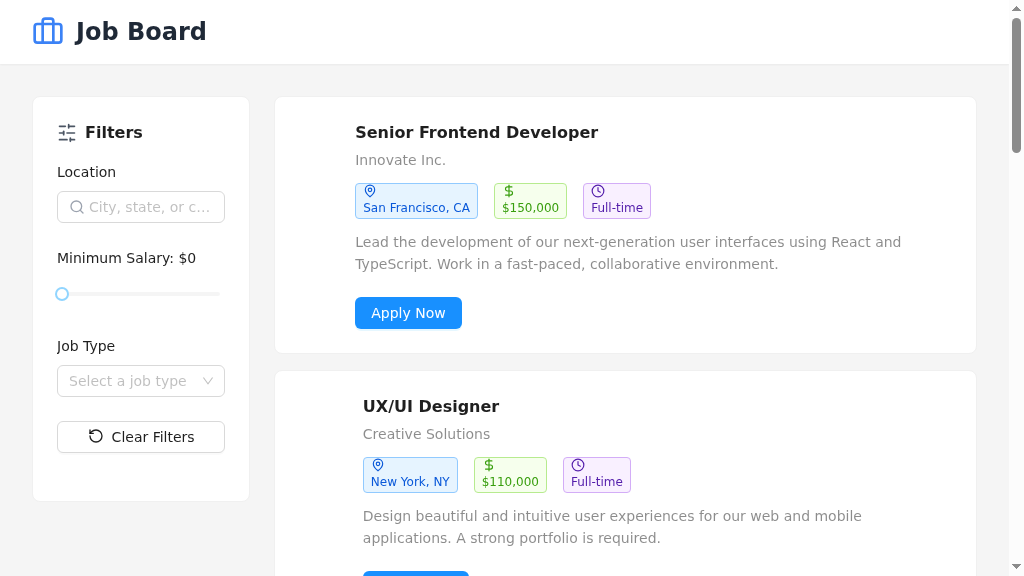} \\[0.6em] \small Gemini-2.5-Pro
    \end{minipage}
    \\[5em] 
    \begin{minipage}{\linewidth}
        \centering
        \includegraphics[width=\linewidth]{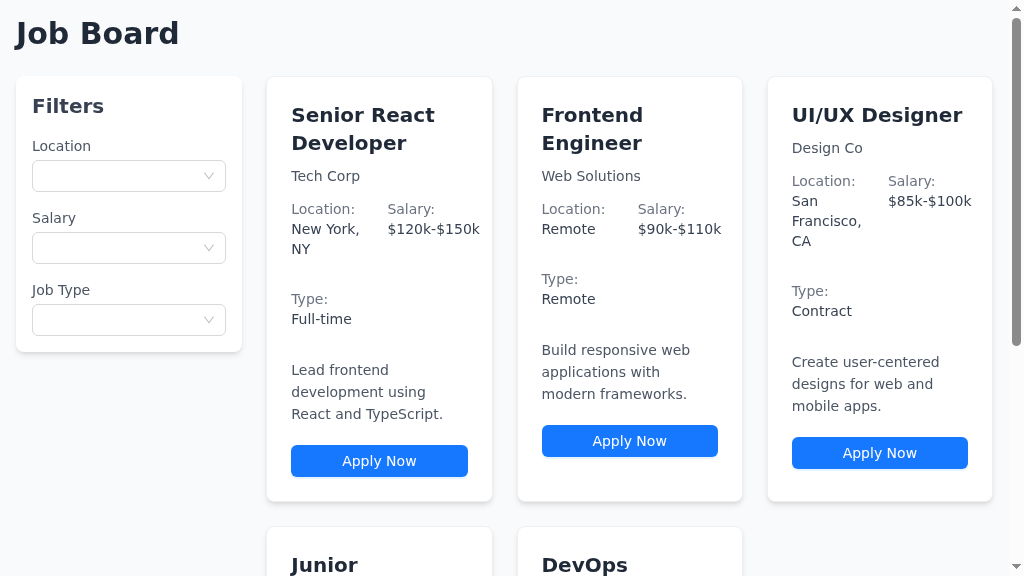} \\[0.6em] \small Qwen3-32B
    \end{minipage} &
    \begin{minipage}{\linewidth}
        \centering
        \includegraphics[width=\linewidth]{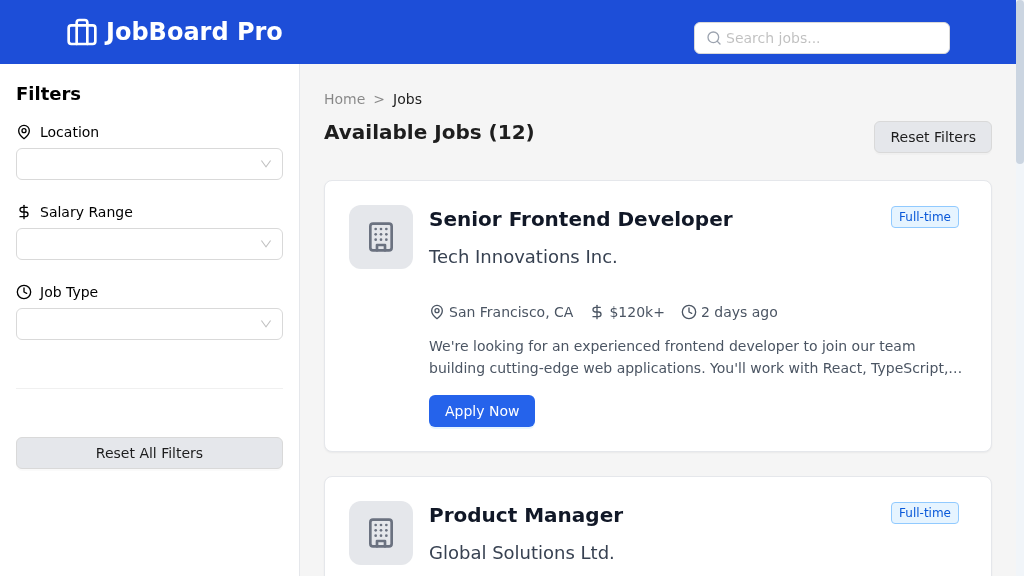} \\[0.6em] \small Qwen3-Coder-30B-A3B
    \end{minipage} &
    \begin{minipage}{\linewidth}
        \centering
        \includegraphics[width=\linewidth]{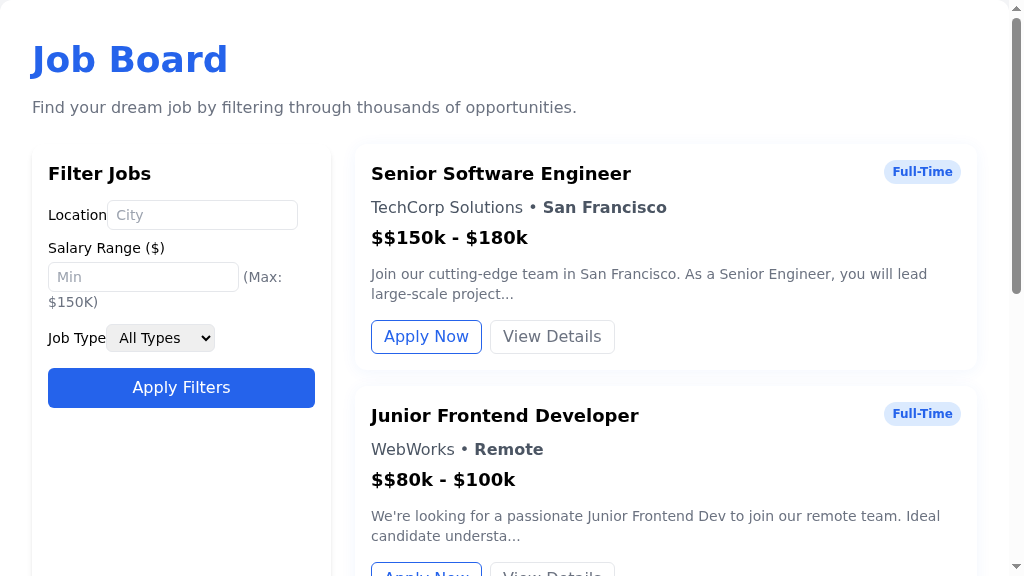} \\[0.6em] \small \textbf{WebGen-R1-7B (Ours)}
    \end{minipage}
\end{tabular}
\\ 
\midrule
\midrule
\textbf{Instruction:}
clone of warrior cat website
\\ 
\midrule
\begin{tabular}{@{}p{0.33\textwidth} p{0.33\textwidth} p{0.33\textwidth}@{}}
    \begin{minipage}{\linewidth}
        \centering
        \includegraphics[width=\linewidth]{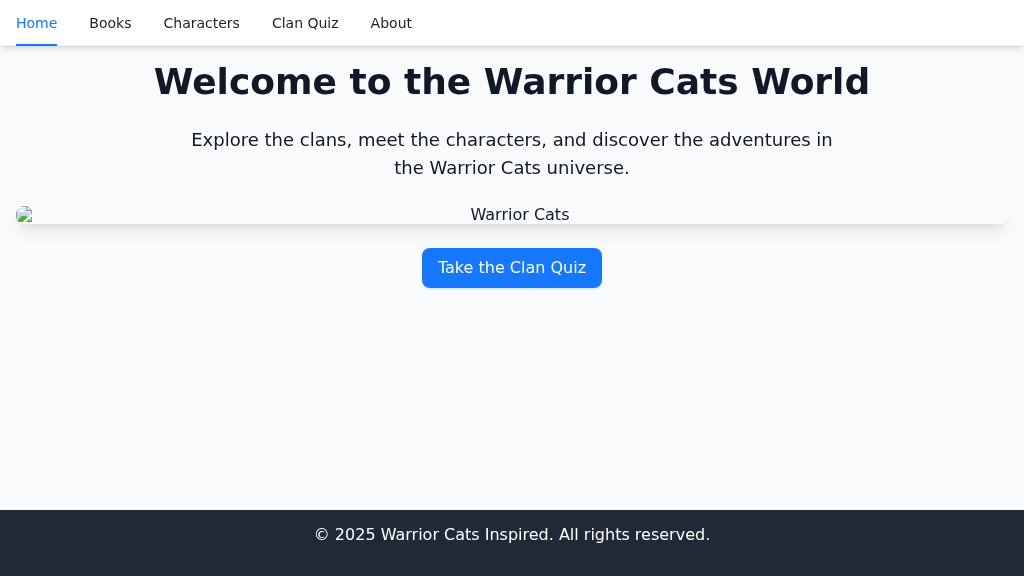} \\[0.6em] \small GPT-5
    \end{minipage} &
    \begin{minipage}{\linewidth}
        \centering
        \includegraphics[width=\linewidth]{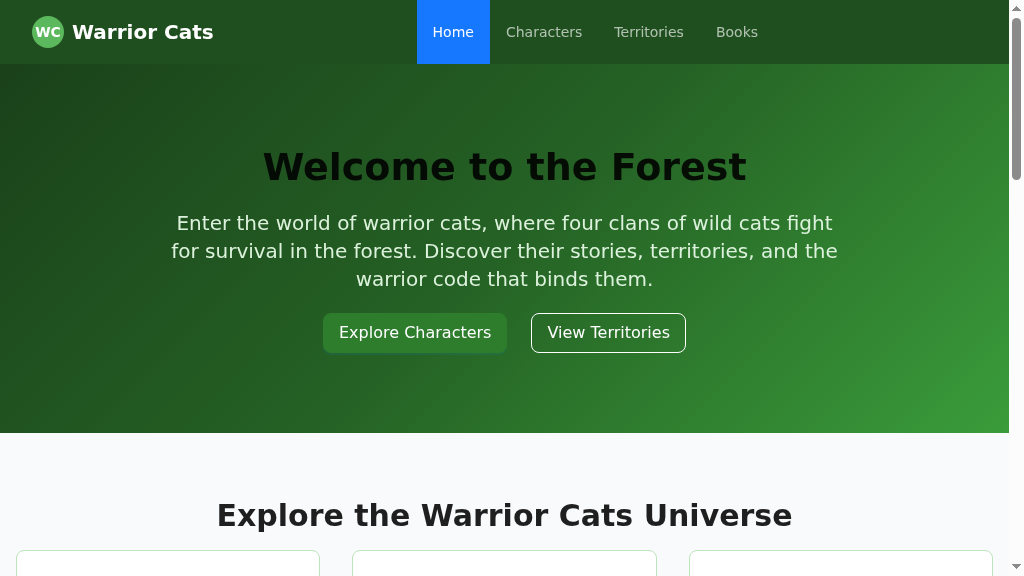} \\[0.6em] \small Claude-Sonnet-4
    \end{minipage} &
    \begin{minipage}{\linewidth}
        \centering
        \includegraphics[width=\linewidth]{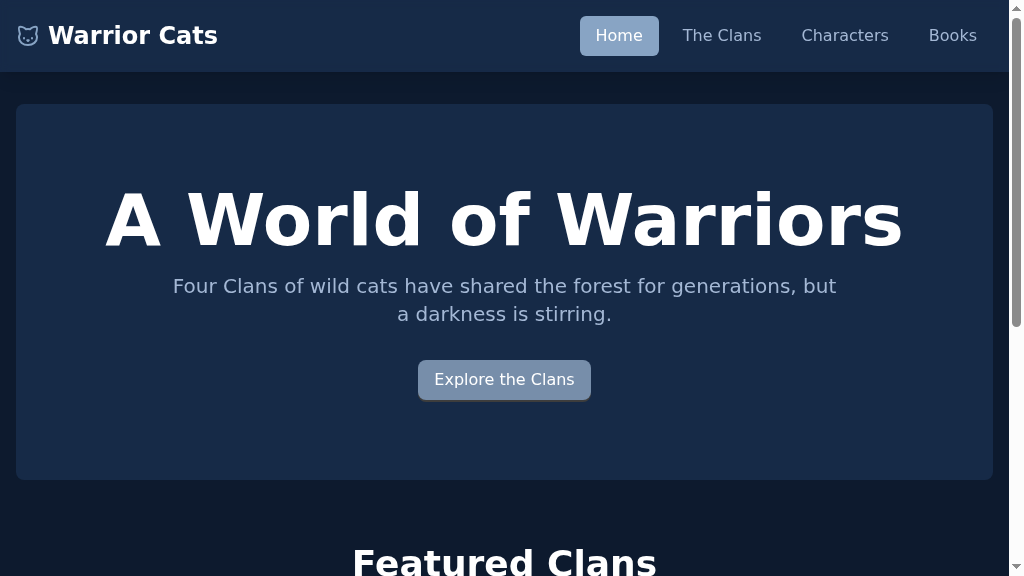} \\[0.6em] \small Gemini-2.5-Pro
    \end{minipage}
    \\[5em]
    \begin{minipage}{\linewidth}
        \centering
        \includegraphics[width=\linewidth]{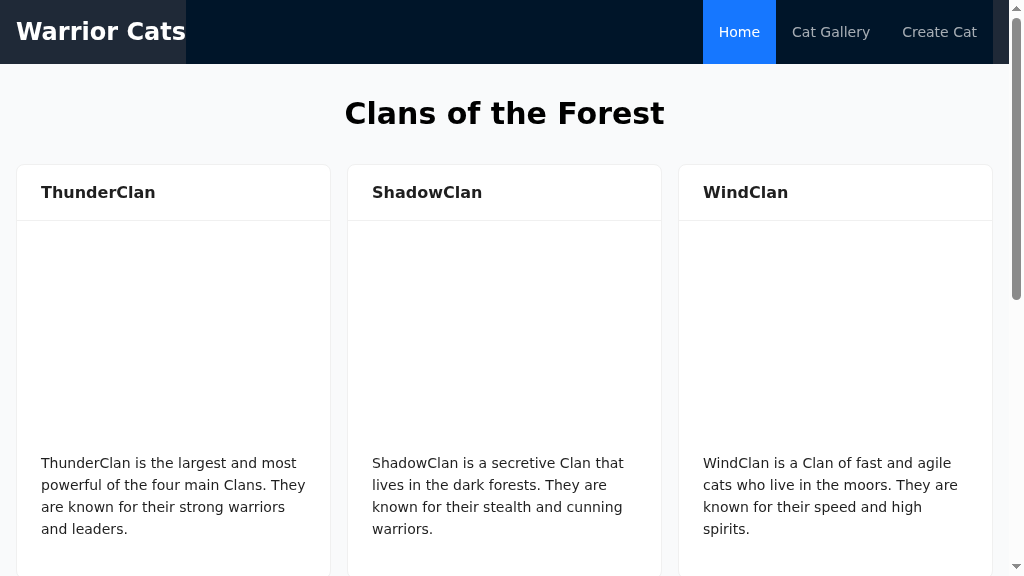} \\[0.6em] \small Qwen3-32B
    \end{minipage} &
    \begin{minipage}{\linewidth}
        \centering
        \includegraphics[width=\linewidth]{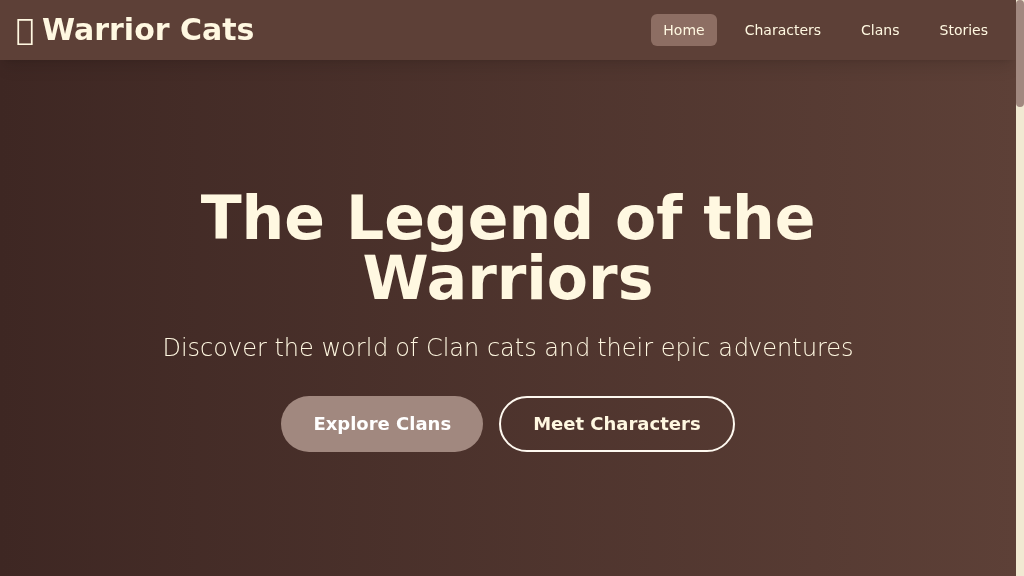} \\[0.6em] \small Qwen3-Coder-30B-A3B
    \end{minipage} &
    \begin{minipage}{\linewidth}
        \centering
        \includegraphics[width=\linewidth]{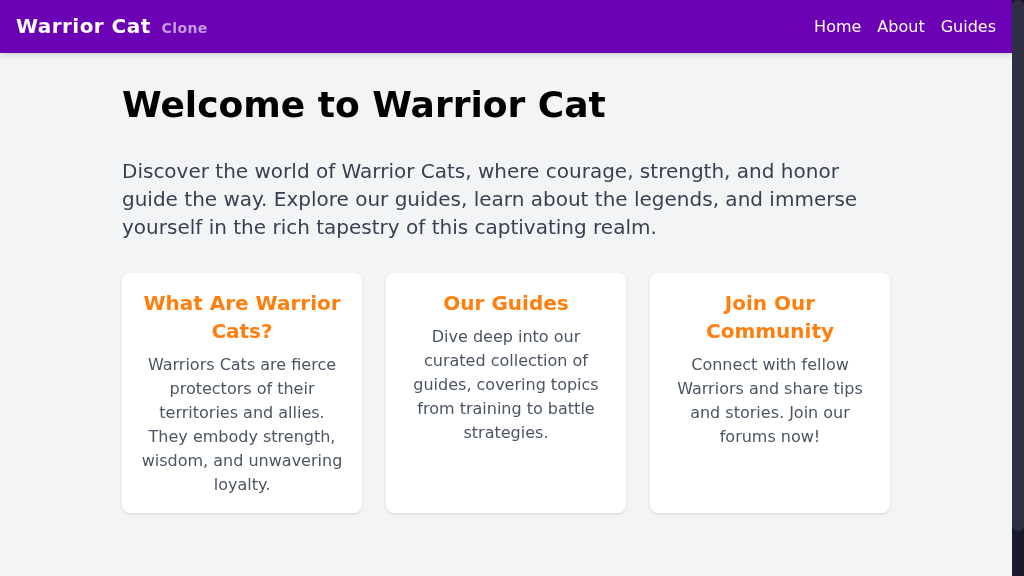} \\[0.6em] \small \textbf{WebGen-R1-7B (Ours)}
    \end{minipage}
\end{tabular}
\\
\bottomrule
\end{tabular}
\label{tab:webdev_arena_model_comparison_part6}
\end{table}

\begin{table}[!ht]
\centering
\caption{UI agent testing processes resulting in ``YES''. The top and bottom rows show the UI verification tasks for two different websites, respectively.}
\setlength{\tabcolsep}{1pt} 
\begin{tabular}{@{}p{\textwidth}@{}}
\toprule
\textbf{Task:} Verify that the website uses `honeydew' as the background color and `dark olive green' as the component color, as specified in the design requirements. \\
\textbf{Expected Result:} The website has a background color of `honeydew' and components (such as buttons, cards, headers, etc.) are styled with the color `dark olive green', accurately reflecting the design instruction.
\\ 
\midrule
\begin{tabular}{@{}p{0.33\textwidth} p{0.33\textwidth} p{0.33\textwidth}@{}}
    \begin{minipage}{\linewidth}
        {\centering
        \includegraphics[width=\linewidth]{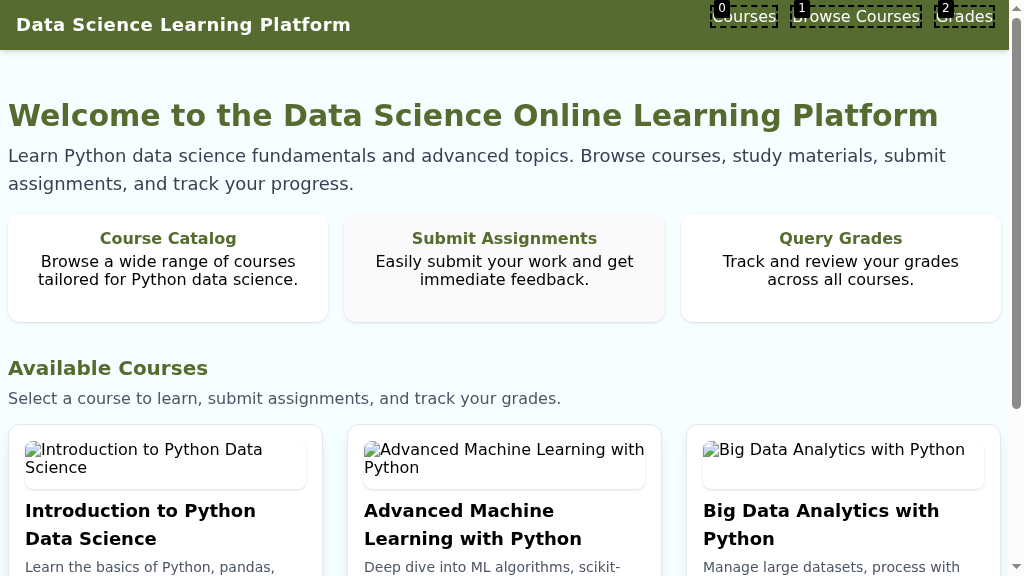}\par} 
        \vspace{0.2em} 
    \end{minipage} 
    &
    \begin{minipage}{\linewidth}
        {\centering
        \includegraphics[width=\linewidth]{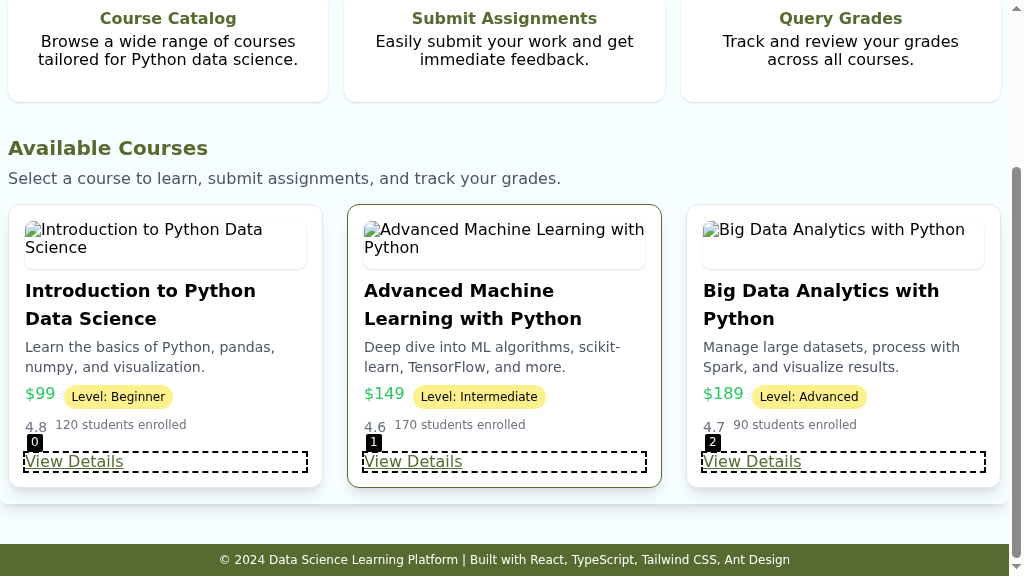} \par} 
        \vspace{0.2em} 
    \end{minipage} 
    &
    \begin{minipage}{\linewidth}
        {\centering
        \includegraphics[width=\linewidth]{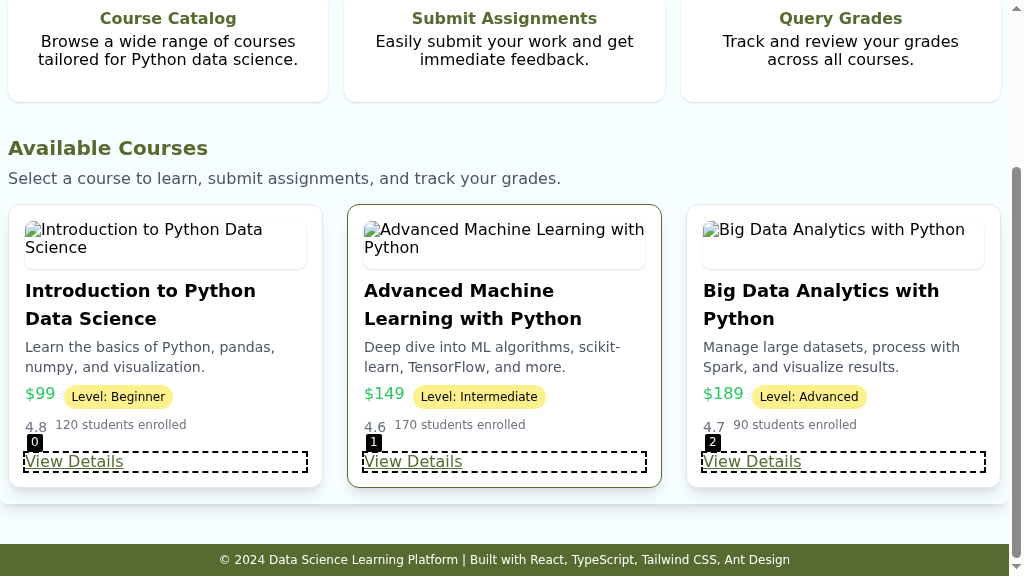} \par} 
        \vspace{0.2em} 
    \end{minipage} \\
    {\small\raggedright 
    \textbf{Thought:} The initial webpage appears to have a ``honeydew'' background color and components (like headings and buttons) styled in ``dark olive green.'' However, I need to verify these colors across different pages or sections of the website, if possible. \\
    \textbf{Action:} Scroll [WINDOW]; down} & 
    {\small\raggedright
    \textbf{Thought:} The background color still appears to be ``honeydew,'' and the components like course cards, headings, and buttons seem styled with ``dark olive green.'' It aligns with the specified design so far, but further verification might involve checking other sections or interactions within the website. \\
    \textbf{Action:} Scroll [WINDOW]; down} & 
    {\small\raggedright
    \textbf{Thought:} The page didn't change after scrolling down. It seems we have reached the bottom of this section. The background color and component color still align with the design requirement. Additional steps to verify more sections of the site may not be necessary. I may now conclude the evaluation.\\
    \textbf{Action:} ANSWER; YES}  
\end{tabular}
\\ 
\midrule
\midrule
\textbf{Task:} Verify that the website uses `light goldenrod' as the background color and `olive drab' as the component color, as specified in the design requirements. \\
\textbf{Expected Result:} The website has a background color of `light goldenrod' and components (such as buttons, cards, headers, etc.) are styled with the color `olive drab', accurately reflecting the design instruction.
\\ 
\midrule
\begin{tabular}{@{}p{0.33\textwidth} p{0.33\textwidth} p{0.33\textwidth}@{}}
    \begin{minipage}{\linewidth}
        {\centering
        \includegraphics[width=\linewidth]{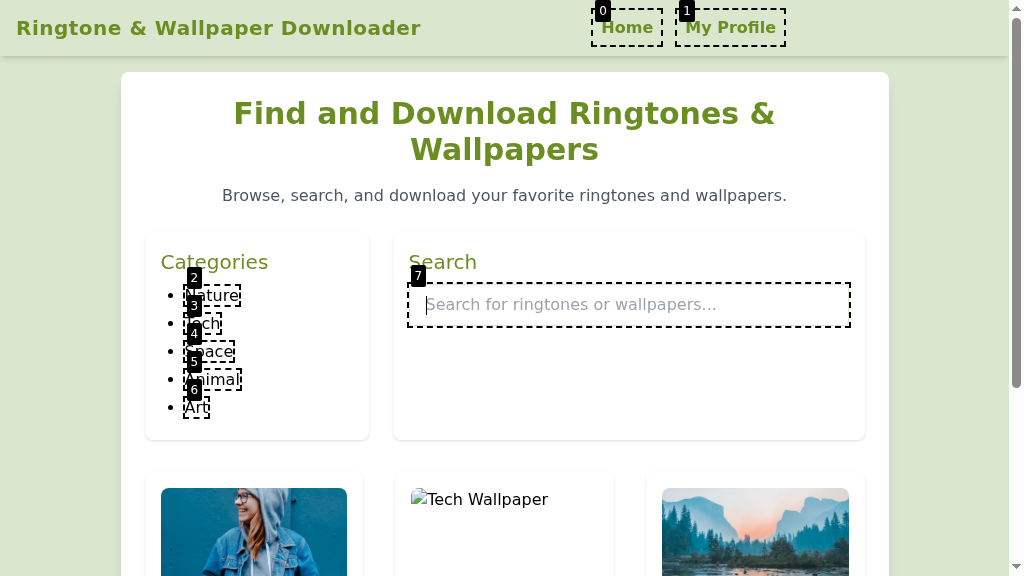} \par} 
        \vspace{0.2em} 
    \end{minipage} &
    \begin{minipage}{\linewidth}
        {\centering
        \includegraphics[width=\linewidth]{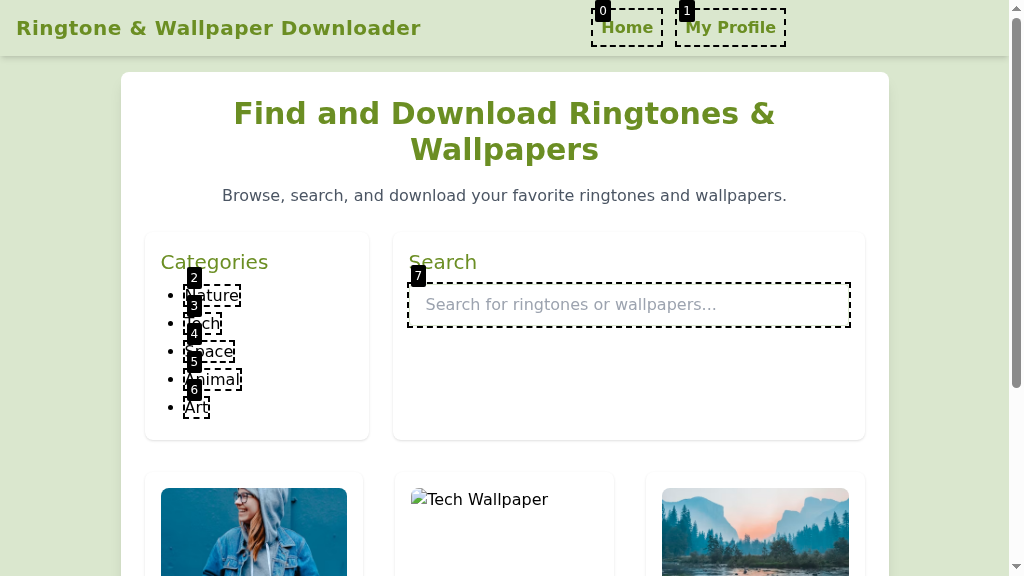} \par} 
        \vspace{0.2em} 
    \end{minipage} &
    \begin{minipage}{\linewidth}
        {\centering
        \includegraphics[width=\linewidth]{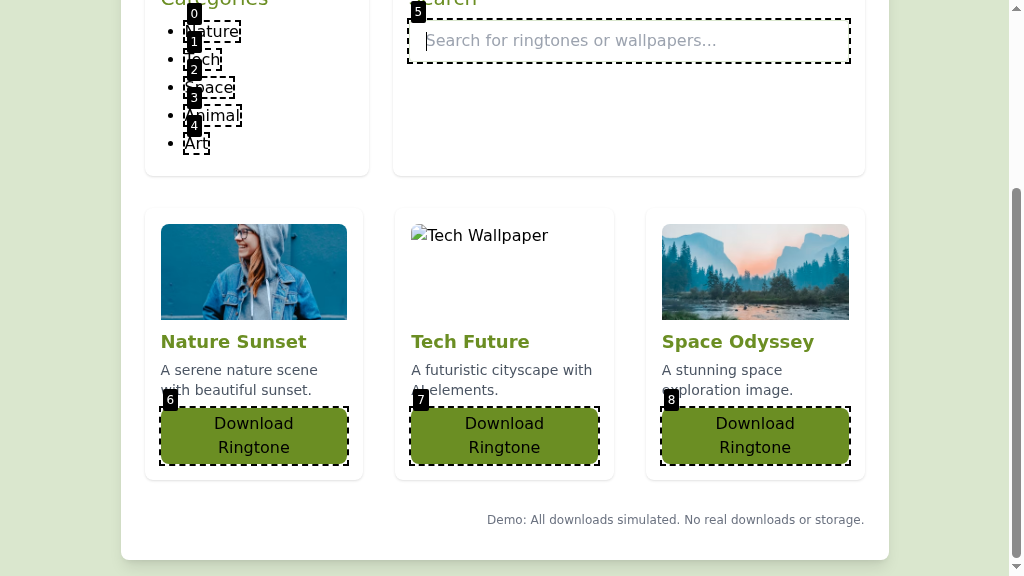} \par} 
        \vspace{0.2em} 
    \end{minipage} \\
    {\small\raggedright
    \textbf{Thought:} The webpage uses a light background color, which could be `light goldenrod', and components such as text and buttons appear in a greenish color which might correspond to `olive drab'. To validate this visually, I'll analyze specific webpage elements further. \\
    \textbf{Action:} Wait} & 
    {\small\raggedright
    \textbf{Thought:} The screenshot seems to confirm that the background color of the webpage aligns with a shade close to `light goldenrod,' and the component color (like heading, buttons) appears to resemble `olive drab.' It is visually consistent, but further interaction or verification might still be required.\\
    \textbf{Action:} Scroll [WINDOW]; down} & 
    {\small\raggedright
    \textbf{Thought:} The buttons labeled ``Download Ringtone'' have a greenish color that matches the description of `olive drab,' and the background still aligns with `light goldenrod.' All displayed elements comply visually with the task's design requirements. Further verification may not be necessary.\\
    \textbf{Action:} ANSWER; YES}
\end{tabular}
\\
\bottomrule
\end{tabular}
\label{tab:webgen_bench_ui_tests_yes}
\end{table}

\begin{table}[!ht]
\centering
\caption{UI agent testing processes resulting in ``PARTIAL''. The top and bottom rows show the UI verification tasks for two different websites, respectively.}
\setlength{\tabcolsep}{1pt} 
\begin{tabular}{@{}p{\textwidth}@{}}
\toprule
\textbf{Task:} Verify that the website uses `light gray' as the background color and `dark red' as the component color, as specified in the design requirements. \\
\textbf{Expected Result:} The website has a background color of `light gray' and components (such as buttons, cards, headers, etc.) are styled with the color `dark red', accurately reflecting the design instruction.
\\ 
\midrule
\begin{tabular}{@{}p{0.33\textwidth} p{0.33\textwidth} p{0.33\textwidth}@{}}
    \begin{minipage}{\linewidth}
        {\centering
        \includegraphics[width=\linewidth]{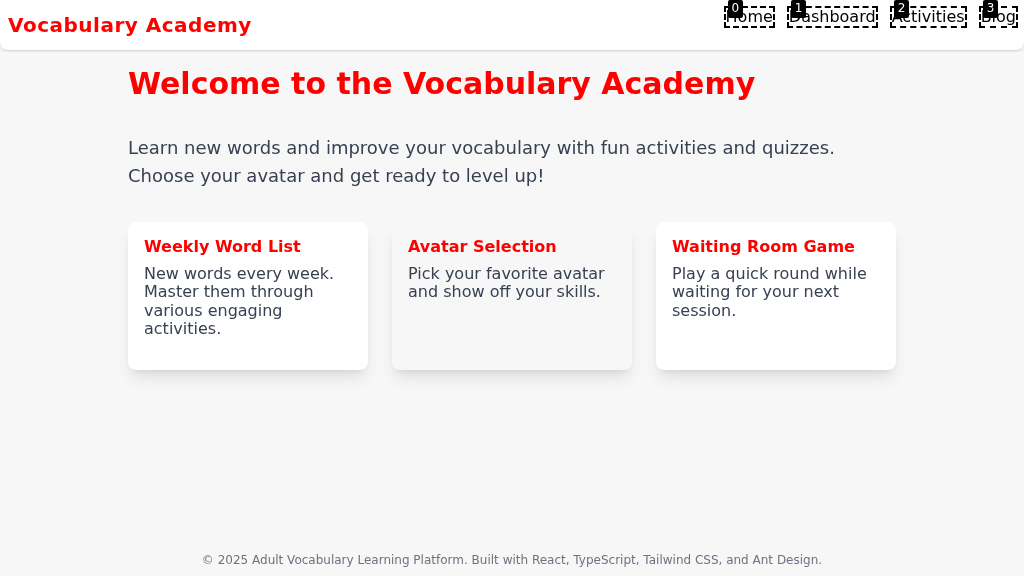}\par} 
        \vspace{0.2em} 
    \end{minipage} 
    &
    \begin{minipage}{\linewidth}
        {\centering
        \includegraphics[width=\linewidth]{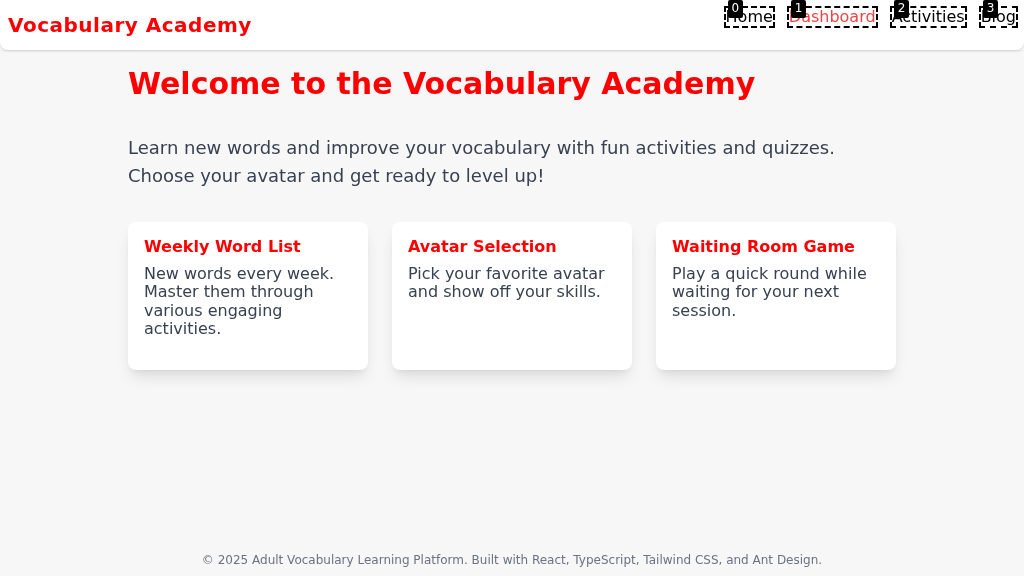} \par} 
        \vspace{0.2em} 
    \end{minipage} 
    &
    \begin{minipage}{\linewidth}
        {\centering
        \includegraphics[width=\linewidth]{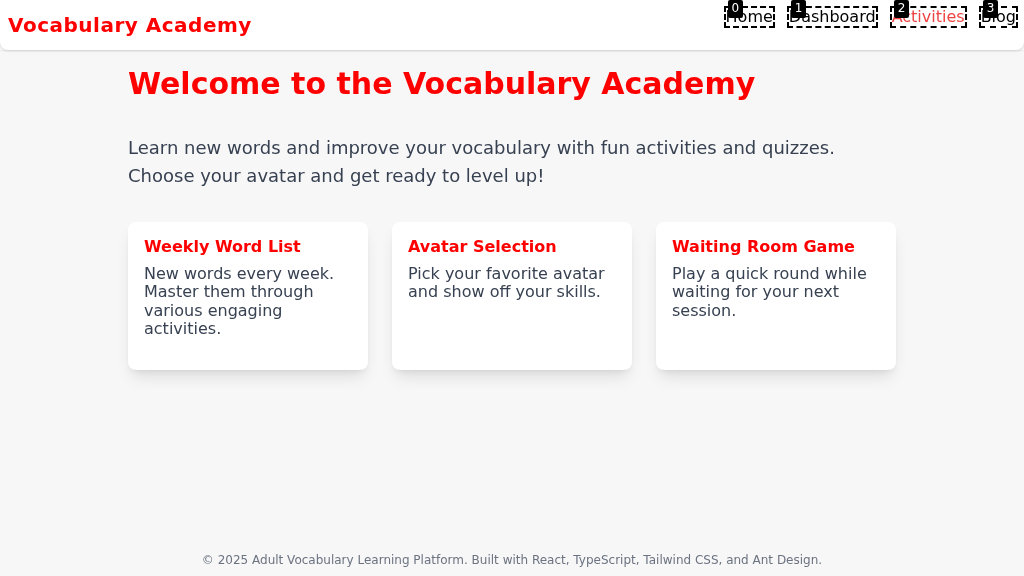} \par} 
        \vspace{0.2em} 
    \end{minipage} \\
    {\small\raggedright 
    \textbf{Thought:} The background of the webpage appears to be light gray, and some components (like headings and texts on cards) seem to use the color dark red. I need to further explore the interface to confirm the design requirement fully, such as validating all component styles and ensuring accuracy. \\
    \textbf{Action:} Click [1] (Dashboard) to explore further information about the webpage style and check other components.} & 
    {\small\raggedright
    \textbf{Thought:} The webpage remains unchanged, and the ``Dashboard'' section didn't open after the previous click. I need to attempt another navigation based on other options available to ensure I can explore the webpage for additional validations. \\
    \textbf{Action:} Click [2] (Activities) to explore the ``Activities'' section.} & 
    {\small\raggedright
    \textbf{Thought:} The webpage still remains unchanged after multiple navigation attempts. It might be possible that the clickable options are not leading to other sections. Based on the observation seen so far, I have to make a final assessment. \\
    \textbf{Action:} ANSWER; PARTIAL}  
\end{tabular}
\\ 
\midrule
\midrule
\textbf{Task:} Validate the ability to view historical air quality index (AQI) data for the past two weeks. \\
\textbf{Expected Result:} Users should be able to access a time series or historical graph/chart that shows AQI data for each of the last 14 days.
\\ 
\midrule
\begin{tabular}{@{}p{0.33\textwidth} p{0.33\textwidth} p{0.33\textwidth}@{}}
    \begin{minipage}{\linewidth}
        {\centering
        \includegraphics[width=\linewidth]{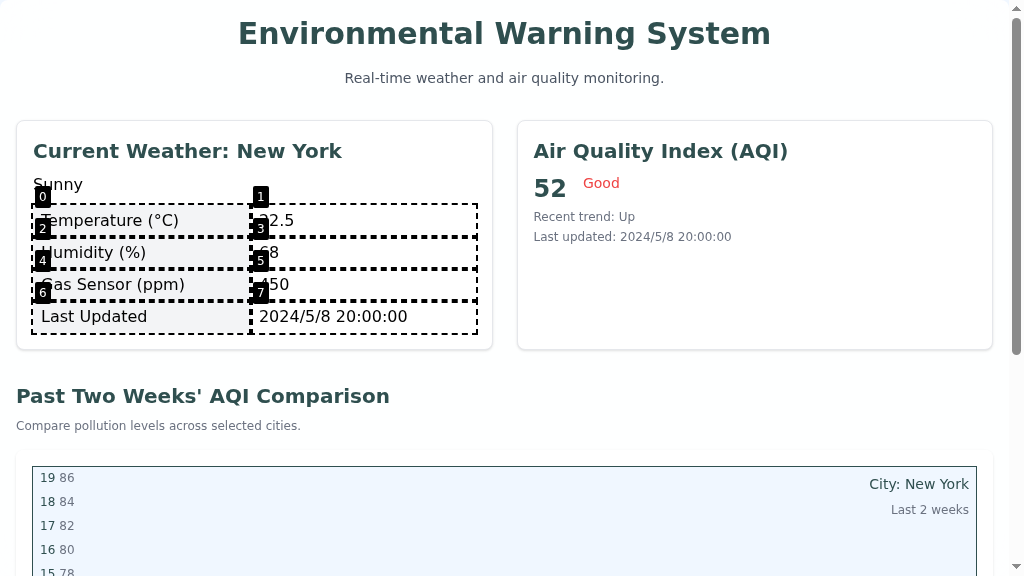} \par} 
        \vspace{0.2em} 
    \end{minipage} &
    \begin{minipage}{\linewidth}
        {\centering
        \includegraphics[width=\linewidth]{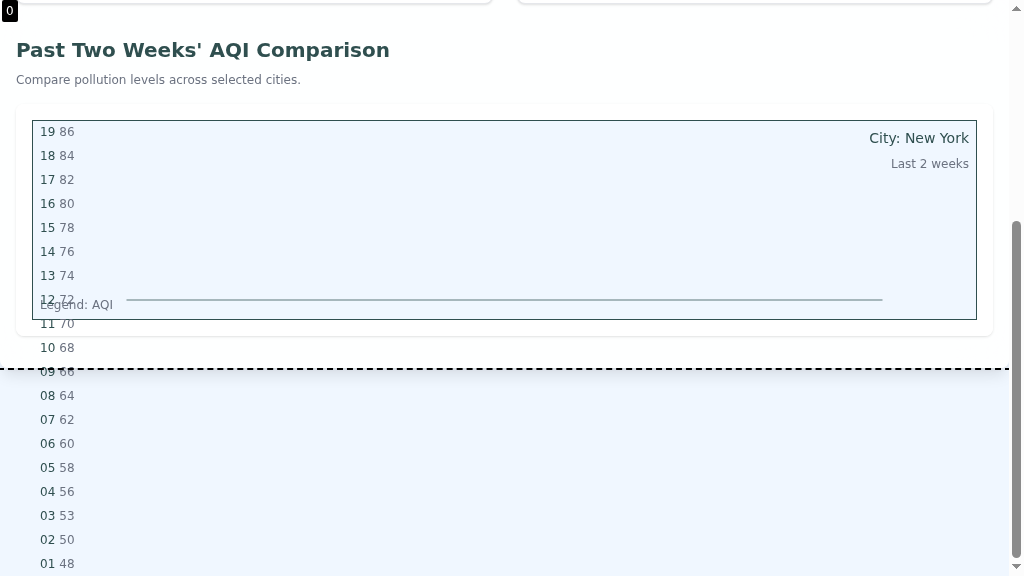} \par} 
        \vspace{0.2em} 
    \end{minipage} &
    \begin{minipage}{\linewidth}
        {\centering
        \includegraphics[width=\linewidth]{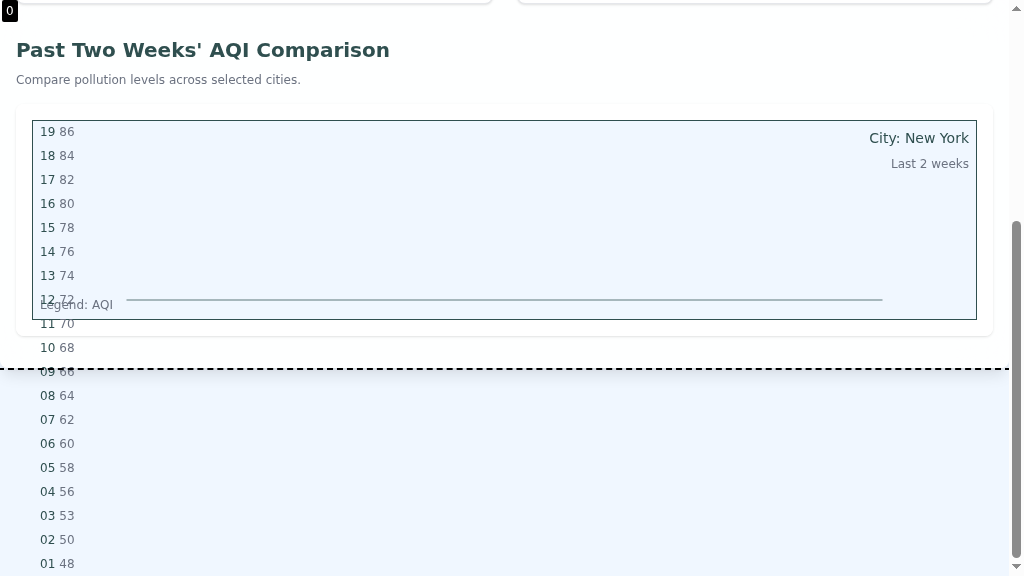} \par} 
        \vspace{0.2em} 
    \end{minipage} \\
    {\small\raggedright
    \textbf{Thought:} The webpage includes a section labeled ``Past Two Weeks' AQI Comparison'' which appears to list AQI data, but further details or a time series graph/chart showing the AQI data for each of the last 14 days is not immediately visible. The list may require scrolling to view more information. \\
    \textbf{Action:} Scroll [WINDOW]; down} & 
    {\small\raggedright
    \textbf{Thought:} The ``Past Two Weeks' AQI Comparison'' section showcases numerical AQI data for the last 14 days, which matches the expected result in terms of tabular data representation. However, a time series graph/chart is still not visible. To ensure, I should continue scrolling for additional views or graphical representations if present. \\
    \textbf{Action:} Scroll [WINDOW]; down} & 
    {\small\raggedright
    \textbf{Thought:} After scrolling further, no additional graphical representation like a time series chart for the last 14 days is visible. Only numerical AQI data in a tabular fashion is presented, which fulfills part of the task but not entirely.\\
    \textbf{Action:} ANSWER; PARTIAL}
\end{tabular}
\\
\bottomrule
\end{tabular}
\label{tab:webgen_bench_ui_tests_partial}
\end{table}

\begin{table}[!ht]
\centering
\caption{UI agent testing processes resulting in ``NO''. The top and bottom rows show the UI verification tasks for two different websites, respectively.}
\setlength{\tabcolsep}{1pt} 
\begin{tabular}{@{}p{\textwidth}@{}}
\toprule
\textbf{Task:} Confirm navigation from the home page to the `About Us' page. \\
\textbf{Expected Result:} Clicking the `About Us' link takes the user to the `About Us' page without any errors.
\\ 
\midrule
\begin{tabular}{@{}p{0.33\textwidth} p{0.33\textwidth} p{0.33\textwidth}@{}}
    \begin{minipage}{\linewidth}
        {\centering
        \includegraphics[width=\linewidth]{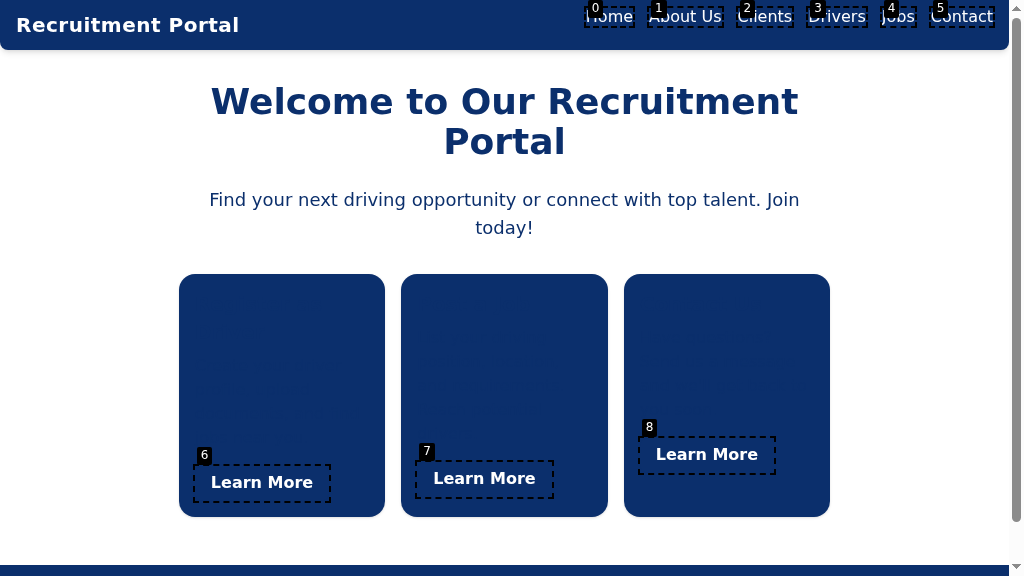}\par} 
        \vspace{0.2em} 
    \end{minipage} 
    &
    \begin{minipage}{\linewidth}
        {\centering
        \includegraphics[width=\linewidth]{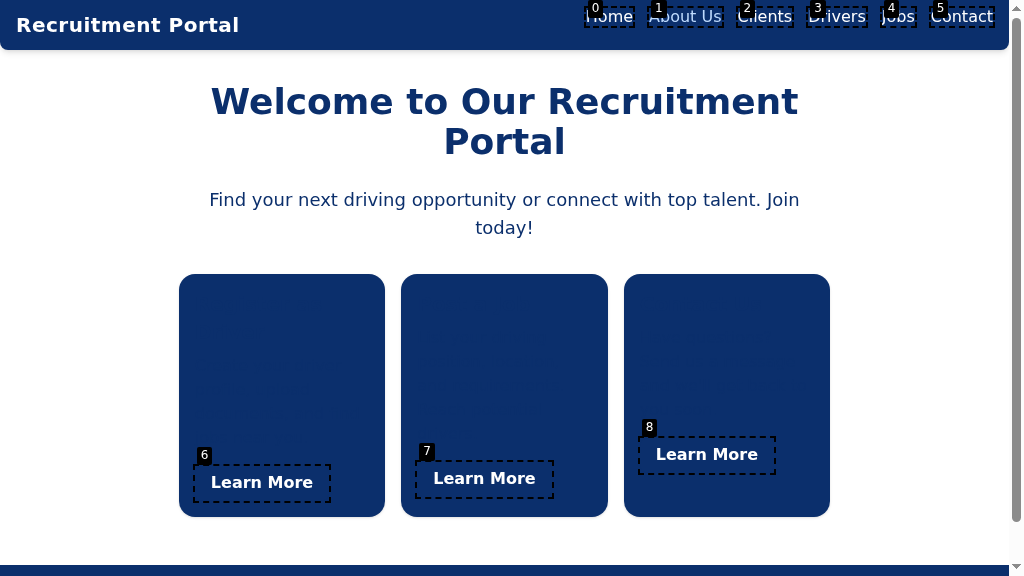} \par} 
        \vspace{0.2em} 
    \end{minipage} 
    &
    \begin{minipage}{\linewidth}
        {\centering
        \includegraphics[width=\linewidth]{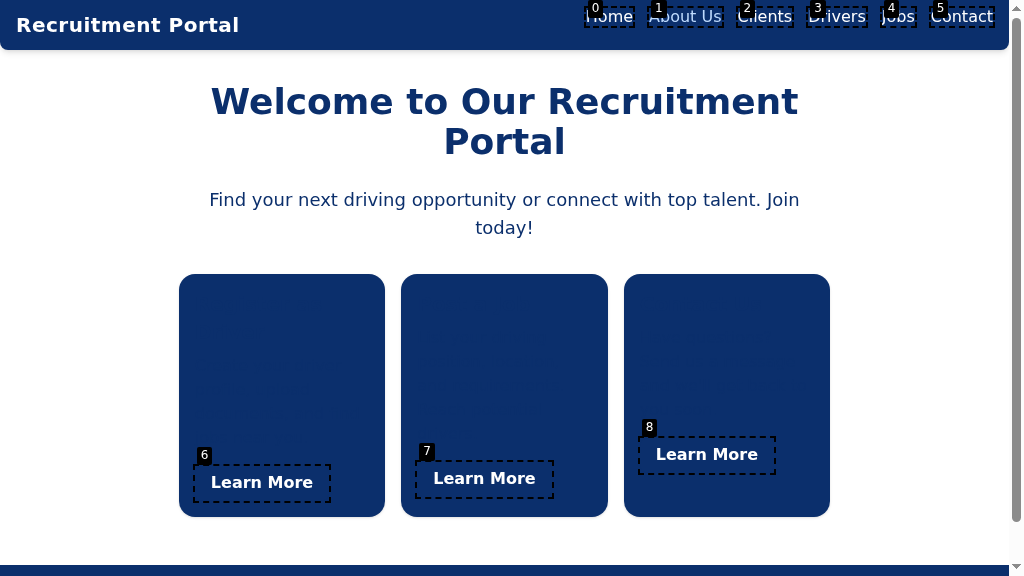} \par} 
        \vspace{0.2em} 
    \end{minipage} \\
    {\small\raggedright 
    \textbf{Thought:} The webpage labeled elements include navigation links such as ``About Us.'' To confirm the navigation from the Home page to the ``About Us'' page, I'll click the ``About Us'' link (Label [1]). \\
    \textbf{Action:} Click [1].} & 
    {\small\raggedright
    \textbf{Thought:} The screen remains on the same page as before, suggesting the ``About Us'' link might not be functional. I will attempt to click it again to ensure it wasn't a one-time issue. \\
    \textbf{Action:} Click [1].} & 
    {\small\raggedright
    \textbf{Thought:} The screen still remains unchanged after clicking the ``About Us'' link twice, confirming that the link is non-functional or does not navigate properly. Now, I will answer based on the observations. \\
    \textbf{Action:} ANSWER; NO}  
\end{tabular}
\\ 
\midrule
\midrule
\textbf{Task:} Test the supplier management interface for adding or updating a golf package supplier. \\
\textbf{Expected Result:} The user should be able to successfully add or update supplier details with confirmation that changes have been saved in the system.
\\ 
\midrule
\begin{tabular}{@{}p{0.33\textwidth} p{0.33\textwidth} p{0.33\textwidth}@{}}
    \begin{minipage}{\linewidth}
        {\centering
        \includegraphics[width=\linewidth]{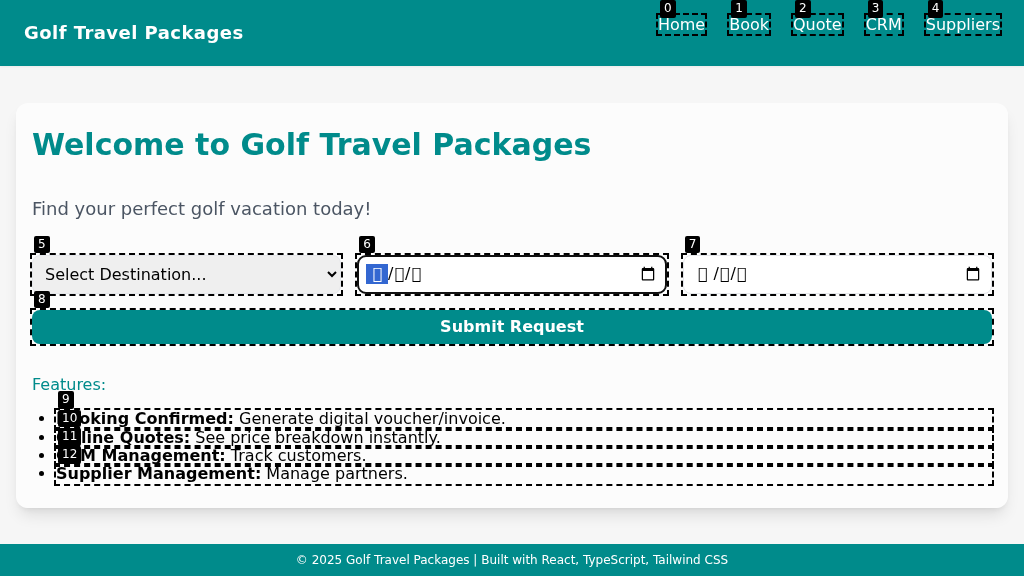} \par} 
        \vspace{0.2em} 
    \end{minipage} &
    \begin{minipage}{\linewidth}
        {\centering
        \includegraphics[width=\linewidth]{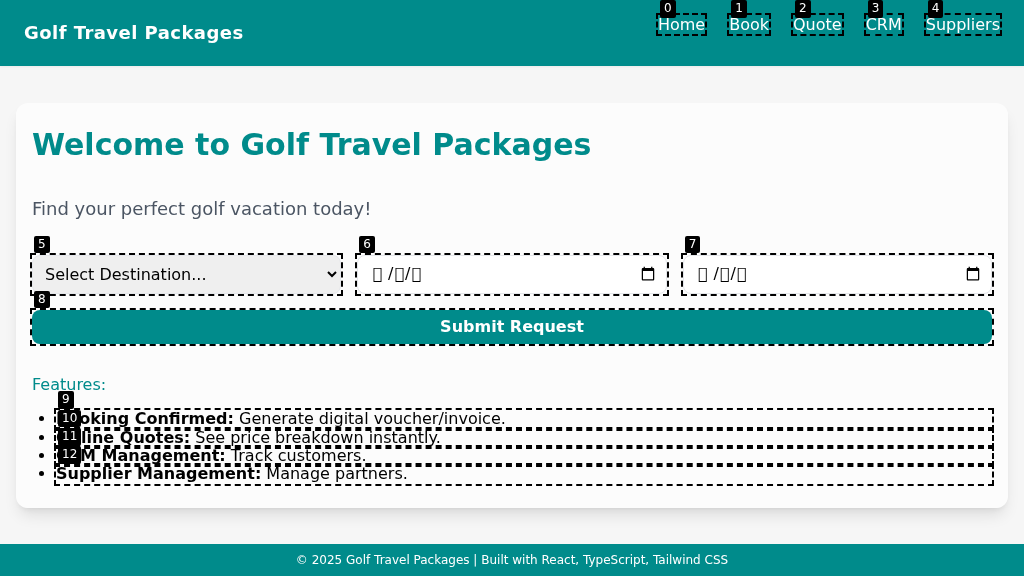} \par} 
        \vspace{0.2em} 
    \end{minipage} &
    \begin{minipage}{\linewidth}
        {\centering
        \includegraphics[width=\linewidth]{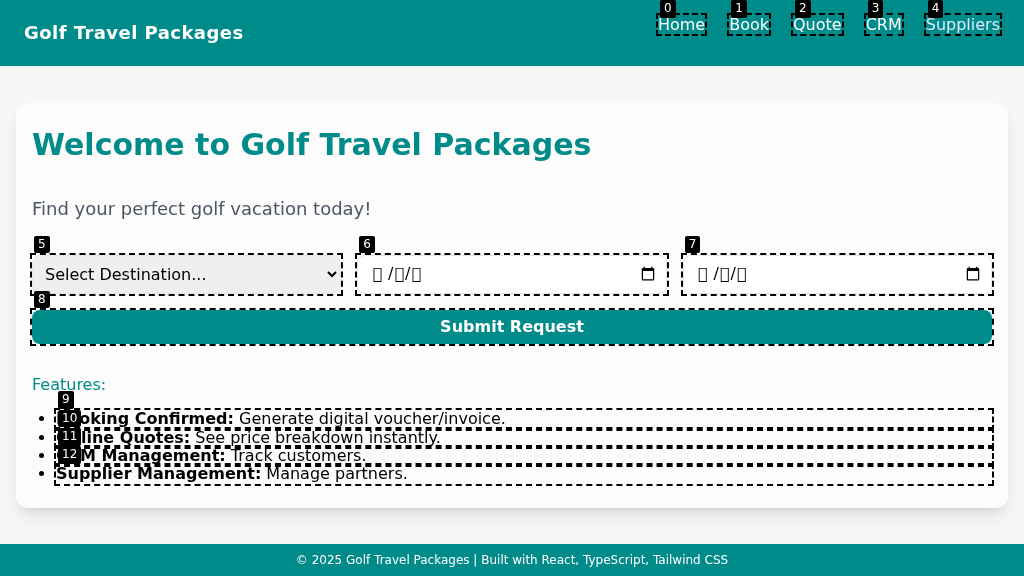} \par} 
        \vspace{0.2em} 
    \end{minipage} \\
    {\small\raggedright
    \textbf{Thought:} To manage suppliers, the ``Suppliers'' navigation option (label [4]) seems appropriate. This should provide access to the supplier management interface. \\
    \textbf{Action:} Click [4].} & 
    {\small\raggedright
    \textbf{Thought:} The page remains unchanged, and no new supplier management interface is visible after the previous action. It seems the ``Suppliers'' option ([4]) did not navigate as expected. I will try interacting with it again to ensure the action registers. \\
    \textbf{Action:} Click [4].} & 
    {\small\raggedright
    \textbf{Thought:} Even after performing the action twice, the page remains unchanged, and the ``Suppliers'' button ([4]) does not navigate to or display the expected supplier management interface. I suspect it might not be functioning or implemented correctly. I cannot proceed further on this page. \\
    \textbf{Action:} ANSWER; NO}
\end{tabular}
\\
\bottomrule
\end{tabular}
\label{tab:webgen_bench_ui_tests_no}
\end{table}




\end{document}